\documentclass[10pt,journal,compsoc]{IEEEtran}

\ifCLASSOPTIONcompsoc
  \usepackage[nocompress]{cite}
\else
  \usepackage{cite}
\fi

\usepackage[utf8]{inputenc} 
\usepackage{hyperref}       

\usepackage{amsmath,amsfonts}
\usepackage{algorithmic}
\usepackage{algorithm}
\usepackage{array}
\usepackage[caption=false,font=normalsize,labelfont=sf,textfont=sf]{subfig}
\usepackage{textcomp}
\usepackage{stfloats}
\usepackage{url}
\usepackage{graphicx}
\usepackage{cite}
\usepackage{booktabs}
\usepackage{multirow}
\usepackage{marvosym}
\usepackage{ragged2e}
\usepackage{xcolor}

\usepackage{amssymb}
\usepackage{mathtools}
\usepackage{amsthm}
\usepackage{dsfont}

\newtheorem{assumption}{Assumption}
\newtheorem{proposition}{Proposition}
\newtheorem{lemma}{Lemma}
\newtheorem{theorem}{Theorem}

\newcommand{\beginsupplement}{%
    \setcounter{table}{0}
    \renewcommand{\thetable}{S\arabic{table}}%
    \setcounter{figure}{0}
    \renewcommand{\thefigure}{S\arabic{figure}}%
    \setcounter{equation}{0}
    \renewcommand{\theequation}{S\arabic{equation}}%
    \setcounter{section}{0}
    \renewcommand{\thesection}{S\arabic{section}}%
    \renewcommand{\thesubsection}{S\arabic{section}.\arabic{subsection}}%
}

\def\algo{{\textsc{DeCon}}}

\usepackage[capitalize]{cleveref}
\Crefname{section}{Sec.}{Secs.}
\Crefname{section}{Section}{Sections}
\Crefname{table}{Tab.}{Tabs.}
\Crefname{table}{Table}{Tables}
\Crefname{assumption}{Assumption}{Assumptions}

\newcommand{\ms}[2]{{#1}{\footnotesize $\,\pm${#2}}}

\begin{document}



\title{Decouple then Converge: Handling Unknown Unlabeled Distributions in Long-Tailed Semi-Supervised Learning}

\author{Kai~Gan,
        Tong~Wei*,
        and~Min-Ling~Zhang,~\IEEEmembership{Senior Member,~IEEE}
\IEEEcompsocitemizethanks{\IEEEcompsocthanksitem Kai~Gan, Tong~Wei (* corresponding author), and Min-Ling~Zhang are with the School of Computer
Science and Engineering, Southeast University, Nanjing 210096, China,
and the Key Laboratory of Computer Network and Information Integration (Southeast University), Ministry of Education, China. E-mail:
\{gank, weit, zhangml\}@seu.edu.cn.}
}



\markboth{IEEE TRANSACTIONS ON PATTERN ANALYSIS AND MACHINE INTELLIGENCE}%
{IEEE TRANSACTIONS ON PATTERN ANALYSIS AND MACHINE INTELLIGENCE}


\IEEEtitleabstractindextext{%

\begin{abstract}\justifying
While long-tailed semi-supervised learning (LTSSL) has attracted growing attention in many real-world classification tasks, existing LTSSL algorithms typically assume that labeled and unlabeled data share nearly identical class distributions. When this assumption is violated, these methods can perform poorly because they rely on biased model-generated pseudo-labels. To address this issue, we propose a simple yet effective approach called \algo\ for LTSSL with unknown unlabeled class distributions. Specifically, \algo\ decouples learning into two specialized branches: a standard branch that focuses on head classes and a balanced branch that focuses on tail classes. During training, the two branches interact and gradually converge, allowing them to complement each other and ultimately achieve strong performance across all classes. Despite its simplicity, we show that \algo\ achieves state-of-the-art performance on a variety of standard LTSSL benchmarks, e.g., an averaged 2.7\% absolute increase in test accuracy against existing algorithms when the class distributions of labeled and unlabeled data are mismatched. Even when the class distributions are identical, \algo\ consistently outperforms many sophisticated LTSSL algorithms. Furthermore, we conduct extensive ablation analyses to tease apart the factors that are the most important to the success of \algo. The source code is available at \url{https://github.com/Gank0078/DeCon}.
\end{abstract}

}

\maketitle

\IEEEdisplaynontitleabstractindextext

\IEEEpeerreviewmaketitle

\section{Introduction}
\IEEEPARstart{S}{emi-supervised} learning (SSL) is an effective way to use unlabeled data to improve the generalization of deep neural networks (DNNs) \cite{he2016deep,krizhevsky2017imagenet,amodei2016deep} when only a limited amount of labeled data is accessible \cite{tarvainen2017mean,miyato2018virtual,berthelot2019mixmatch,sohn2020fixmatch,ouyang2025semantic,dong2024pseudo,dong2022rethinking}. The core idea of most SSL algorithms is to generate pseudo-labels for unlabeled data and select confident ones to train models. Recent progress on SSL has revealed promising performance in various tasks, such as image recognition \cite{sohn2020fixmatch} and text categorization \cite{DBLP:conf/nips/XieDHL020,weit2020tnnls}. Popular SSL methods \cite{zhang2021flexmatch,wang2022freematch,chen2023softmatch} employ a confidence threshold to select reliable pseudo-labels, thereby enhancing the model's generalization capability. However, most existing SSL algorithms assume that the datasets are class-balanced, i.e., each class is associated with an equivalent number of samples in both labeled and unlabeled datasets. In contrast, class distributions in many real-world tasks are long-tailed \cite{zhou2020bbn,xiang2020learning,wang2020long,li2022nested,cui2021parametric}. It is well known that classifiers trained on long-tailed datasets tend to be biased towards majority classes, leading to low test accuracy on minority classes\cite{liu2019large,Wei_2021_RoLT,beierxERM}.

To improve the performance, many long-tailed semi-supervised learning (LTSSL) algorithms have been proposed to generate unbiased pseudo-labels. They pursue class-balanced classifiers using techniques including re-sampling\cite{lee2021abc}, re-weighting\cite{DBLP:conf/icml/LaiWGCC22}, label smoothing\cite{wei2022transfer}, and pseudo-label alignment\cite{kim2020distribution,wei2021crest,ye2023bridging}. These algorithms have shown strong generalization for the minority class by assuming that the class distributions of labeled and unlabeled data are almost identical. However, this assumption is frequently violated in real-world applications, for instance, if the labeled and unlabeled data are collected from different tasks. Unlabeled data may have a large class distribution deviation from labeled data, and the use of an erroneous assumption can severely deteriorate performance\cite{oh2022daso,DBLP:conf/icml/LaiWGCC22}. Recently, some LTSSL methods \cite{oh2022daso,wei2023towards} attempt to address this class distribution shift between labeled and unlabeled data. However, these methods consider specific distributions of unlabeled data, rendering them ineffective in handling varying unlabeled class priors.


\begin{figure*}[t]
\centering
\subfloat[\textit{Consistent}]{\includegraphics[width=2.3in]{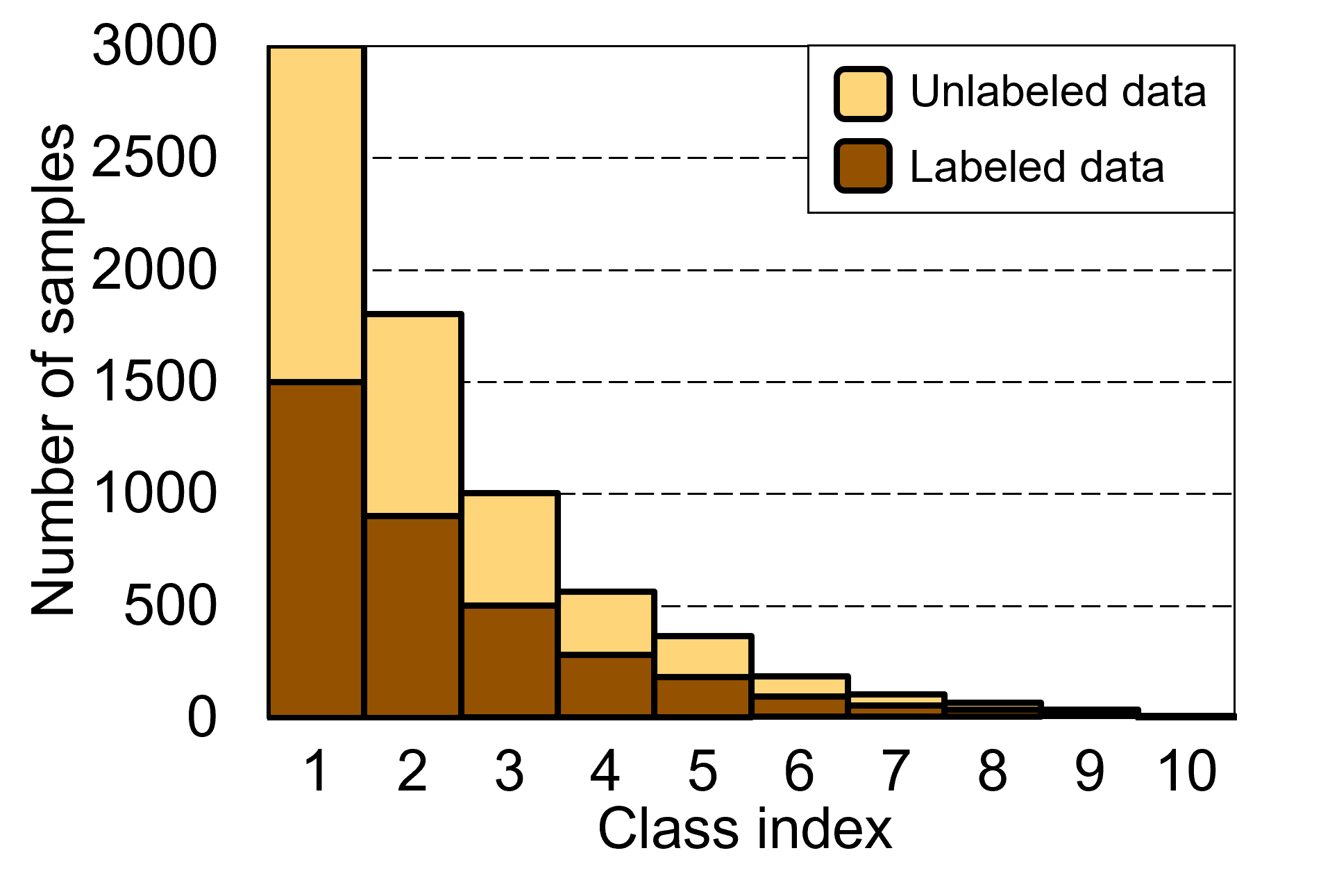}%
\label{fig:dist_con}}
\hfil
\subfloat[\textit{Uniform}]{\includegraphics[width=2.3in]{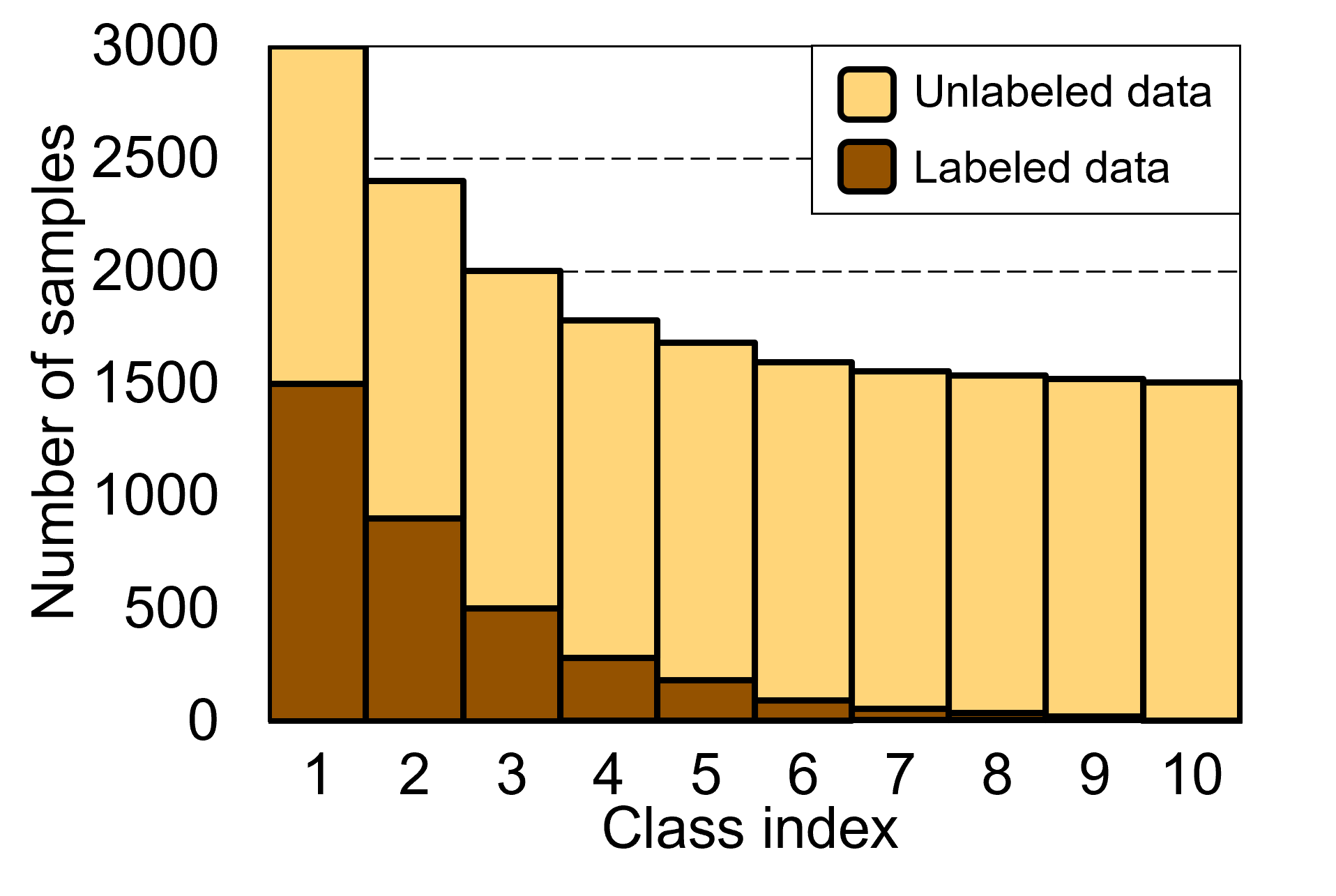}
\label{fig:dist_uni}}
\hfil
\subfloat[\textit{Reversed}]{\includegraphics[width=2.3in]{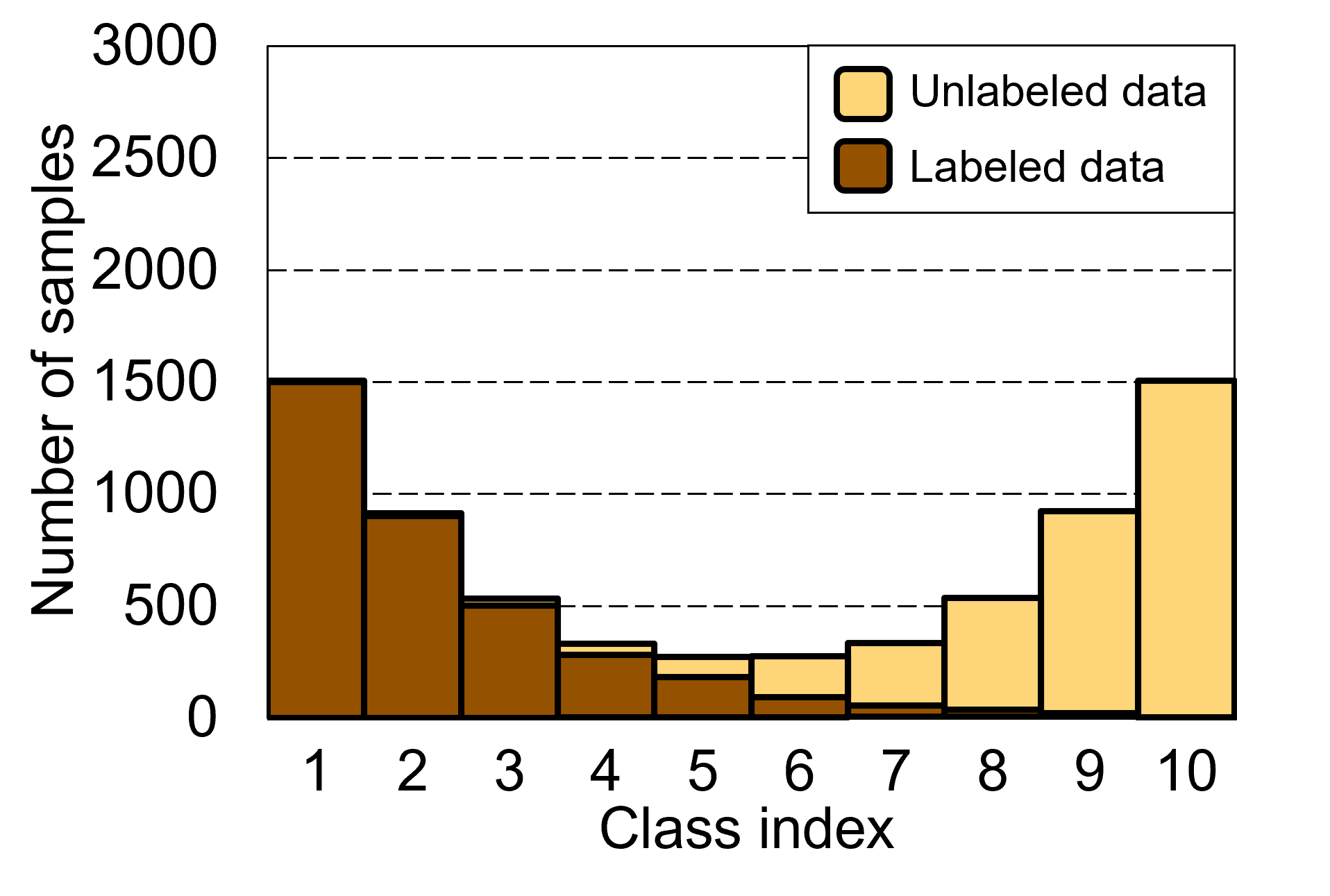}
\label{fig:dist_rev}}

\caption{\labelcref{fig:dist_con,fig:dist_uni,fig:dist_rev}): Three typical types of class distribution of unlabeled data: \textit{consistent}, \textit{uniform}, and \textit{reversed}.}
\label{fig:dist}
\end{figure*}

This paper studies an underexplored yet practical LTSSL setting in which the labeled data follow a long-tailed class distribution, whereas the unlabeled samples are drawn from an unknown class distribution. In particular, we start with three representative types of class distributions of unlabeled data following \cite{oh2022daso}, i.e., \textit{consistent}, \textit{uniform}, and \textit{reversed}, as illustrated in \Cref{fig:dist}. It should be noted that this paper focuses on a wide range of unlabeled class priors that are not limited to three representative distributions. We then propose a simple but effective method \algo, built on the popular SSL algorithm FixMatch \cite{sohn2020fixmatch}, to improve learning performance by tackling mismatched class priors between labeled and unlabeled data.




In realistic LTSSL settings, if we directly apply SSL methods to the imbalanced labeled and unlabeled data, the model's predictions will significantly favor the head classes. However, recent LTSSL methods \cite{wei2023towards,oh2022daso} that introduce balanced modules can result in biased model predictions towards tail classes, thus affecting the performance of head classes. To this end, our method \algo, first decouples the learning objective by employing a dual-branch architecture, designed to handle head and tail classes separately. The first branch, referred to as the \textbf{standard branch}, aims to maintain the performance of head classes. The standard branch does not implement any adjustment on pseudo-labels, ensuring it does not overly shift towards the tail classes like recent LTSSL methods. The other branch, termed the \textbf{balanced branch}, attempts to maintain the performance on tail classes by adjusting the model's predictions on the unlabeled data. However, even if both branches ideally achieve their intended objectives, integrating two branches to achieve favorable performance across all classes during inference remains a significant challenge, which has been extensively discussed in previous studies \cite{zhang2022self,wang2020long,li2022nested} and a satisfactory solution to this problem remains elusive. Therefore, in this paper, we do not combine the predictions of the two branches as the final inference result. Instead, we facilitate interactions between the branches during the training, enabling both to achieve superior performance on both head and tail classes progressively.

Specifically, we avoid directly applying any balancing strategy to the pseudo-labels derived from unlabeled data for the standard branch, allowing it to achieve strong performance on head classes \cite{zhang2022self}. In addition, for the balanced branch, we apply a popular rebalancing strategy in long-tail learning, i.e., logit adjustment \cite{menon2020long}, to alleviate the model's excessive focus on head classes, thus enabling it to sustain strong performance on tail classes. Regarding the interaction between two branches, this is primarily manifested in two aspects: i) the output of the standard branch for unlabeled samples undergo rebalancing adjustment and then serve as the targets for the balanced branch, whereas ii) the distribution predicted by the balanced branch for unlabeled data will be used to align the labeled data in standard branch. As training progresses, it is observed that the predictive tendencies of the two branches gradually converge. Furthermore, both branches, especially the balanced branch, can achieve high pseudo-label accuracy on unlabeled datasets and achieve outstanding performance on both head and tail classes. During the inference phase, we significantly improved the model's performance on test set using post-hoc adjustment, which further balances the predictions.


%


We demonstrate the effectiveness of the proposed approach under various realistic LTSSL scenarios by varying the class distributions of unlabeled data. Despite its simplicity, the proposed algorithm improves recent LTSSL algorithms in all test cases, e.g., our method improves DARP\cite{wei2021crest}, DASO\cite{oh2022daso}, ACR\cite{wei2023towards} with up to \textbf{11.2\%}, \textbf{8.1\%}, and \textbf{2.4\%} absolute increase in test accuracy, respectively.
In addition to three types of representative class distributions, i.e., \textit{consistent}, \textit{uniform}, and \textit{reversed}, we also test \algo\ under many other class distributions in Supplementary Sec. S1. As expected, our method significantly improves performance when class distributions are mismatched between labeled and unlabeled data.




\section{Related Work}

\noindent\textbf{Long-tail learning.} Research on long-tail learning has attracted increasing attention recently, which is a challenging and practical task. The main directions to improve the performance include: 1) data manipulation \cite{zhou2020bbn,kang2019decoupling}, 2) representation learning \cite{zhong2021improving,cui2021parametric}, and model output adjustment \cite{menon2020long,ren2020balanced}. Data manipulation is often encompassed by the design of resampling strategies and the implementation of data augmentation techniques. Numerous studies have demonstrated performance enhancements through the adoption of a two-stage training paradigm, where the initial stage is dedicated to learning data representations, followed by a subsequent stage that focuses on classifier training \cite{kang2019decoupling}. The adjustment of the model's output can be executed during the training phase through the optimization of unbiased loss functions, or post-hoc adjustments can be implemented. In this paper, considering the firm statistical grounding and competitive empirical performance of the logit adjustment \cite{menon2020long}, \algo\ predominantly employs this technique to mitigate the effects of the long-tailed distribution.

\noindent\textbf{Semi-supervised learning.} A popular class of Semi-supervised learning (SSL) algorithms use unlabeled data to improve the performance via learning to predict the pseudo-labels produced by the model, which can be viewed as a self-training process \cite{tarvainen2017mean,miyato2018virtual,berthelot2019mixmatch,berthelot2019remixmatch}. Recent SSL algorithms \cite{sohn2020fixmatch,berthelot2019remixmatch} combine pseudo labeling and consistency regularization, which encourages similar predictions between two different views of an image, to improve the robustness of DNNs. As a representative approach, FixMatch \cite{sohn2020fixmatch} achieves significantly better performance than many other SSL algorithms in the image recognition task. Hence, the performance of the SSL algorithm is quite sensitive to the quality of pseudo-labels. To improve the quality of pseudo-labels, \cite{zhang2021flexmatch,wang2022freematch,chen2023softmatch} employ thresholding or weighting schemes to select high-quality pseudo-labels for training. FlexMatch \cite{zhang2021flexmatch} allocates unique thresholds to individual classes, considering the varying degrees of learning difficulty and the aggregate pace of learning. FreeMatch \cite{wang2022freematch} seamlessly integrates global and local threshold adjustments, complemented by a class-fairness regularization technique that promotes diversity in the model's predictions. SoftMatch \cite{chen2023softmatch} effectively balances the trade-off between quantity and quality through a unified approach by sample reweighting, facilitated by a soft confidence threshold mechanism. However, most existing SSL algorithms assume balanced class distributions of labeled and unlabeled data, resulting in poor generalization of the minority class due to biased pseudo-labels. Our \algo\ is built upon FixMatch, which has been frequently used as the base model for SSL tasks.

\noindent\textbf{Long-tailed semi-supervised learning.} Long-tailed semi-supervised learning (LTSSL) has received significant attention for its practicality in many real-world tasks. For instance, DARP \cite{kim2020distribution} and CReST \cite{wei2021crest} propose eliminating biased pseudo-labels generated by the model by distribution alignment to refine pseudo-labels according to the class distribution of labeled data. ABC\cite{lee2021abc} uses an auxiliary balanced classifier trained by down-sampling majority classes to improve the generalization. CoSSL \cite{fan2022cossl} designs a novel feature enhancement module for the minority class using mixup \cite{DBLP:conf/iclr/ZhangCDL18} to train balanced classifiers. Although these algorithms can significantly enhance performance, they assume identical class distributions of labeled and unlabeled data which is not always satisfied in real-world scenarios. DASO \cite{oh2022daso} proposes to handle this issue by employing a dynamic combination of linear and semantic pseudo-labels based on the current estimated class distribution of unlabeled data. It is noted that the accuracy of semantic pseudo-labels in DASO relies on the discrimination of learned representations. However, long-tailed class distribution negatively impacts representation learning, reducing the reliability of semantic pseudo-labels. SimPro \cite{du2024simpro}, which is grounded in a probabilistic model, refines the expectation-maximization (EM) algorithm by explicitly decoupling the modeling of conditional and marginal class distributions to handle unknown unlabeled data distribution. The recent state-of-the-art method ACR \cite{wei2023towards} employs three fixed anchor distributions to dynamically tune the intensity of logit adjustment, thereby effectively addressing the challenges presented by \textit{consistent}, \textit{uniform}, and \textit{reversed}. Nevertheless, ACR has been observed to perform optimally only with the three specified unlabeled data distributions due to the rigidity of the anchor distributions. When confronted with more diverse unlabeled data distributions, the efficacy of ACR is notably diminished. \algo\ imposes no assumptions on unlabeled data distribution, enabling robust performance across diverse unlabeled data class prior. Notably, when considering \textit{consistent}, \textit{uniform}, and \textit{reversed}, \algo\ still significantly outperforms ACR by a considerable margin.

\section{The Proposed Method}

In this section, we first introduce the preliminaries related to LTSSL tasks. Next, we present the \algo\ to handle the unknown class distribution of unlabeled data with two processes and post-hoc adjustment to further improve the performance during the inference. Finally, we discuss the key distinctions between \algo\ and ACR to highlight the superiority of \algo. 


\subsection{Preliminaries}
\noindent
\textbf{Problem setup.} In LTSSL tasks, we are given a labeled dataset $\mathcal{D}^{l}=\{(x_i^{(l)},y_i^{(l)})\}^{N}_{i=1}$ of size $N$ and an unlabeled dataset $\mathcal{D}^{u}=\{x_j^{(u)}\}^{M}_{j=1}$ of size $M$, where $x_i^{(l)}, x_j^{(u)} \in \mathbb{R}^d$ are $d$-dimensional training samples. Each labeled sample is associated with a ground-truth label $y_i^{(l)} \in \{0, 1\}^{C}$, where $C$ is the number of classes. When $y_i^{(l)}$ appears as a subscript, it denotes the corresponding class index. Let $N_c$ denote the number of labeled samples in class $c$. We assume $N_1 \ge N_2 \ge ... \ge N_C$, and denote the imbalance ratio of the labeled dataset by $\gamma_l = \frac{N_1}{N_C}$. Similarly, let $M_c$ denote the number of unlabeled samples in class $c$, and define the imbalance ratio of the unlabeled dataset as $\gamma_u = \frac{\max_c M_c}{\min_c M_c}$. Since we do not assume any prior knowledge of the unlabeled class distribution, we mainly consider three representative settings, namely \textit{consistent}, \textit{uniform}, and \textit{reversed}. Specifically, i) in the \textit{consistent} setting, $M_1 \ge M_2 \ge \dots \ge M_C$ and $\gamma_u = \gamma_l$; ii) in the \textit{uniform} setting, $M_1 = M_2 = \dots = M_C$, i.e., $\gamma_u=1$; iii) in the \textit{reversed} setting, $M_1 \le M_2 \le \dots \le M_C$ and $\gamma_u = 1/\gamma_l$. We also consider more diverse unlabeled distributions in Supplementary Sec. S1, including \textit{middle} and \textit{headtail} \cite{du2024simpro}. Our goal is to train a classifier using $\mathcal{D}^{l}$ and $\mathcal{D}^{u}$. Unless otherwise specified, $f: \mathbb{R}^d \rightarrow \mathbb{R}^C$ denotes the logit function parameterized by $\theta$, and $\delta(f(x))\in[0,1]^C$ denotes its softmax probability vector.


\noindent
\textbf{Semi-supervised learning.} Many existing SSL algorithms seek to minimize a supervised classification loss on labeled data and an unsupervised regularizer on unlabeled data. Formally, the objective function is given as follows:
\begin{equation}
    \min_{\theta \in \Theta} \underbrace{\sum_{i=1}^N \ell(f(x_i^{(l)}; \theta), y_i^{(l)})}_{\mathsf{supervised} \ (\mathcal{L}_{\textsf{labeled}})} + \underbrace{\sum_{j=1}^M \Omega(x_j^{(u)}; \theta)}_{\mathsf{unsupervised} },
\end{equation}
where $\Theta$ is the parameter space of the classifier, $\ell$ denotes the cross-entropy loss, and $\Omega(x_j^{(u)}; \theta)$ is a per-sample regularizer. In FixMatch\cite{sohn2020fixmatch}, $\Omega$ is instantiated as the per-sample consistency regularization:
\begin{equation} \label{eq:conloss}
    \mathcal{L}_{\textsf{con}}= \sum_{j=1}^M \; \underbrace{M(x_j^{(u)})}_{\mathsf{sample\ mask}} \; \underbrace{\ell\left(f(\mathcal{A} (x_j^{(u)})), q_j\right)}_{\mathsf{consistency}},
\end{equation}
where $q_j=\arg\max_c \delta(f(x_j^{(u)}))_c$ is the pseudo-label of $x_j^{(u)}$ predicted by $f$, and the sample mask $M(x_j^{(u)}) := \mathbb{I}\left(\max \left( \delta(f(x_j^{(u)})) \right) \geq \rho \right)$ selects unlabeled samples whose predicted confidence is higher than a predefined threshold $\rho$ ($\rho=0.95$ in FixMatch). Here, $\delta(\cdot)$ is the softmax function and $\mathbb{I}(\cdot)$ is the indicator function. To generate another view for each sample, $\mathcal{A}(x_j^{(u)})$ denotes the augmentation scheme applied to $x_j^{(u)}$, such as Cutout\cite{devries2017improved} and RandAugment\cite{cubuk2020randaugment}. Incorporating the consistency regularizer improves the model's robustness to spurious feature patterns.

Recent LTSSL methods are often inspired by long-tail learning strategies such as re-sampling \cite{wei2021crest} and logit adjustment \cite{menon2020long,wei2023towards}. However, methods that emphasize tail classes too strongly can hurt head-class performance, whereas standard SSL methods are usually biased toward head classes. In our LTSSL setting, we define head and tail classes using the full training set, including both labeled and unlabeled data, rather than using only labeled data as in traditional LTSSL work. Achieving strong performance on both head and tail classes at the same time is therefore challenging. To address this, we employ a dual-branch network with a \textbf{standard branch (denoted by $f$)} and a \textbf{balanced branch (denoted by $\widetilde{f}$)} to learn a standard classifier and a class-balanced classifier, which are designed to favor head and tail classes, respectively, during training. Both $f$ and $\widetilde{f}$ output logits, and their softmax probabilities are denoted by $\delta(f(\cdot))$ and $\delta(\widetilde{f}(\cdot))$, respectively.

\textit{Can the dual-branch network handle unknown class distributions of unlabeled data?} Our proposed method, \algo, addresses this challenge through a process of decoupling then converging.
\begin{itemize}
\setlength\itemsep{0.1em}
\item \textbf{Decoupling.} We first decouple the training objective by establishing a dual-branch network. A standard branch preserves performance on head classes, while a balanced branch is designed to improve performance on tail classes.
\item \textbf{Converging.} We then introduce interaction mechanisms that encourage the two branches to incrementally converge, sharing knowledge and eventually guaranteeing strong performance for all classes.
\end{itemize}


\subsection{Decoupling via Dual-Branch Training}
\label{sec:dual_learning}
To achieve the first process, we leverage supervision constraints on labeled data to guide two branches in focusing on head and tail classes, respectively.

For the balanced branch, we intentionally bias predictions toward minority classes to reduce the model's preference for head classes and improve tail-class performance. To do so, we optimize $\textit{balanced\ softmax}$\cite{ren2020balanced,menon2020long} on labeled data, which modifies the standard cross-entropy as follows:
\begin{equation}
    \mathcal{L}_{\textsf{b-labeled}} = - \sum_{i=1}^N \log \frac{e^{\widetilde{f}_{y_i^{(l)}}(x_i^{(l)})+\log \pi_{y_i^{(l)}}}}{\sum_{c=1}^C e^{\widetilde{f}_c(x_i^{(l)})+\log \pi_c}},
    \label{eq:bal_CE}
\end{equation}
%
where $\pi=(\pi_1,\dots,\pi_C)$ denotes the class-prior vector and $\pi_c=\mathbb{P}(y=c)$ is its $c$-th entry, which is approximated by the empirical frequency on the labeled samples. We can obtain more balanced predictions by minimizing \Cref{eq:bal_CE}.

For the standard branch, our goal is to produce more accurate pseudo-labels for head classes. Although standard cross-entropy is effective for head classes, it can increasingly bias the model during training, especially when the unlabeled distribution differs from the labeled one. This bias can lead to many incorrect pseudo-labels. To reduce it, we align the predictions of the standard branch to a decoupled distribution using accurate supervision from labeled data:
\begin{equation}
    \mathcal{L}_{\textsf{labeled}} = - \sum_{i=1}^N \log \frac{e^{f_{y_i^{(l)}}(x_i^{(l)})+\tau_{2} \cdot \Delta_{y_i^{(l)}}}}{\sum_{c=1}^C e^{f_c(x_i^{(l)})+\tau_{2} \cdot \Delta_c}},
    \label{eq:std_CE}
\end{equation}
where $\tau_{2}$ controls the alignment intensity of the standard branch, $\Delta_c = \log \pi_c - \log \pi^{b}_c$ is the adjustment term for class $c$, and $\pi^{b}=(\pi^{b}_1,\dots,\pi^{b}_C)$ denotes the class-prior vector estimated by the balanced branch. Here, $\pi^{b}$ is updated by an exponential moving average (EMA) as follows:
\begin{align}
    \pi^{b} &= m \pi^{b} + (1 - m) \widetilde{p},
    \label{eq:ema_est}
    \\
    \widetilde{p} &= \frac{1}{B + \mu B} (\sum_{i=1}^{B}\widetilde{p}_{i} + \sum_{j=1}^{\mu B}\widetilde{p}_{j}),
    \label{eq:cal_pb}
\end{align}
where $m\in[0, 1)$ is the EMA coefficient, $B$ is the labeled mini-batch size, and $\mu$ controls the unlabeled-to-labeled batch-size ratio, so the unlabeled mini-batch size is $\mu B$. The vector $\widetilde{p}$ on the left-hand side of \Cref{eq:cal_pb} is the mini-batch class-prior estimate produced by $\widetilde{f}$ over all $B+\mu B$ samples, where $\widetilde{p}_{i}=\delta(\widetilde{f}(x_i^{(l)}))$ and $\widetilde{p}_{j}=\delta(\widetilde{f}(x_j^{(u)}))$ denote the softmax prediction vectors of the $i$-th labeled sample and the $j$-th unlabeled sample, respectively. 

The adjustment based on $\pi_c$ in \Cref{eq:bal_CE} mitigates labeled-data imbalance, while $\pi^{b}_c$ helps $f$ align more closely with the decoupled distribution. Notably, the decoupled distribution in \Cref{eq:std_CE} is estimated by the balanced branch $\widetilde{f}$ rather than by $f$ itself, which is why we refer to $\pi^{b}_c$ as the decoupled distribution. We posit that this strategy can mitigate the confirmation bias often observed in SSL tasks \cite{arazo2020pseudo,wang2021self} and prevent the accumulation of distributional errors. In fact, if we align the model's tendency to the distribution estimated by $f$, it will gradually bias itself toward its own predictions, leading to accumulated preferences and incorrect pseudo-labels that ultimately hurt accuracy. Aligning to $\pi^{b}_c$ instead reduces reliance on $f$ alone and mitigates the risk of error accumulation. 

Despite \Cref{eq:std_CE} seeming to lack deliberate attention to head classes with the alignment to $\pi^{b}_c$, it is worth noting that the decoupled distribution portrays a subtle bias towards balanced distribution and more importantly, it also reflects the true distribution to a large degree, wherein predictions for head classes remain significantly more pronounced than those for tail classes. Consequently, the alignment in \Cref{eq:std_CE} can effectively contribute to enhancing the performance of head classes in standard branch.


\subsection{Converging via Cross-Branch Interaction}




Having established the two decoupled branches in \Cref{sec:dual_learning}, we now introduce the interaction mechanisms that drive their convergence. One key interaction already appears in the alignment to $\pi^{b}$ in \Cref{eq:std_CE}, through which the balanced branch influences the standard branch. This alignment also links labeled and unlabeled data, which helps unlock the model's performance potential \cite{huang2023flatmatch,huang2024interlude}.

We further apply a simple logit adjustment (a.k.a post-hoc logit adjustment\cite{menon2020long}) to the output of the standard classifier $f$ whose predicted distribution tends to favor the head classes. The refined logits are used to generate pseudo-labels, which will be treated as targets for balanced branch. In this way, \algo\ can mitigate the confirmation bias commonly observed in SSL methods. Specifically, the pseudo-label of the $j$-th unlabeled data $x_j^{(u)}$ used in the consistency regularizer in balanced branch is generated as:
\begin{equation}
    \widetilde{q}(x^{(u)}_j) = \arg\max_{c} \left[f_c\left(x_j^{(u)}\right) - \tau_{1} \cdot \log \pi^{s}_c\right],
    \label{eq:std2bal_pseu}
\end{equation}
where $\pi^{s}=(\pi^{s}_1,\dots,\pi^{s}_C)$ denotes the unlabeled class-prior vector estimated from the standard-branch predictions on unlabeled data, $\tau_{1}$ controls the adjustment intensity for balanced pseudo-label generation, and $\widetilde{q}_j=\widetilde{q}(x_j^{(u)})$ is the resulting pseudo-label for the $j$-th unlabeled sample. The consistency regularizer for the balanced branch is:
\begin{equation} \label{eq:b-conloss}
    \mathcal{L}_{\textsf{b-con}}= \sum_{j=1}^M \psi(x_j^{(u)}) \ell\left(\widetilde{f}(\mathcal{A} (x_j^{(u)})), \widetilde{q}_j\right),
\end{equation}
where $\psi(x_j^{(u)}) = \gamma_t \cdot \eta(x_j^{(u)})$ is the importance weight for the $j$-th unlabeled sample, $\eta(x_j^{(u)}) = \max_c [\delta(f(x_j^{(u)})) \odot \delta(f(x_j^{(u)}) - \tau_{1} \cdot \log \pi^{s})]_c$ measures the consistency between the predictions before and after adjustment, $\odot$ denotes element-wise multiplication, and $\log \pi^{s}$ is applied element-wise. The term $\gamma_t$ is an epoch-dependent scaling factor that scales the overall magnitude of the sample weights to align with the confidence of $f$ at epoch $t$:
\begin{equation}
    \gamma_t = \sum_{j=1}^{\mu B} \max \delta(f_t(x_j^{(u)})) \hspace{3pt} / \hspace{3pt} \sum_{j=1}^{\mu B} \eta(x_j^{(u)}).
    \label{eq:gamma_t}
\end{equation}
Notably, $f_t$ represents the standard branch at epoch $t$, indicating that $\gamma_t$ changes dynamically during training. By definition, $\eta(x_j^{(u)})$ measures the similarity between the standard-branch outputs before and after the adjustment in \Cref{eq:std2bal_pseu}. A large distribution shift during the adjustment produces a small $\eta(x_j^{(u)})$, indicating that the adjustment substantially changes the sample's prediction and that the adjusted prediction may be unreliable. Assigning a small weight to such a sample can therefore reduce its negative influence on training. In contrast, when $\eta(x_j^{(u)})$ is large, the prediction changes little after adjustment, suggesting that the sample is more reliable and should receive a larger weight. Further analysis of sample weights is provided in \Cref{exp:sys_analysis}.

The scaling factor $\gamma_t$ in \Cref{eq:gamma_t} attempts to maintain the overall magnitude of the weights and the model's confidence consistently. $\gamma_t$ guarantees that initially, due to insufficient training, the weights remain relatively low. As training progresses and the model becomes increasingly confident in its predictions, the corresponding weights should be concurrently elevated, which enhances the contribution of accurate pseudo-labels to the consistency regularization loss.

Through \Cref{eq:b-conloss}, the standard branch provides pseudo-labels and consistency weights to the balanced branch, further illustrating the interaction between the two branches. The predicted unlabeled distributions of the two branches tend to converge, as shown in \Cref{exp:sys_analysis} and \Cref{fig:kl_qhatbal}. More importantly, the converged balanced branch achieves high pseudo-label accuracy on both head and tail classes in \Cref{fig:pseu_acc}.


\begin{table*}[t]
\centering
\caption{Test accuracy of previous LTSSL algorithms and our proposed \algo\ under consistent class distributions, i.e., $\gamma_l = \gamma_u$, on CIFAR-10-LT and CIFAR-100-LT datasets. The best results are in \textbf{bold}.}
\resizebox{\textwidth}{!}{%
\begin{tabular}{@{}llccccccccccc@{}}
\toprule
 &
   &
  \multicolumn{5}{c}{CIFAR-10-LT} &
   &
  \multicolumn{5}{c}{CIFAR-100-LT} \\ \midrule
 &
   &
  \multicolumn{2}{c}{$\gamma=\gamma_l=\gamma_u=100$} &
   &
  \multicolumn{2}{c}{$\gamma=\gamma_l=\gamma_u=150$} &
   &
  \multicolumn{2}{c}{$\gamma=\gamma_l=\gamma_u=10$} &
   &
  \multicolumn{2}{c}{$\gamma=\gamma_l=\gamma_u=20$} \\ \cmidrule(lr){3-4} \cmidrule(lr){6-7} \cmidrule(lr){9-10} \cmidrule(l){12-13} 
Algorithm &
   &
  \begin{tabular}[c]{@{}c@{}}$N_1=500$\\ $M_1=4000$\end{tabular} &
  \begin{tabular}[c]{@{}c@{}}$N_1=1500$\\ $M_1=3000$\end{tabular} &
   &
  \begin{tabular}[c]{@{}c@{}}$N_1=500$\\ $M_1=4000$\end{tabular} &
  \begin{tabular}[c]{@{}c@{}}$N_1=1500$\\ $M_1=3000$\end{tabular} &
   &
  \begin{tabular}[c]{@{}c@{}}$N_1=50$\\ $M_1=400$\end{tabular} &
  \begin{tabular}[c]{@{}c@{}}$N_1=150$\\ $M_1=300$\end{tabular} &
   &
  \begin{tabular}[c]{@{}c@{}}$N_1=50$\\ $M_1=400$\end{tabular} &
  \begin{tabular}[c]{@{}c@{}}$N_1=150$\\ $M_1=300$\end{tabular} \\ \cmidrule(r){1-1} \cmidrule(lr){3-4} \cmidrule(lr){6-7} \cmidrule(lr){9-10} \cmidrule(l){12-13} 
\begin{tabular}[c]{@{}l@{}}Supervised\\ \quad w/LA\cite{menon2020long}\end{tabular} &
   &
  \begin{tabular}[c]{@{}c@{}}\ms{47.3}{0.95}\\ \ms{53.3}{0.44}\end{tabular} &
  \begin{tabular}[c]{@{}c@{}}\ms{61.9}{0.41}\\ \ms{70.6}{0.21}\end{tabular} &
   &
  \begin{tabular}[c]{@{}c@{}}\ms{44.2}{0.33}\\ \ms{49.5}{0.40}\end{tabular} &
  \begin{tabular}[c]{@{}c@{}}\ms{58.2}{0.29}\\ \ms{67.1}{0.78}\end{tabular} &
   &
  \begin{tabular}[c]{@{}c@{}}\ms{29.6}{0.57}\\ \ms{30.2}{0.44}\end{tabular} &
  \begin{tabular}[c]{@{}c@{}}\ms{46.9}{0.22}\\ \ms{48.7}{0.89}\end{tabular} &
   &
  \begin{tabular}[c]{@{}c@{}}\ms{25.1}{1.14}\\ \ms{26.5}{1.31}\end{tabular} &
  \begin{tabular}[c]{@{}c@{}}\ms{41.2}{0.15}\\ \ms{44.1}{0.42}\end{tabular} \\ \cmidrule(r){1-1} \cmidrule(lr){3-4} \cmidrule(lr){6-7} \cmidrule(lr){9-10} \cmidrule(l){12-13} 
\begin{tabular}[c]{@{}l@{}}FixMatch \cite{sohn2020fixmatch}\\ \quad w/ DARP \cite{kim2020distribution}\\ \quad w/ CReST+ \cite{wei2021crest}\\ \quad w/ DASO \cite{oh2022daso} \\ \quad w/ SimPro \cite{du2024simpro}\\ \quad w/ ACR\cite{wei2023towards}\\ \quad w/ CDMAD\cite{CDMAD} \\ \quad w/ FARAD\cite{gufourier}\end{tabular} &
   &
  \begin{tabular}[c]{@{}c@{}}\ms{67.8}{1.13}\\ \ms{74.5}{0.78}\\ \ms{76.3}{0.86}\\ \ms{76.0}{0.37}\\ \ms{80.7}{0.30}\\ \ms{81.6}{0.19}\\ - \\ - \end{tabular} &
  \begin{tabular}[c]{@{}c@{}}\ms{77.5}{1.32}\\ \ms{77.8}{0.63}\\ \ms{78.1}{0.42}\\ \ms{79.1}{0.75}\\ - \\ \ms{84.1}{0.39}\\ \ms{83.6}{0.46} \\ \ms{85.6}{0.38}\end{tabular} &
   &
  \begin{tabular}[c]{@{}c@{}}\ms{62.9}{0.36}\\ \ms{67.2}{0.32}\\ \ms{67.5}{0.45}\\ \ms{70.1}{1.81}\\ \ms{74.2}{0.90}\\ \ms{77.0}{1.19}\\ - \\ - \end{tabular} &
  \begin{tabular}[c]{@{}c@{}}\ms{72.4}{1.03}\\ \ms{73.6}{0.73}\\ \ms{73.7}{0.34}\\ \ms{75.1}{0.77}\\ - \\ \ms{80.9}{0.22}\\ \ms{80.8}{0.86}\\ \ms{82.5}{0.64}\end{tabular} &
   &
  \begin{tabular}[c]{@{}c@{}}\ms{45.2}{0.55}\\ \ms{49.4}{0.20}\\ \ms{44.5}{0.94}\\ \ms{49.8}{0.24}\\ - \\ \ms{51.3}{0.48}\\ - \\ -\end{tabular} &
  \begin{tabular}[c]{@{}c@{}}\ms{56.5}{0.06}\\ \ms{58.1}{0.44}\\ \ms{57.4}{0.18}\\ \ms{49.2}{0.35}\\ - \\ \ms{61.1}{0.11}\\ \ms{61.0}{0.24} \\\ms{60.2}{0.32}\end{tabular} &
   &
  \begin{tabular}[c]{@{}c@{}}\ms{40.0}{0.96}\\ \ms{43.4}{0.87}\\ \ms{40.1}{1.28}\\ \ms{43.6}{0.09}\\ \ms{43.1}{0.40}\\ \ms{44.8}{0.21}\\ - \\ -\end{tabular} &
  \begin{tabular}[c]{@{}c@{}}\ms{50.7}{0.25}\\ \ms{52.2}{0.66}\\ \ms{52.1}{0.21}\\ \ms{52.9}{0.42}\\ - \\ \ms{55.9}{0.31} \\ \ms{54.3}{0.44}\\ \ms{55.9}{0.34}\end{tabular} \\ \cmidrule(r){1-1} \cmidrule(lr){3-4} \cmidrule(lr){6-7} \cmidrule(lr){9-10} \cmidrule(l){12-13}
FixMatch w/ Ours &
   &
  \ms{\textbf{84.1}}{0.56} &
  \ms{\textbf{86.7}}{0.54} &
   &
  \ms{\textbf{81.1}}{0.92} &
  \ms{\textbf{84.1}}{0.21} &
   &
  \ms{\textbf{52.0}}{0.25} &
  \ms{\textbf{62.1}}{0.35} &
   &
  \ms{\textbf{45.5}}{0.33} &
  \ms{\textbf{56.6}}{0.49} \\ \bottomrule
\end{tabular}%
}
\label{tab:main}
\end{table*}

\begin{table*}[ht]
\centering
\caption{Test accuracy of previous LTSSL algorithms and our proposed \algo\ under inconsistent class distributions, i.e., $\gamma_l \not= \gamma_u$, on CIFAR-10-LT and STL10-LT datasets. The $\gamma_l$ is fixed to 100 for CIFAR-10-LT, while it is set to 10 and 20 for STL10-LT dataset. The best results are in \textbf{bold}.}
\resizebox{\textwidth}{!}{%
\begin{tabular}{@{}llcclcclccccc@{}}
\toprule
 &
   &
  \multicolumn{5}{c}{CIFAR-10-LT ($\gamma_l \neq \gamma_u$)} &
   &
  \multicolumn{5}{c}{STL10-LT ($\gamma_u = N/A$)} \\ \midrule
 &
   &
  \multicolumn{2}{c}{$\gamma_u=1$ (uniform)} &
   &
  \multicolumn{2}{c}{$\gamma_u=1/100$ (reversed)} &
   &
  \multicolumn{2}{c}{$\gamma_l=10$} &
  \multicolumn{1}{l}{} &
  \multicolumn{2}{c}{$\gamma_l=20$} \\ \cmidrule(lr){3-4} \cmidrule(lr){6-7} \cmidrule(lr){9-10} \cmidrule(l){12-13} 
Algorithm &
   &
  \begin{tabular}[c]{@{}c@{}}$N_1=500$\\ $M_1=4000$\end{tabular} &
  \begin{tabular}[c]{@{}c@{}}$N_1=1500$\\ $M_1=3000$\end{tabular} &
   &
  \begin{tabular}[c]{@{}c@{}}$N_1=500$\\ $M_1=4000$\end{tabular} &
  \begin{tabular}[c]{@{}c@{}}$N_1=1500$\\ $M_1=3000$\end{tabular} &
   &
  \begin{tabular}[c]{@{}c@{}}$N_1=150$\\ $M_1=100k$\end{tabular} &
  \begin{tabular}[c]{@{}c@{}}$N_1=450$\\ $M_1=100k$\end{tabular} &
   &
  \begin{tabular}[c]{@{}c@{}}$N_1=150$\\ $M_1=100k$\end{tabular} &
  \begin{tabular}[c]{@{}c@{}}$N_1=450$\\ $M_1=100k$\end{tabular} \\ \cmidrule(r){1-1} \cmidrule(lr){3-4} \cmidrule(lr){6-7} \cmidrule(lr){9-10} \cmidrule(l){12-13} 
\begin{tabular}[c]{@{}l@{}}FixMatch \cite{sohn2020fixmatch}\\ \quad w/ DARP \cite{kim2020distribution}\\ \quad w/ CReST \cite{wei2021crest}\\ \quad w/ CReST+ \cite{wei2021crest}\\ \quad w/ DASO \cite{oh2022daso}\\ \quad w/ SimPro \cite{du2024simpro}\\ \quad w/ ACR \cite{wei2023towards}\\ \quad w/ CDMAD\cite{CDMAD}\\ \quad w/ FARAD\cite{gufourier}\\ \quad w/ Ours\end{tabular} &
   &
  \begin{tabular}[c]{@{}c@{}}\ms{73.0}{3.81}\\ \ms{82.5}{0.75}\\ \ms{83.2}{1.67}\\ \ms{82.2}{1.53}\\ \ms{86.6}{0.84}\\ \ms{93.8}{0.10}\\ \ms{92.1}{0.18}\\ - \\ - \\ \ms{\textbf{93.9}}{0.02}\end{tabular} &
  \begin{tabular}[c]{@{}c@{}}\ms{81.5}{1.15}\\ \ms{84.6}{0.34}\\ \ms{87.1}{0.28}\\ \ms{86.4}{0.42}\\ \ms{88.8}{0.59}\\ - \\ \ms{93.5}{0.11}\\ \ms{87.5}{0.46}\\ \ms{91.5}{0.64}\\ \ms{\textbf{94.1}}{0.08}\end{tabular} &
  \multicolumn{1}{c}{} &
  \begin{tabular}[c]{@{}c@{}}\ms{62.5}{0.94}\\ \ms{70.1}{0.22}\\ \ms{70.7}{2.02}\\ \ms{62.9}{1.39}\\ \ms{71.0}{0.95}\\ \ms{85.8}{0.48}\\ \ms{84.9}{0.09}\\ - \\ - \\ \ms{\textbf{86.6}}{0.33}\end{tabular} &
  \begin{tabular}[c]{@{}c@{}}\ms{71.8}{1.70}\\ \ms{80.0}{0.93}\\ \ms{80.8}{0.39}\\ \ms{72.9}{2.00}\\ \ms{80.3}{0.65}\\ - \\ \ms{89.5}{0.17}\\ \ms{86.3}{0.54}\\ \ms{87.6}{0.45}\\ \ms{\textbf{89.9}}{0.21}\end{tabular} &
  \multicolumn{1}{c}{} &
  \begin{tabular}[c]{@{}c@{}}\ms{56.1}{2.32}\\ \ms{66.9}{1.66}\\ \ms{61.7}{2.51}\\ \ms{61.2}{1.27}\\ \ms{70.0}{1.19}\\ - \\ \ms{77.1}{0.24}\\ - \\ - \\ \ms{\textbf{83.3}}{0.46}\end{tabular} &
  \begin{tabular}[c]{@{}c@{}}\ms{72.4}{0.71}\\ \ms{75.6}{0.45}\\ \ms{71.6}{1.17}\\ \ms{71.5}{0.96}\\ \ms{78.4}{0.80}\\ \ms{84.5}{0.39}\\ \ms{83.0}{0.32}\\ \ms{79.9}{0.23}\\ \ms{84.6}{0.43}\\ \ms{\textbf{86.6}}{0.15}\end{tabular} &
   &
  \begin{tabular}[c]{@{}c@{}}\ms{47.6}{4.87}\\ \ms{59.9}{2.17}\\ \ms{57.1}{3.67}\\ \ms{56.0}{3.19}\\ \ms{65.7}{1.78}\\ - \\ \ms{75.1}{0.70}\\ - \\ - \\ \ms{\textbf{82.1}}{0.52}\end{tabular} &
  \begin{tabular}[c]{@{}c@{}}\ms{64.0}{2.27}\\ \ms{72.3}{0.60}\\ \ms{68.6}{0.88}\\ \ms{68.5}{1.88}\\ \ms{75.3}{0.44}\\ \ms{82.5}{0.25}\\ \ms{81.5}{0.25}\\ \ms{75.2}{0.40}\\ \ms{83.2}{0.52} \\ \ms{\textbf{85.7}}{0.53}\end{tabular} \\ \bottomrule
\end{tabular}%
}
\label{tab:cifar10_stl10}
\end{table*}

\begin{table}[h]
\centering
\caption{Test accuracy on CIFAR-100-LT dataset under \textit{uniform} and \textit{reversed} class distributions. The best results are in \textbf{bold}.}
\begin{small}
\begin{sc}
\resizebox{1.0\linewidth}{!}{%
\begin{tabular}{@{}llcclcc@{}}
\toprule
 &
   &
  \multicolumn{5}{c}{CIFAR-100-LT ($\gamma_l \neq \gamma_u$)} \\ \midrule
 &
   &
  \multicolumn{2}{c}{$\gamma_u=1$ (uniform)} &
   &
  \multicolumn{2}{c}{$\gamma_u=1/10$ (reversed)} \\ \cmidrule(lr){3-4} \cmidrule(l){6-7} 
Algorithm &
   &
  \begin{tabular}[c]{@{}c@{}}$N_1=50$\\ $M_1=400$\end{tabular} &
  \begin{tabular}[c]{@{}c@{}}$N_1=150$\\ $M_1=300$\end{tabular} &
   &
  \begin{tabular}[c]{@{}c@{}}$N_1=50$\\ $M_1=400$\end{tabular} &
  \begin{tabular}[c]{@{}c@{}}$N_1=150$\\ $M_1=300$\end{tabular} \\ \cmidrule(r){1-1} \cmidrule(lr){3-4} \cmidrule(l){6-7} 
\begin{tabular}[c]{@{}l@{}}FixMatch \cite{sohn2020fixmatch}\\ \quad w/ DARP \cite{kim2020distribution}\\ \quad w/ CReST \cite{wei2021crest}\\ \quad w/ CReST+ \cite{wei2021crest}\\ \quad w/ DASO \cite{oh2022daso}\\ \quad w/ ACR \cite{wei2023towards}\\ \quad w/ Ours\end{tabular} &
   &
  \begin{tabular}[c]{@{}c@{}}\ms{45.5}{0.71}\\ \ms{43.5}{0.95}\\ \ms{43.5}{0.30}\\ \ms{43.6}{1.60}\\ \ms{53.9}{0.66}\\ \ms{57.2}{0.19}\\ \ms{\textbf{60.4}}{0.72}\end{tabular} &
  \begin{tabular}[c]{@{}c@{}}\ms{58.1}{0.72}\\ \ms{55.9}{0.32}\\ \ms{59.2}{0.25}\\ \ms{58.7}{0.16}\\ \ms{61.8}{0.98}\\ \ms{66.7}{0.30}\\ \ms{\textbf{68.0}}{0.21}\end{tabular} &
  \multicolumn{1}{c}{} &
  \begin{tabular}[c]{@{}c@{}}\ms{44.2}{0.43}\\ \ms{36.9}{0.48}\\ \ms{39.0}{1.11}\\ \ms{39.1}{0.77}\\ \ms{51.0}{0.19}\\ \ms{51.6}{0.12}\\ \ms{\textbf{53.3}}{0.20}\end{tabular} &
  \begin{tabular}[c]{@{}c@{}}\ms{57.3}{0.19}\\ \ms{51.8}{0.92}\\ \ms{56.4}{0.62}\\ \ms{56.4}{0.78}\\ \ms{60.0}{0.31}\\ \ms{62.9}{0.25}\\ \ms{\textbf{63.5}}{0.22}\end{tabular} \\ \bottomrule
\end{tabular}%
}
\end{sc}
\end{small}
\label{tab:cifar100_uni_rev}
\end{table}

\Cref{alg:alg1} summarizes the overall framework. The consistency regularizer for the standard branch is similar to FixMatch presented in \Cref{eq:conloss}. In general, each branch of the network has two losses to minimize, that is, the classification loss and the consistency regularizer. Put it together, our total objective function is:
\begin{equation}
\begin{split}
    \mathcal{L}_{\textsf{total}} = \underbrace{\mathcal{L}_{\textsf{labeled}} + \mathcal{L}_{\textsf{con}}}_{\textsf{standard branch}} + \underbrace{\mathcal{L}_{\textsf{b-labeled}} + \mathcal{L}_{\textsf{b-con}}}_{\textsf{balanced branch}}.
    \label{eq:total_loss}
\end{split}
\end{equation}

The backbone parameters are updated by gradients from both branches and each branch is updated only by its corresponding losses. A schematic overview of the full workflow and cross-branch interactions is provided in Supplementary Figure S5 (Supplementary Sec. S5). We also provide theoretical analysis on the stability of the two-branch interaction and the behavior of EMA-based prior signals under label shift in Supplementary Sec. S7.


\subsection{Closing the Distribution Gap during Inference}


Although the balanced branch $\widetilde{f}$ improves pseudo-label quality for both head and tail classes on unlabeled data, a distribution gap can still remain between the training data and the test data, which can substantially affect inference performance.


Consequently, we apply post-hoc logit adjustment \cite{menon2020long} to the balanced-branch outputs to further reduce prediction bias:
\begin{equation}
    \widetilde{q}_{\mathrm{test}}(x) = \arg\max_{c} \left[\widetilde{f}_c\left(x\right) - \tau_{3} \cdot \log \pi^{b}_c\right],
    \label{eq:post_hoc}
\end{equation}
where $\tau_{3}$ controls the intensity of post-hoc logit adjustment and $\pi^{b}$ is the balanced-branch prior estimate introduced in \Cref{eq:std_CE}. We use $\widetilde{q}_{\mathrm{test}}(x)$ as the final prediction for test sample $x$.

Although post-hoc adjustment is common in LTSSL, we find that it yields especially large gains for \algo. We attribute this to its higher pseudo-label accuracy during training, which is beneficial for the model’s representation learning \cite{kang2019decoupling} and makes the model more re-balanceable, that is, more amenable to further improvement through post-hoc balancing. It is evident that models with well-learned representations are more amenable to performance improvements on balanced test distributions through post-hoc re-balancing \cite{kang2019decoupling}. It is worth emphasizing that we do not assume a balanced test distribution; the test set may be either balanced or long-tailed. Because the classifier is trained on long-tailed data, it can still retain head-class bias at test time. The post-hoc logit adjustment in \Cref{eq:post_hoc} is therefore used to reduce this \textit{residual bias}, rather than to impose a balanced-test assumption. Given imbalanced unlabeled data, it is difficult to achieve both high pseudo-label accuracy during training and strong final test performance. We therefore argue that an effective LTSSL method should not only generate accurate pseudo-labels during training, but also remain amenable to substantial gains from post-hoc adjustment at inference. More discussion is provided in \Cref{exp:sys_analysis}.




\begin{algorithm}[h]
\caption{The Proposed \algo.}\label{alg:alg1}
\begin{algorithmic}[1]
\STATE {\bfseries Input:} labeled dataset $\mathcal{D}^{l}$ and unlabeled dataset $\mathcal{D}^{u}$, standard branch $f$, balanced branch $\widetilde{f}$, number of iterations $T$
\FOR{$t=1$ {\bfseries to} $T$}
\STATE $\{(x_i^{(l)},y_i^{(l)})\}^{B-1}_{i=0} \leftarrow$ Sample a batch of labeled data
\STATE $\{x_j^{(u)}\}^{\mu B-1}_{j=0} \leftarrow$ Sample a batch of unlabeled data
\STATE Update estimated class distribution $\pi^{b}$ by \Cref{eq:ema_est}
\STATE Generate the pseudo-labels $\widetilde{q}(x^{(u)}_j)$ for balanced branch
\STATE Obtain samples' weights via $\gamma_t$ and $\eta(x_j^{(u)})$
\STATE Compute the $\mathcal{L}_{\textsf{labeled}}$ by \Cref{eq:std_CE}
\STATE Compute the $\mathcal{L}_{\textsf{b-labeled}}$ by \Cref{eq:bal_CE}
\STATE Compute the $\mathcal{L}_{\textsf{con}}$ by \Cref{eq:conloss}
\STATE Compute the $\mathcal{L}_{\textsf{b-con}}$ by \Cref{eq:b-conloss}
\STATE Compute the total loss $\mathcal{L}_{\textsf{total}}$ by \Cref{eq:total_loss}
\STATE Update $f$ and $\widetilde{f}$ based on $\nabla\mathcal{L}_{\textsf{total}}$ using SGD
\ENDFOR
\end{algorithmic}
\label{alg1}
\end{algorithm}

\subsection{Distinctions between \algo\ and ACR}

Although \algo\ and ACR both use a dual-branch structure and logit adjustment to address imbalance in LTSSL, \algo\ includes several key improvements that make it applicable to a broader range of settings and lead to better performance.

The main distinctions between ACR and \algo\ are as follows:
\begin{itemize}
\setlength\itemsep{0.1em}
\item \textbf{Distribution Assumption.} ACR employs three typical anchor distributions to tackle diverse unlabeled class distributions, limiting its performance to \textit{consistent}, \textit{uniform}, and \textit{reversed} scenarios. However, \algo\ facilitates the dynamic convergence of the predictive distributions of the two branches without any assumptions about the unlabeled distribution.
\item \textbf{Adjustment strategy.} In order to alleviate biases introduced by the imbalanced distribution, ACR performs logit adjustment on predictions of unlabeled data according to the distribution of labeled data, which differs significantly from the real distribution of unlabeled data, leading to poor performance. Conversely, \algo\ adjusts model predictions by the estimated distribution from $f$, which can dynamically reflect the model's inclination as training progresses, thus enhancing the versatility and performance.
\item \textbf{Selection of pseudo-labels.} ACR only uses a fixed threshold to filter out less confident pseudo-labels, while \algo\ weights the consistency loss of $\widetilde{f}$ based on adjustments of outputs from $f$, mitigating the impact of erroneous pseudo-labels with excessive tail shift on training during logit adjustment.
\item \textbf{Further improvement during inference.} \algo\ employs post-hoc adjustment to further boost performance, whereas ACR, due to over-balancing during training, fails to gain considerable performance enhancement through post-hoc adjustment. More details can be found in \Cref{exp:sys_analysis}.
\end{itemize}

A more detailed clarification of how \algo\ differs from prior two-branch designs beyond ACR is provided in Supplementary Sec. S4.



\begin{figure*}[h]
\centering
\subfloat{\includegraphics[width=2.35in]{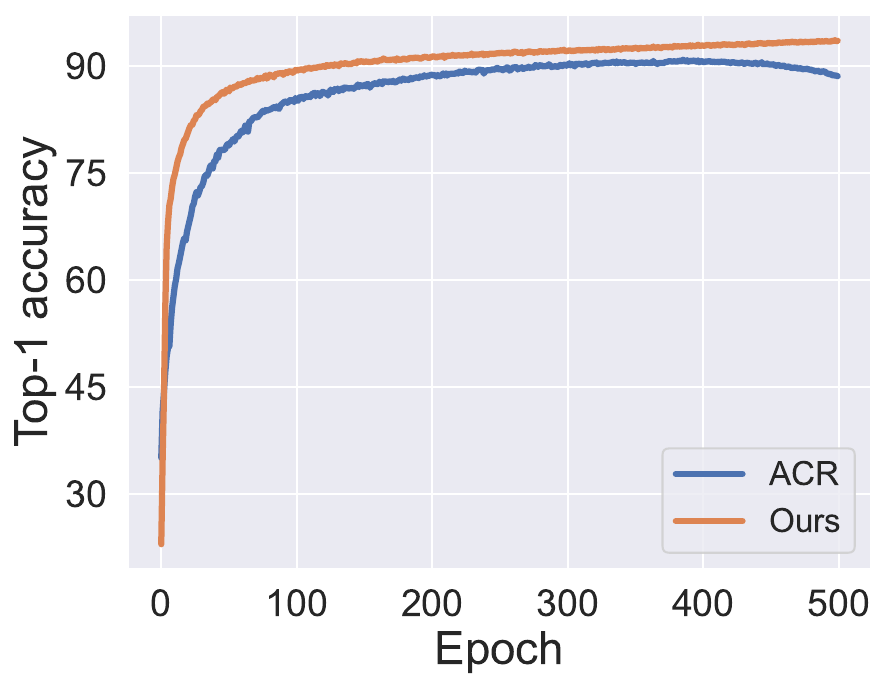}\label{fig:pseu_acc}}
\hfil
\subfloat{\includegraphics[width=2.35in]{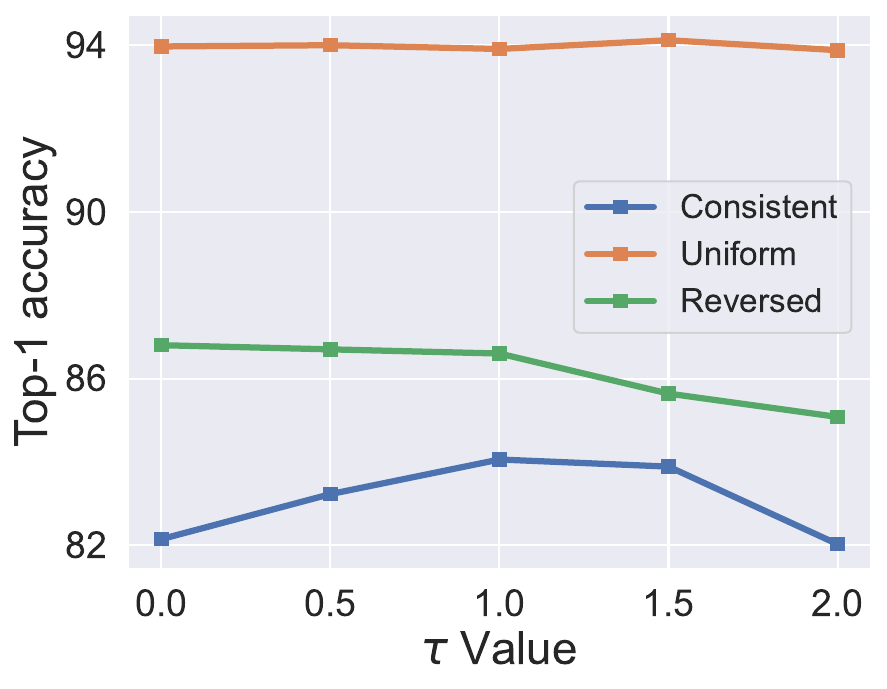}\label{fig:sen_tau}}
\hfil
\subfloat{\includegraphics[width=2.35in]{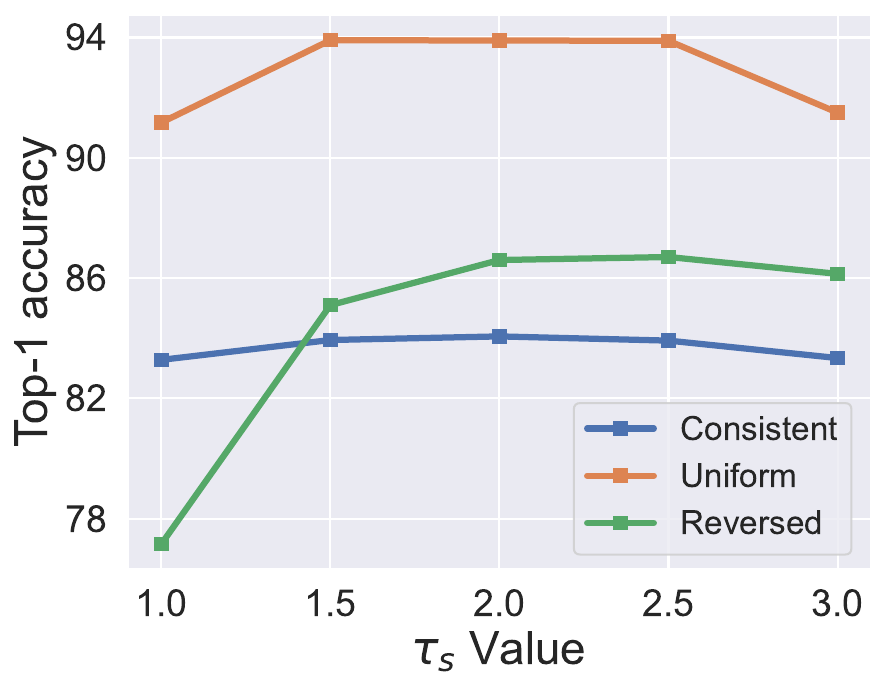}\label{fig:sen_taus}}
\caption{(\labelcref{fig:pseu_acc}): Pseudo-label accuracy for unlabeled data. The reported results are based on the average accuracy of \textit{consistent}, \textit{uniform}, and \textit{reversed} with $\gamma_l=100$ on CIFAR-10-LT. (\labelcref{fig:sen_tau,fig:sen_taus}): The sensitivity of $\tau_{1}$ and $\tau_{2}$ under various settings on CIFAR-10-LT.}
\label{fig:sen_analysis}
\end{figure*}

\begin{figure*}[h]
\centering
\subfloat[KL distance between branches]{\includegraphics[width=2.35in,height=1.83in]{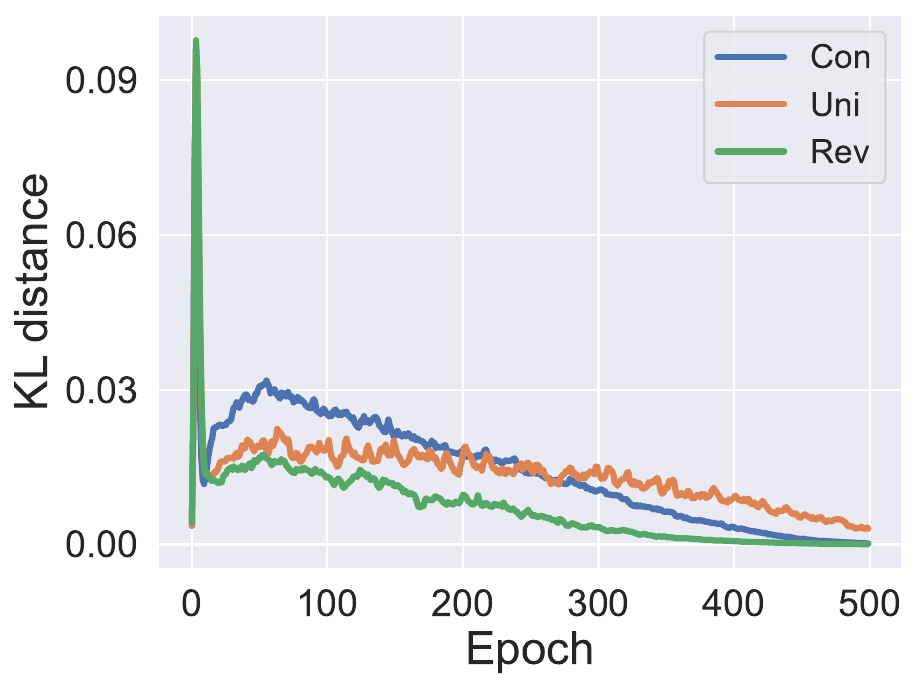}%
\label{fig:kl_qhatbal}}
\hfil
\subfloat[Head classes]{\includegraphics[width=2.35in,height=1.83in]{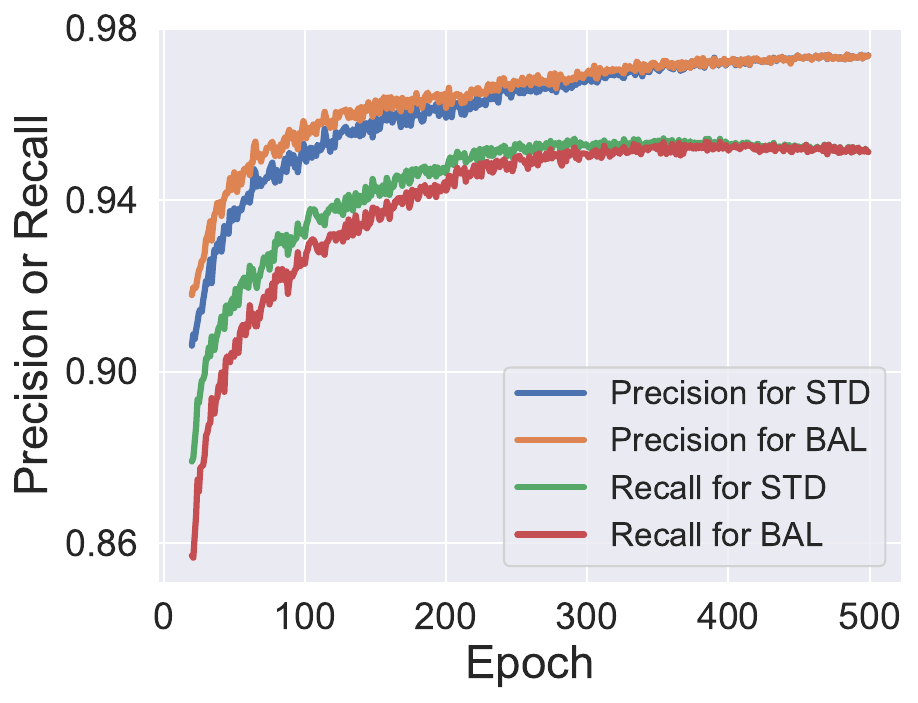}%
\label{fig:cifar10_head}}
\hfil
\subfloat[Tail classes]{\includegraphics[width=2.35in,height=1.83in]{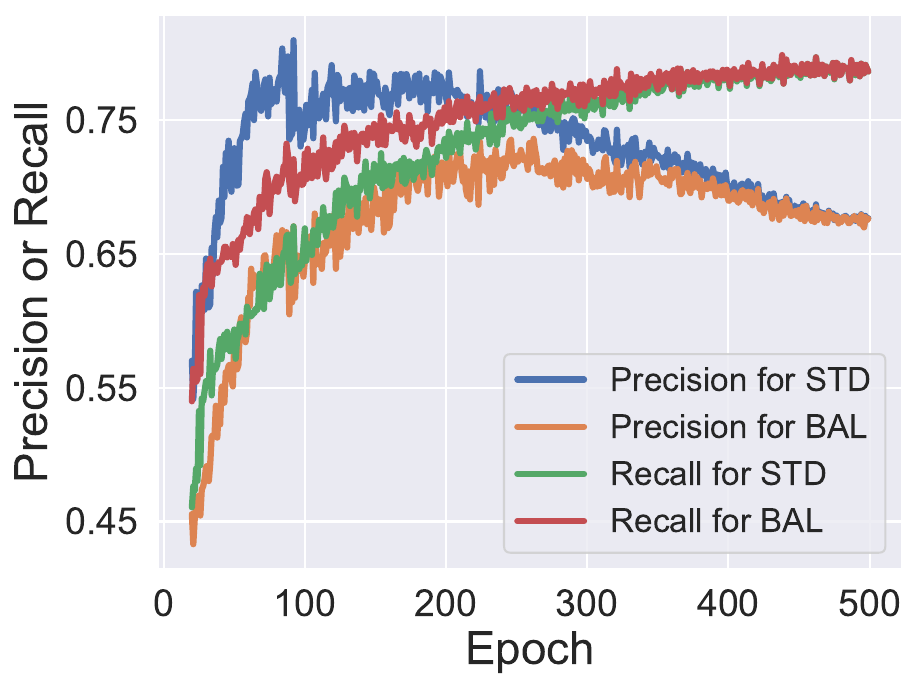}%
\label{fig:cifar10_tail}}
\caption{(\labelcref{fig:kl_qhatbal}): The KL distance of predicted unlabeled data distribution between standard and balanced branch. The experiments are conducted on CIFAR-100-LT with $N_1=50, M_1=400,$ and $\gamma_l=10$. \textit{Con}, \textit{Uni}, and \textit{Rev} represent \textit{consistent}, \textit{uniform}, and \textit{reversed} for short. (\labelcref{fig:cifar10_head,fig:cifar10_tail}): The average precision and recall across different settings for head and tail classes. The experiments are conducted on CIFAR-10-LT with $N_1=500, M_1=4000,$ and $\gamma_l=100$. \textit{STD} and \textit{BAL} represent standard branch and balanced branch, respectively.}
\label{fig:converge}
\end{figure*}

\begin{table}[h]
\centering
\small
\caption{Test accuracy on ImageNet-127. The best results are in \textbf{bold}.}
\resizebox{0.8\linewidth}{!}{%
\begin{tabular}{@{}llcc@{}}
\toprule
Algorithm &
   &
  $32 \times 32$ &
  $64 \times 64$ \\ \cmidrule(r){1-1} \cmidrule(l){3-4} 
\begin{tabular}[c]{@{}l@{}}FixMatch \cite{sohn2020fixmatch}\\ \quad w/ DARP \cite{kim2020distribution}\\ \quad w/ DARP+cRT \cite{kim2020distribution}\\ \quad w/ CReST+ \cite{wei2021crest}\\ \quad w/ CReST++LA \cite{menon2020long}\\ \quad w/ CoSSL \cite{fan2022cossl}\\ \quad w/ SimPro \cite{du2024simpro}\\ \quad w/ ACR \cite{wei2023towards}\\ \quad w/ CDMAD\cite{CDMAD}\\ \quad w/ FARAD\cite{gufourier}\\ \quad w/ Ours\end{tabular} &
   &
  \begin{tabular}[c]{@{}c@{}}29.7\\ 30.5\\ 39.7\\ 32.5\\ 40.9\\ 43.7\\ 59.1 \\ 57.2\\ 48.4\\ 50.6\\ \textbf{60.5}\end{tabular} &
  \begin{tabular}[c]{@{}c@{}}42.3\\ 42.5\\ 51.0\\ 44.7\\ 55.9\\ 53.9\\ 67.0 \\ 63.6\\ 59.3\\ 62.1\\ \textbf{68.4}\end{tabular} \\ \bottomrule
\end{tabular}%
}
\label{tab:small_imagenet}
\end{table}

\section{Experiments}

We conduct extensive experiments to demonstrate the effectiveness of the proposed \algo\ under various class distributions of unlabeled data.

\subsection{Experimental Setup}
The experiments we conducted are based on widely used datasets, including CIFAR-10-LT\cite{krizhevsky2009learning}, CIFAR-100-LT\cite{krizhevsky2009learning}, STL10-LT\cite{coates2011analysis}, and ImageNet-127\cite{fan2022cossl}. 
Recall that parameter $\gamma_l$ is used to control the imbalance ratio of the labeled dataset, and we can decide the number of labeled samples for class $c$ as $N_c = N_1 \cdot \gamma_l^{-\frac{c-1}{C-1}}$ once $N_1$ is given. Likewise, given the imbalance ratio of unlabeled dataset $\gamma_u$ and $M_1$ (or $M_C$ in the \textit{reversed} setting), we set $M_c$ as we did for the labeled dataset.

\begin{itemize}
    \item \textbf{CIFAR-10-LT}: Following DASO\cite{oh2022daso}, we test our method under $N_1=500, M_1=4000$ and $N_1=1500, M_1=3000$ settings. We report results with imbalance ratios $\gamma_l=\gamma_u=100$ and $\gamma_l=\gamma_u=150$. For \textit{uniform} and \textit{reversed} settings, we fix $\gamma_l=100$ and adjust $\gamma_u\in\{1, 1/100\}$ to simulate various class distribution of unlabeled data.
    \item \textbf{CIFAR-100-LT}: We test our method under $N_1=50, M_1=400$ and $N_1=150, M_1=300$ settings. The imbalance ratio is set to $\gamma_l=\gamma_u=10$ and $\gamma_l=\gamma_u=20$. With a fixed $\gamma_l=10$, we also test our method under $\gamma_u\in\{1,1/10\}$ for the \textit{uniform} and \textit{reversed} unlabeled data class distributions.
    \item \textbf{STL10-LT}: Since ground-truth labels of unlabeled data in STL-10 are unknown, we conduct experiments by controlling the imbalance ratio of labeled data. We set $\gamma_l\in\{10,20\}$.
    \item \textbf{ImageNet-127}: ImageNet-127 is a naturally long-tailed dataset, so we do not need to construct the datasets manually. Following CoSSL \cite{fan2022cossl}, we down-sample the image size to $32 \times 32$ and $64 \times 64$ due to limited resources. 
\end{itemize}


Following prior work\cite{oliver2018realistic,fan2022cossl}, we implement our method with Wide ResNet-28-2\cite{zagoruyko2016wide} on CIFAR-10-LT, CIFAR-100-LT, and STL10-LT, and with ResNet-50 on ImageNet-127. Following FixMatch, we train for 500 epochs with 500 mini-batches per epoch and a batch size of 64 using SGD with momentum\cite{sutskever2013importance,polyak1964some,nesterov27method}. We use cosine learning rate decay\cite{loshchilov2016sgdr}, which sets the learning rate to $\eta_0 \cos(\frac{7\pi t}{16T})$, where $\eta_0$ is the initial learning rate, $t$ is the current training step, and $T$ is the total number of training steps. We use $\eta_0$ here to avoid confusion with the sample-consistency score $\eta(x_j^{(u)})$ in \Cref{eq:b-conloss}. We set $\tau_{2}=2.0$ for all datasets, as defined in \Cref{eq:std_CE}. For $\tau_{1}$, we use $1.0$ by default and $2.0$ on CIFAR-100-LT. For post-hoc enhancement, we set $\tau_{3}=2.0$ on CIFAR-10-LT and $\tau_{3}=1.0$ on the remaining datasets. The EMA coefficient $m$ is set to $0.99$ by default. We compare \algo\ with existing LTSSL algorithms, including DARP\cite{kim2020distribution}, CReST\cite{wei2021crest}, DASO\cite{oh2022daso}, ABC\cite{lee2021abc}, TRAS\cite{wei2022transfer}, SimPro\cite{du2024simpro}, and ACR\cite{wei2023towards}. Performance is measured by top-1 test accuracy, and we report the mean and standard deviation over three independent runs. Following \cite{sohn2020fixmatch}, we report the final performance of the balanced branch $\widetilde{f}$ using an exponential moving average of model parameters, since it achieves superior pseudo-label accuracy in \Cref{fig:pseu_acc}. All experiments can be conducted in PyTorch on a single NVIDIA V100 32GB GPU.

\subsection{Results on CIFAR-10/100-LT and STL10-LT}
We first evaluate the performance when the class distributions are \textit{consistent} (i.e., $\gamma_l = \gamma_u$) in \Cref{tab:main}. Subsequently, in \Cref{tab:cifar10_stl10} and \Cref{tab:cifar100_uni_rev}, we report results when the unlabeled data class distribution is \textit{uniform} or \textit{reversed} (e.g., $\gamma_u=1$ or $\gamma_u=1/100$). It is noteworthy that the distribution of unlabeled data in STL10-LT is completely unknown.

\noindent
\textbf{Case of} $\gamma_l = \gamma_u$. We compare our approach \algo\ with several state-of-the-art LTSSL methods: DARP\cite{kim2020distribution}, CReST+\cite{wei2021crest}, DASO\cite{oh2022daso}, SimPro\cite{du2024simpro}, and ACR\cite{wei2023towards}. Results are reported in \Cref{tab:main}. Without exception, \algo\ consistently outperforms existing methods by a large margin, even though some of these methods are particularly developed based on the assumption that labeled and unlabeled data share the same class distribution. This observation verifies the superior performance of our method. Specifically, \algo\ exhibits a 1.9\% average advantage over ACR, particularly excelling on the CIFAR-10-LT where the average advantage reaches 3.1\%.

\noindent
\textbf{Case of} $\gamma_l \not= \gamma_u$. In real-world datasets, the class distribution of unlabeled data is likely to be significantly inconsistent with labeled data. Therefore, we consider \textit{uniform} and \textit{reversed} class distributions, e.g., $\gamma_u=1$ or $\gamma_u=1/100$ for CIFAR-10-LT. On the STL10-LT dataset, due to the unknown ground-truth labels of the unlabeled data, we can only control the imbalance ratio of labeled data. The results are summarized in the \Cref{tab:cifar10_stl10} and \Cref{tab:cifar100_uni_rev}.

It can be seen that \algo\ achieves the best results when the class distributions of unlabeled data are inconsistent. For example, \algo\ obtains  16.8\% and 21.1\% absolute performance gains over FixMatch under $\gamma_u=1$ and $\gamma_u=1/100$ on CIFAR-10-LT, respectively. Similarly, the CIFAR-100-LT results show that our method outperforms ACR by an average of 1.7\% accuracy increase. For STL10-LT, \algo\ achieves the best results with averaged 5.3\% accuracy gain compared with ACR, even if the unlabeled data distribution is unknown. Generally speaking, empirical results under unknown class distributions of unlabeled data on three datasets justify that \algo\ can effectively utilize unlabeled data to alleviate the negative impact of class imbalance.

\subsection{Results on ImageNet-127} \label{sec:imageNet_127}
ImageNet-127 was introduced in prior work\cite{huh2016makes} and later adopted for LTSSL by CReST\cite{wei2021crest}, which groups the 1000 classes of ImageNet\cite{deng2009imagenet} into 127 classes based on the WordNet hierarchy. Compared with other datasets, we do not need to construct the dataset artificially because it naturally follows a long-tailed class distribution with an imbalance ratio $\gamma \approx 286$, and we also employ post-hoc logit adjustment reduce \textit{residual bias} from long-tailed training. Following CoSSL\cite{fan2022cossl}, we down-sample the original images to smaller sizes of $32\times 32$ or $64 \times 64$ pixels using the box method from the Pillow library and randomly select 10\% training samples to form the labeled set. It is worth noting that the test set of ImageNet-127 is also long-tailed. The results are summarized in \mbox{\Cref{tab:small_imagenet}}. We can see that \algo\ achieves superior results for both image sizes $32\times 32$ and $64\times 64$ with 3.3\% and 4.8\% absolute improvement on test accuracy compared with ACR \cite{wei2023towards}, respectively. The results show that \algo\ can be successfully applied to tasks with long-tailed test datasets.

\begin{figure*}[t]
\centering
\subfloat[Gap for \textit{consistent}]{\includegraphics[width=2.35in]{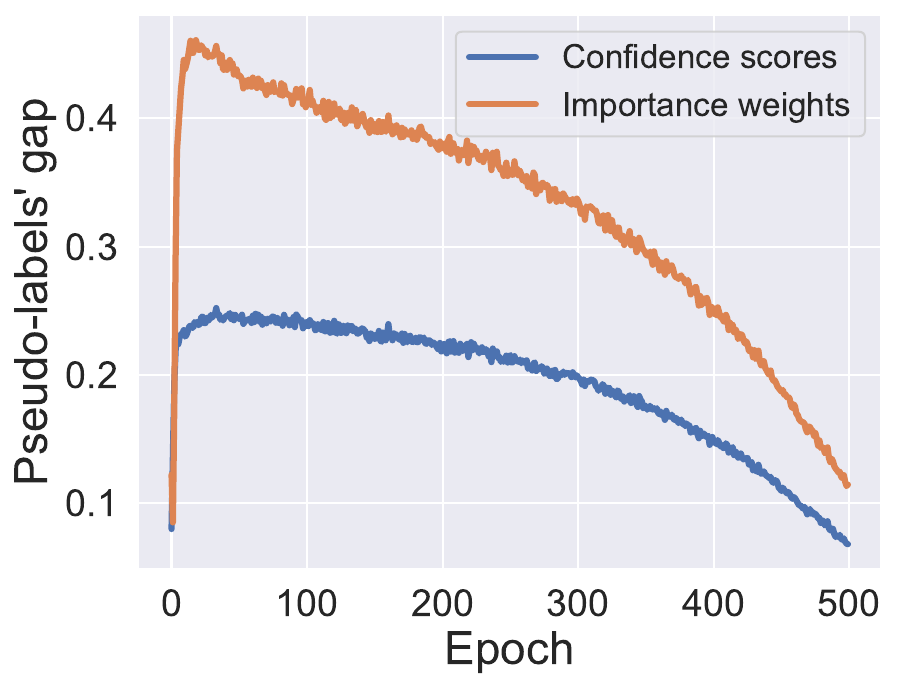}%
\label{fig:diff_con}}
\hfil
\subfloat[Gap for \textit{uniform}]{\includegraphics[width=2.35in]{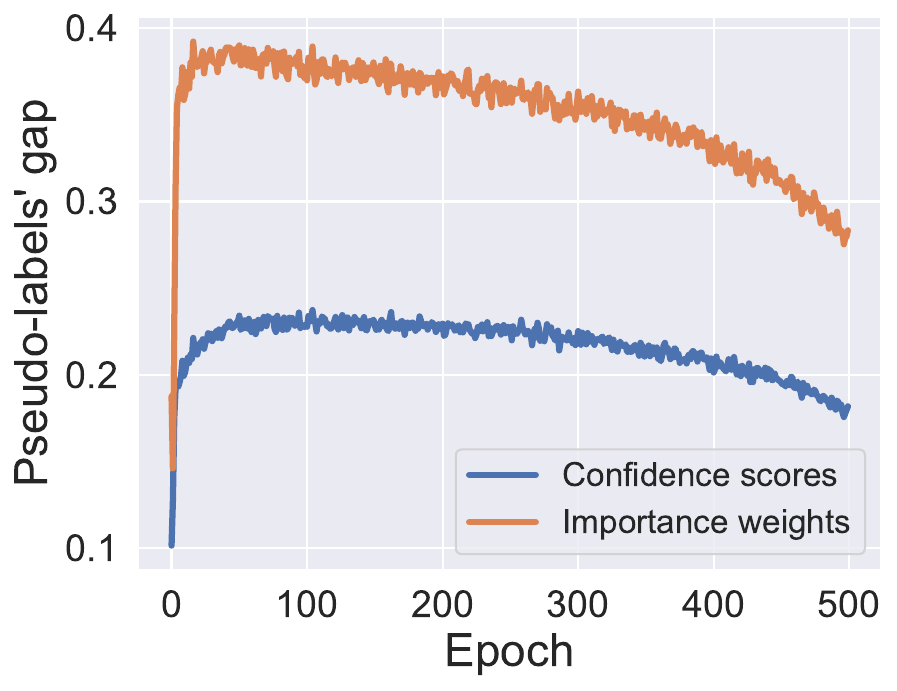}%
\label{fig:diff_uni}}
\hfil
\subfloat[Gap for \textit{reversed}]{\includegraphics[width=2.35in]{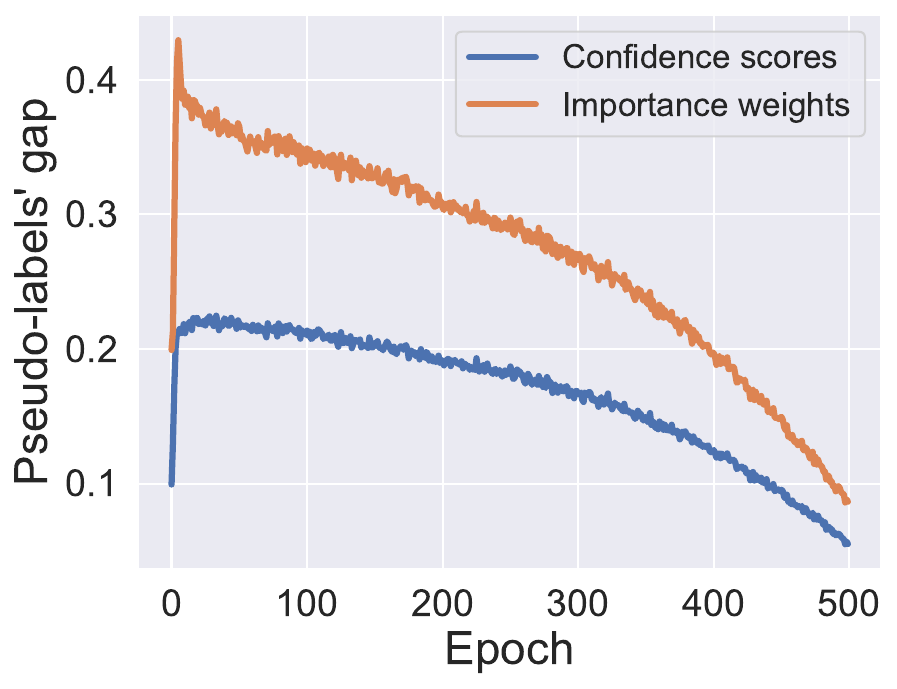}
\label{fig:diff_rev}}

\caption{The confidence scores and importance weights gap between accurate and incorrect pseudo-labels of different settings. The experiments are conducted on CIFAR-100-LT with $N_1=50, M_1=400,$ and $\gamma_l=10$.}
\label{fig:diff}
\end{figure*}

\begin{figure*}[t]
\centering
\subfloat[KL distance for \textit{consistent}]{\includegraphics[width=2.35in]{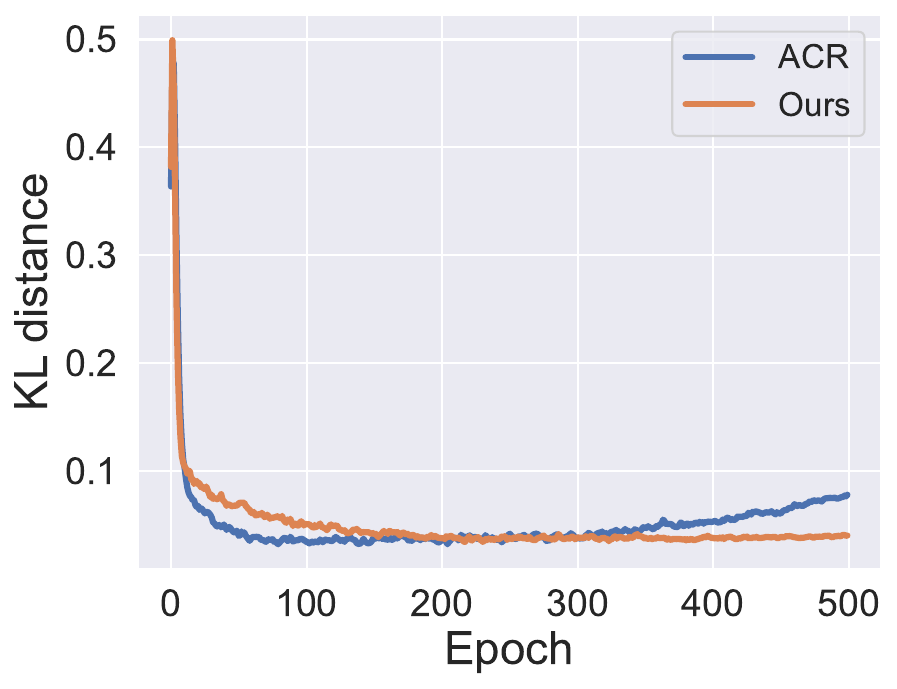}%
\label{fig:kl_con}}
\hfil
\subfloat[KL distance for \textit{uniform}]{\includegraphics[width=2.35in]{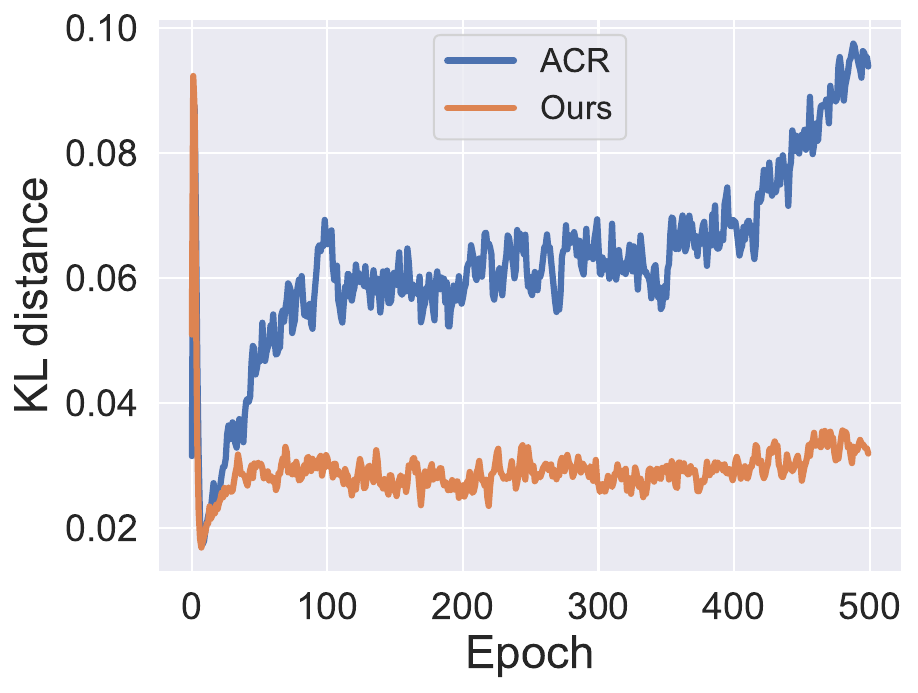}%
\label{fig:kl_uni}}
\hfil
\subfloat[KL distance for \textit{reversed}]{\includegraphics[width=2.35in]{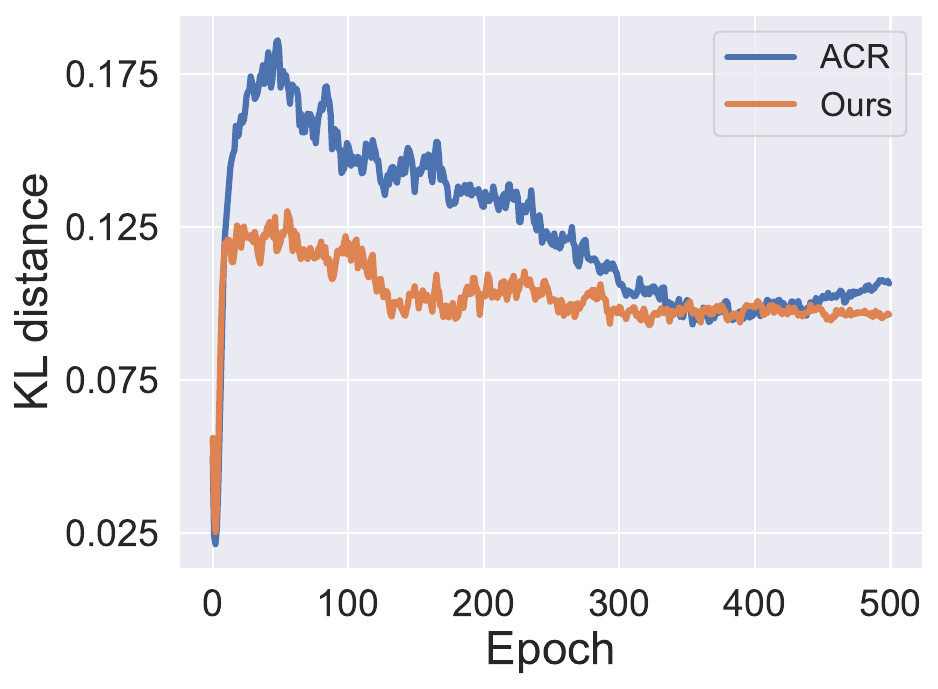}
\label{fig:kl_rev}}

\caption{The KL distance between the predicted and true distributions of the unlabeled data. The experiments are conducted on CIFAR-100-LT with $N_1=50, M_1=400,$ and $\gamma_l=10$.}
\label{fig:kl_dist}
\end{figure*}

\begin{table}[t]
\centering
\caption{Comparison of ACR and \algo\ under post-hoc enhancement with different $\tau_{3}$ values. The experiments are conducted on CIFAR-10-LT with $\gamma_l=100$, $N_1=500$, and $M_1=4000$.}
\resizebox{\linewidth}{!}{%
\begin{tabular}{@{}llccc@{}}
\toprule
 &
   &
  \multicolumn{3}{c}{CIFAR-10-LT} \\ \midrule
Algorithm &
   &
  $\gamma_u=100$ &
  $\gamma_u=1$ &
  $\gamma_u=1/100$ \\ \cmidrule(r){1-1} \cmidrule(l){3-5} 
\begin{tabular}[c]{@{}l@{}}ACR \cite{wei2023towards} ($\tau_{3}=0$)\\ \quad w/ $\tau_{3}=1$\\ \quad w/ $\tau_{3}=2$\\ \algo\ ($\tau_{3}=0$)\\ \quad w/ $\tau_{3}=1$\\ \quad w/ $\tau_{3}=2$\end{tabular} &
   &
  \begin{tabular}[c]{@{}c@{}}$81.6 \pm 0.19$\\ $82.4 \pm 0.45$\\ $82.0 \pm 0.42$\\ $77.8 \pm 0.43$\\ $82.3 \pm 0.44$\\ $84.1 \pm 0.56$\end{tabular} &
  \begin{tabular}[c]{@{}c@{}}$92.1 \pm 0.18$\\ $92.1 \pm 0.17$\\ $92.2 \pm 0.21$\\ $93.8 \pm 0.01$\\ $93.8 \pm 0.03$\\ $93.9 \pm 0.02$\end{tabular} &
  \begin{tabular}[c]{@{}c@{}}$84.9 \pm 0.09$\\ $85.4 \pm 0.19$\\ $86.0 \pm 1.23$\\ $85.5 \pm 0.33$\\ $86.3 \pm 0.26$\\ $86.6 \pm 0.33$\end{tabular} \\ \bottomrule
\end{tabular}%
}
\label{tab:post_hoc}
\end{table}

\begin{table}[ht]
\centering
\caption{Ablation studies of our proposed \algo\ algorithm. \textit{Con}, \textit{Uni}, and \textit{Rev} represent \textit{consistent}, \textit{uniform}, and \textit{reversed} for short.}
\resizebox{1\linewidth}{!}{%
\begin{tabular}{@{}lcccccc@{}}
\toprule
\multirow{2}{*}{Ablations} & \multicolumn{3}{c}{CIFAR-10-LT} & \multicolumn{3}{c}{CIFAR-100-LT} \\ \cmidrule(l){2-7} 
                           & Con      & Uni      & Rev      & Con       & Uni      & Rev      \\ \midrule
Ours                       & 84.1     & 93.9     & 86.6     & 52.0      & 60.4     & 53.3     \\
w/o alignment              & 84.2     & 84.6     & 66.8     & 51.9      & 60.2     & 50.1     \\
w/o balanced pseudo-labels & 82.2     & 93.9     & 86.2     & 51.7      & 58.0     & 49.9     \\
w/o balanced softmax & 84.0     & 93.5     & 86.1     & 51.1      & 59.6     & 52.7     \\
w/o weighting              & 83.8     & 93.3     & 86.5     & 51.8      & 58.3     & 52.1     \\ \bottomrule
\end{tabular}%
}
\label{tab:ablation}
\end{table}

\subsection{Systematic analysis of the proposed method}
\label{exp:sys_analysis}

To better understand our method, we conduct extensive studies to demonstrate the effectiveness of \algo.

\noindent
\textbf{Quality of pseudo-labels.} \Cref{fig:pseu_acc} compares the pseudo-label accuracy of ACR and \algo. We observe that \algo\ consistently achieves higher accuracy during training than ACR, and this improvement is important because pseudo-label quality plays a pivotal role in SSL performance \cite{chen2023softmatch}.


\noindent
\textbf{The iterative convergence of standard and balanced branch.} The core idea of \algo\ is to first decouple learning for head and tail classes and then have them converge. The standard branch initially focuses on head classes, while the balanced branch improves tail classes. Both branches progressively converge during training, ultimately achieving satisfactory performance across both head and tail classes. \Cref{fig:kl_qhatbal} shows that, across different settings, the distance between the predicted unlabeled data distribution of the two branches progressively decreases, indicating that the branches are converging towards one another. In addition, we present the precision and recall of pseudo-labels for head and tail classes in \Cref{fig:cifar10_head,fig:cifar10_tail}. Regarding the head classes, it is noted that the standard branch exhibits lower precision but higher recall compared to the balanced branch during the training, which indicates that the standard branch tends to favor the head classes \cite{wei2021robust}. Similarly, regarding the tail classes, the balanced branch shows lower precision and higher recall, suggesting its inclination towards the tail classes. In line with our expectations, the differences in precision and recall between the two branches reduced progressively for head and tail classes during the training. Simultaneously, there exists a general upward trend for most precision and recall, demonstrating that both branches are converging and achieving improved performance for all classes. The precision for tail classes exhibits a declining trend in later stages, and we attribute this to the expected phenomenon of tailward inclination, which helps mitigate the bias introduced by long-tailed distributions in the dataset. In conjunction with \Cref{fig:pseu_acc}, we can conclude that as the two branches progressively converge, the accuracy of pseudo-labels continually increases, thus improving the generalization performance of our model.


\noindent
\textbf{Hyperparameter sensitivity.} Despite the strong performance achieved by our method, the inclusion of certain crucial parameters necessitates an examination of whether the exceptional results are solely due to parameter tuning. Therefore, we explore the sensitivity of $\tau_{1}$ and $\tau_{2}$ in \Cref{fig:sen_analysis}. $\tau_{1}$ in \Cref{eq:std2bal_pseu} controls the logit adjustment intensity for the balanced branch's pseudo-labels. Our observations indicate marginal fluctuations in performance for both \textit{consistent} and \textit{reversed}, while the performance of \textit{uniform} displays a stable tendency. Generally, when $\tau_{1}$ equals $1$, better overall performance can be achieved. When considering $\tau_{2}$, we find that lower values result in poorer performance for \textit{reversed}. We attribute this to the significant distribution disparities between labeled and unlabeled data, and higher $\tau_{2}$ can provide strong balancing intensity to alleviate biases introduced by labeled data and shift towards the real distribution of unlabeled data. Therefore, we set $\tau_{2}=2$ in our experiments.




\noindent
\textbf{Why post-hoc adjustment is needed?} To illustrate the necessity of post-hoc adjustment for \algo, \Cref{tab:post_hoc} compares the performance effects of various enhancement intensities between ACR and \algo. Upon removing post-hoc adjustment ($\tau_{3} = 0$), we noticed that \algo\ didn't exhibit substantial performance benefits and even performed 3.8\% worse than ACR under \textit{consistent}. However, with the inclusion of post-hoc adjustment ($\tau_{3} = 2$), \algo\ shows a notable enhancement in performance averaging at 2.5\%, contrasting with the marginal 0.5\% improvement of ACR. This indicates that \algo\ possesses strong re-balanceable ability, which suggests that \algo\ can achieve substantial performance gains through simple post-hoc balancing adjustment. We attribute this superior performance to more attention of \algo\ on the accuracy of pseudo-labels during training. As depicted in \Cref{fig:pseu_acc}, \algo\ achieves higher pseudo-label accuracy compared to ACR, which is beneficial
for the model’s representation learning \cite{kang2019decoupling}. In contrast, ACR's strong logit adjustment during training introduces more erroneous pseudo-labels, hindering representation learning as evident in Supplementary Sec. S1 and consequently limiting the extent of performance gains achievable through post-hoc enhancement. Based on these results and findings, we conclude that an effective LTSSL model should prioritize enhancing the accuracy of pseudo-labels to achieve better representations during training. This prioritization ensures that model possesses a robust foundation for subsequent performance improvement through post-hoc adjustment.


\noindent
\textbf{Role of sample weights.} During optimization of the consistency regularizer in the balanced branch, we assign weight $\psi(x_j^{(u)})$ to the sample $x_j^{(u)}$. We hope to prioritize correct pseudo-labels with higher weights and give lower weights to incorrect samples. Therefore, we first define the gap between accurate and incorrect pseudo-labels:
%
\begin{align}
    \psi_{gap} = 
    \frac{1}{| \mathcal{U}^{+} |} \sum_{x_j^{(u)} \in \mathcal{U}^{+}} \psi(x_j^{(u)}) - 
    \frac{1}{| \mathcal{U}^{-} |} \sum_{x_j^{(u)} \in \mathcal{U}^{-}} \psi(x_j^{(u)})
    \label{eq:gap_weigts}
\end{align}
where $\mathcal{U}^{+} = \{x_j^{(u)} \mid \widetilde{q}_j = y_j^{(u)}, x_j^{(u)} \in \mathcal{D}^{u}\}$ and $\mathcal{U}^{-} = \{x_j^{(u)} \mid \widetilde{q}_j \neq y_j^{(u)}, x_j^{(u)} \in \mathcal{D}^{u}\}$ represent the collections of unlabeled samples with accurate and incorrect pseudo-labels, respectively. $y_j^{(u)}$ is the real label for unlabeled sample $x_j^{(u)}$. It is evident that we hope the disparity in weights between accurate and incorrect pseudo-labels is significant, so $\psi_{gap}$ should be as large as possible. To confirm the superiority of sample weights, we can also calculate the gap in confidence scores between accurate and incorrect pseudo-labels. The results are shown in \Cref{fig:diff}, and we can conclude that sample weights exhibit a larger gap compared to confidence scores across all settings, indicating that sample weights can more easily distinguish between accurate and incorrect pseudo-labels, thus mitigating the damage to the model's performance caused by erroneous pseudo-labels.

\noindent
\textbf{Estimation of unlabeled data distribution.} In our method, the alignment of the unlabeled data distribution in \Cref{eq:std_CE} and the adjustments made during post-hoc enhancement both demonstrate that the accuracy of the predicted unlabeled data distribution $\pi^{b}$ is critical to the performance of \algo. Therefore, we present the KL distance between the predicted distribution and the true unlabeled data distribution for different 
scenarios in \Cref{fig:kl_dist}. Evidently, as training progresses, \algo\ achieves more accurate distribution prediction $\pi^{b}$ compared to ACR, which is significantly beneficial for improving the performance of \algo.

\subsection{Ablation analysis}
\label{sec:ablation}

To better understand \algo, we tease apart the factors that contribute significantly to its success in \Cref{tab:ablation}.

\noindent
\textbf{Impact of alignment.} The alignment in \Cref{eq:std_CE} aims to leverage the correct label information from the labeled data to guide the predictions of the standard branch to align more closely with the inclination of the balanced branch. Without alignment, the labeled portion of the standard branch loss will be directly calculated using cross-entropy, which will consistently be influenced by biases introduced by imbalanced labeled data. In our experiments, we notice a significant performance decline upon removing the alignment, especially for \textit{uniform} and \textit{reversed}, averaging 8.1\%. The phenomenon suggests that alignment can effectively steer the model in the correct direction without being skewed by imbalanced labeled data.

\noindent
\textbf{Impact of balanced pseudo-labels.} In our proposed \algo, the balanced branch employs pseudo-labels from standard branch after adjustment by \Cref{eq:std2bal_pseu}. If we remove adjustment, we observe a decrease averaged 1.4\% in accuracy, indicating that balanced pseudo-labels can effectively aid balanced branch in mitigating bias towards head classes while maintaining high performance on tail classes.

  
  

\noindent
\textbf{Impact of balanced softmax.} The balanced softmax in \Cref{eq:bal_CE} effectively mitigates the issue of labeled data imbalance within the balanced branch, leading to more balanced predictions. The exclusion of the balanced softmax results in an average performance drop of 0.6\% across all settings, emphasizing the necessity of incorporating the balanced softmax into the labeled loss in balanced branch.


\noindent
\textbf{Impact of weighting.} In consistency regularizer for balanced branch in \Cref{eq:b-conloss}, we employ importance weights to evaluate the reliability of pseudo-labels. From \Cref{tab:ablation}, we observe a decrease in accuracy averaged 0.7\%, suggesting that importance weights can effectively filter out unstable pseudo-labels and enhance performance.

\section{Conclusion}
This paper introduces \algo, a simple yet effective framework for long-tailed semi-supervised learning with unknown unlabeled distributions. The core strategy of \algo\ is to first decouple the learning problem into standard branch and balanced branch, which specialize in head and tail classes respectively. Subsequently, these branches are encouraged to converge through carefully designed interactions, including cross-branch pseudo-labeling and distribution alignment. This process allows them to share knowledge, leading to a single, powerful model that achieves high accuracy across all classes. We implement the interactions by adjustments to the pseudo-labels and alignment to estimated distribution between two branches. We empirically show that our method significantly outperforms all competing methods under various scenarios, offering a solid baseline for future studies.

\bibliographystyle{IEEEtran}
\bibliography{IEEEabrv,IEEEtran}

@inproceedings{he2016deep,
  title={Deep residual learning for image recognition},
  author={He, Kaiming and Zhang, Xiangyu and Ren, Shaoqing and Sun, Jian},
  booktitle={Proceedings of the IEEE conference on computer vision and pattern recognition},
  pages={770--778},
  year={2016}
}

@article{krizhevsky2017imagenet,
  title={Imagenet classification with deep convolutional neural networks},
  author={Krizhevsky, Alex and Sutskever, Ilya and Hinton, Geoffrey E},
  journal={Communications of the ACM},
  volume={60},
  number={6},
  pages={84--90},
  year={2017},
  publisher={AcM New York, NY, USA}
}

@inproceedings{amodei2016deep,
  title={Deep speech 2: End-to-end speech recognition in english and mandarin},
  author={Amodei, Dario and Ananthanarayanan, Sundaram and Anubhai, Rishita and Bai, Jingliang and Battenberg, Eric and Case, Carl and Casper, Jared and Catanzaro, Bryan and Cheng, Qiang and Chen, Guoliang and others},
  booktitle={International conference on machine learning},
  pages={173--182},
  year={2016},
  organization={PMLR}
}

@article{tarvainen2017mean,
  title={Mean teachers are better role models: Weight-averaged consistency targets improve semi-supervised deep learning results},
  author={Tarvainen, Antti and Valpola, Harri},
  journal={Advances in Neural Information Processing Systems},
  volume={30},
  pages = {1195--1204},
  year={2017}
}

@article{miyato2018virtual,
  title={Virtual adversarial training: a regularization method for supervised and semi-supervised learning},
  author={Miyato, Takeru and Maeda, Shin-ichi and Koyama, Masanori and Ishii, Shin},
  journal={IEEE Transactions on Pattern Analysis and Machine Intelligence},
  volume={41},
  number={8},
  pages={1979--1993},
  year={2018}
}

@article{berthelot2019mixmatch,
  title={Mixmatch: A holistic approach to semi-supervised learning},
  author={Berthelot, David and Carlini, Nicholas and Goodfellow, Ian and Papernot, Nicolas and Oliver, Avital and Raffel, Colin A},
  journal={Advances in Neural Information Processing Systems},
  volume={32},
  pages     = {5050--5060},
  year={2019}
}

@article{sohn2020fixmatch,
  title={Fixmatch: Simplifying semi-supervised learning with consistency and confidence},
  author={Sohn, Kihyuk and Berthelot, David and Carlini, Nicholas and Zhang, Zizhao and Zhang, Han and Raffel, Colin A and Cubuk, Ekin Dogus and Kurakin, Alexey and Li, Chun-Liang},
  journal={Advances in Neural Information Processing Systems},
  volume={33},
  pages={596--608},
  year={2020}
}

@inproceedings{DBLP:conf/nips/XieDHL020,
  author    = {Qizhe Xie and
               Zihang Dai and
               Eduard H. Hovy and
               Thang Luong and
               Quoc Le},
  title     = {Unsupervised Data Augmentation for Consistency Training},
  booktitle = {Advances in Neural Information Processing Systems},
  year      = {2020}
}

@article{weit2020tnnls,
  title={Does Tail Label Help for Large-Scale Multi-Label Learning?},
  author={Wei, Tong and Li, Yu-Feng},
  journal={IEEE Transactions on Neural Networks and Learning Systems},
  volume={31},
  number={7},
  pages={2315--2324},
  year={2019}
}

@inproceedings{zhou2020bbn,
  title={Bbn: Bilateral-branch network with cumulative learning for long-tailed visual recognition},
  author={Zhou, Boyan and Cui, Quan and Wei, Xiu-Shen and Chen, Zhao-Min},
  booktitle={Proceedings of the IEEE/CVF conference on computer vision and pattern recognition},
  pages={9719--9728},
  year={2020}
}

@inproceedings{xiang2020learning,
  title={Learning from multiple experts: Self-paced knowledge distillation for long-tailed classification},
  author={Xiang, Liuyu and Ding, Guiguang and Han, Jungong},
  booktitle={European Conference on Computer Vision},
  pages={247--263},
  year={2020},
  organization={Springer}
}

@article{wang2020long,
  title={Long-tailed recognition by routing diverse distribution-aware experts},
  author={Wang, Xudong and Lian, Long and Miao, Zhongqi and Liu, Ziwei and Yu, Stella X},
  journal={arXiv preprint arXiv:2010.01809},
  year={2020}
}

@inproceedings{li2022nested,
  title={Nested Collaborative Learning for Long-Tailed Visual Recognition},
  author={Li, Jun and Tan, Zichang and Wan, Jun and Lei, Zhen and Guo, Guodong},
  booktitle={Proceedings of the IEEE/CVF Conference on Computer Vision and Pattern Recognition},
  pages={6949--6958},
  year={2022}
}

@inproceedings{cui2021parametric,
  title={Parametric contrastive learning},
  author={Cui, Jiequan and Zhong, Zhisheng and Liu, Shu and Yu, Bei and Jia, Jiaya},
  booktitle={Proceedings of the IEEE/CVF international conference on computer vision},
  pages={715--724},
  year={2021}
}

@inproceedings{liu2019large,
  title={Large-scale long-tailed recognition in an open world},
  author={Liu, Ziwei and Miao, Zhongqi and Zhan, Xiaohang and Wang, Jiayun and Gong, Boqing and Yu, Stella X},
  booktitle={Proceedings of the IEEE/CVF Conference on Computer Vision and Pattern Recognition},
  pages={2537--2546},
  year={2019}
}

@article{Wei_2021_RoLT,
  author    = {Tong Wei and
               Jiang{-}Xin Shi and
               Wei{-}Wei Tu and
               Yu{-}Feng Li},
  title     = {Robust Long-Tailed Learning under Label Noise},
  journal   = {CoRR},
  volume    = {abs/2108.11569},
  year      = {2021}
}

@inproceedings{beierxERM,
  title={Cross-Domain Empirical Risk Minimization for Unbiased Long-tailed Classification},
  author={Zhu, Beier and Niu, Yulei and Hua, Xian-Sheng and Zhang, Hanwang},
  booktitle={Proceedings of the AAAI Conference on Artificial Intelligence},
  year={2022}
}

@article{lee2021abc,
  title={ABC: Auxiliary Balanced Classifier for Class-imbalanced Semi-supervised Learning},
  author={Lee, Hyuck and Shin, Seungjae and Kim, Heeyoung},
  journal={Advances in Neural Information Processing Systems},
  volume={34},
  pages= {7082--7094},
  year={2021}
}

@inproceedings{DBLP:conf/icml/LaiWGCC22,
  author    = {Zhengfeng Lai and
               Chao Wang and
               Henrry Gunawan and
               Sen{-}Ching S. Cheung and
               Chen{-}Nee Chuah},
  title     = {Smoothed Adaptive Weighting for Imbalanced Semi-Supervised Learning: Improve Reliability Against Unknown Distribution Data},
  booktitle = {International Conference on Machine Learning},
  pages     = {11828--11843},
  year      = {2022}
}

@article{wei2022transfer,
  title={Transfer and Share: Semi-Supervised Learning from Long-Tailed Data},
  author={Wei, Tong and Liu, Qian-Yu and Shi, Jiang-Xin and Tu, Wei-Wei and Guo, Lan-Zhe},
  journal={Machine Learning},
  year={2022}
}

@article{kim2020distribution,
  title={Distribution aligning refinery of pseudo-label for imbalanced semi-supervised learning},
  author={Kim, Jaehyung and Hur, Youngbum and Park, Sejun and Yang, Eunho and Hwang, Sung Ju and Shin, Jinwoo},
  journal={Advances in Neural Information Processing Systems},
  volume={33},
  pages={14567--14579},
  year={2020}
}

@inproceedings{wei2021crest,
  title={Crest: A class-rebalancing self-training framework for imbalanced semi-supervised learning},
  author={Wei, Chen and Sohn, Kihyuk and Mellina, Clayton and Yuille, Alan and Yang, Fan},
  booktitle={Proceedings of the IEEE/CVF Conference on Computer Vision and Pattern Recognition},
  pages={10857--10866},
  year={2021}
}

@inproceedings{oh2022daso,
  title={DASO: Distribution-Aware Semantics-Oriented Pseudo-Label for Imbalanced Semi-Supervised Learning},
  author={Oh, Youngtaek and Kim, Dong-Jin and Kweon, In So},
  booktitle={Proceedings of the IEEE/CVF Conference on Computer Vision and Pattern Recognition},
  pages={9786--9796},
  year={2022}
}

@inproceedings{berthelot2019remixmatch,
  title={ReMixMatch: Semi-Supervised Learning with Distribution Matching and Augmentation Anchoring},
  author={Berthelot, David and Carlini, Nicholas and Cubuk, Ekin D and Kurakin, Alex and Sohn, Kihyuk and Zhang, Han and Raffel, Colin},
  booktitle={International Conference on Learning Representations},
  year={2019}
}

@inproceedings{DBLP:conf/iclr/ZhangCDL18,
  author    = {Hongyi Zhang and
               Moustapha Ciss{\'{e}} and
               Yann N. Dauphin and
               David Lopez{-}Paz},
  title     = {mixup: Beyond Empirical Risk Minimization},
  booktitle = {International Conference on Learning Representations},
  year      = {2018}
}

@inproceedings{fan2022cossl,
  title={CoSSL: Co-Learning of Representation and Classifier for Imbalanced Semi-Supervised Learning},
  author={Fan, Yue and Dai, Dengxin and Kukleva, Anna and Schiele, Bernt},
  booktitle={Proceedings of the IEEE/CVF Conference on Computer Vision and Pattern Recognition},
  pages={14574--14584},
  year={2022}
}

@article{devries2017improved,
  title={Improved regularization of convolutional neural networks with cutout},
  author={DeVries, Terrance and Taylor, Graham W},
  journal={arXiv preprint arXiv:1708.04552},
  year={2017}
}

@inproceedings{cubuk2020randaugment,
  title={Randaugment: Practical automated data augmentation with a reduced search space},
  author={Cubuk, Ekin D and Zoph, Barret and Shlens, Jonathon and Le, Quoc V},
  booktitle={Proceedings of the IEEE/CVF conference on computer vision and pattern recognition workshops},
  pages={702--703},
  year={2020}
}

@article{krizhevsky2009learning,
  title={Learning multiple layers of features from tiny images},
  author={Krizhevsky, Alex and Hinton, Geoffrey and others},
  year={2009}
}

@inproceedings{coates2011analysis,
  title={An analysis of single-layer networks in unsupervised feature learning},
  author={Coates, Adam and Ng, Andrew and Lee, Honglak},
  booktitle={Proceedings of the fourteenth international conference on artificial intelligence and statistics},
  pages={215--223},
  year={2011},
  organization={JMLR Workshop and Conference Proceedings}
}

@article{oliver2018realistic,
  title={Realistic evaluation of deep semi-supervised learning algorithms},
  author={Oliver, Avital and Odena, Augustus and Raffel, Colin A and Cubuk, Ekin Dogus and Goodfellow, Ian},
  journal={Advances in neural information processing systems},
  volume={31},
  year={2018}
}

@article{zagoruyko2016wide,
  title={Wide residual networks},
  author={Zagoruyko, Sergey and Komodakis, Nikos},
  journal={arXiv preprint arXiv:1605.07146},
  year={2016}
}

@inproceedings{sutskever2013importance,
  title={On the importance of initialization and momentum in deep learning},
  author={Sutskever, Ilya and Martens, James and Dahl, George and Hinton, Geoffrey},
  booktitle={International conference on machine learning},
  pages={1139--1147},
  year={2013},
  organization={PMLR}
}

@article{polyak1964some,
  title={Some methods of speeding up the convergence of iteration methods},
  author={Polyak, Boris T},
  journal={Ussr computational mathematics and mathematical physics},
  volume={4},
  number={5},
  pages={1--17},
  year={1964},
  publisher={Elsevier}
}

@inproceedings{nesterov27method,
  title={A method of solving a convex programming problem with convergence rate $O(1/k^2)$},
  author={Nesterov, Yurii},
  booktitle={Sov. Math. Dokl},
  volume={27}
}

@article{loshchilov2016sgdr,
  title={Sgdr: Stochastic gradient descent with warm restarts},
  author={Loshchilov, Ilya and Hutter, Frank},
  journal={arXiv preprint arXiv:1608.03983},
  year={2016}
}

@inproceedings{kang2019decoupling,
  title={Decoupling representation and classifier for long-tailed recognition},
  author={Kang, Bingyi and Xie, Saining and Rohrbach, Marcus and Yan, Zhicheng
          and Gordo, Albert and Feng, Jiashi and Kalantidis, Yannis},
  booktitle={International Conference on Learning Representations},
  year={2020}
}

@article{ren2020balanced,
  title={Balanced meta-softmax for long-tailed visual recognition},
  author={Ren, Jiawei and Yu, Cunjun and Ma, Xiao and Zhao, Haiyu and Yi, Shuai and others},
  journal={Advances in Neural Information Processing Systems},
  volume={33},
  pages={4175--4186},
  year={2020}
}

@inproceedings{menon2020long,
  title={Long-tail learning via logit adjustment},
  author={Menon, Aditya Krishna and Jayasumana, Sadeep and Rawat, Ankit Singh and Jain, Himanshu and Veit, Andreas and Kumar, Sanjiv},
  booktitle={International Conference on Learning Representations},
  year={2020}
}

@inproceedings{arazo2020pseudo,
  title={Pseudo-labeling and confirmation bias in deep semi-supervised learning},
  author={Arazo, Eric and Ortego, Diego and Albert, Paul and O’Connor, Noel E and McGuinness, Kevin},
  booktitle={IJCNN},
  pages={1--8},
  year={2020}
}

@inproceedings{wang2021self,
  title={Self-tuning for data-efficient deep learning},
  author={Wang, Ximei and Gao, Jinghan and Long, Mingsheng and Wang, Jianmin},
  booktitle={ICML},
  pages={10738--10748},
  year={2021}
}

@inproceedings{wei2023towards,
  title={Towards realistic long-tailed semi-supervised learning: Consistency is all you need},
  author={Wei, Tong and Gan, Kai},
  booktitle={Proceedings of the IEEE/CVF Conference on Computer Vision and Pattern Recognition},
  pages={3469--3478},
  year={2023}
}

@inproceedings{du2024simpro,
  author       = {Chaoqun Du and
                  Yizeng Han and
                  Gao Huang},
  title        = {SimPro: {A} Simple Probabilistic Framework Towards Realistic Long-Tailed
                  Semi-Supervised Learning},
  booktitle    = {Forty-first International Conference on Machine Learning},
  pages        = {11686--11703},
  year         = {2024}
}

@inproceedings{jia2022visual,
  title={Visual prompt tuning},
  author={Jia, Menglin and Tang, Luming and Chen, Bor-Chun and Cardie, Claire and Belongie, Serge and Hariharan, Bharath and Lim, Ser-Nam},
  booktitle={European Conference on Computer Vision},
  pages={709--727},
  year={2022},
  organization={Springer}
}

@article{huh2016makes,
  title={What makes ImageNet good for transfer learning?},
  author={Huh, Minyoung and Agrawal, Pulkit and Efros, Alexei A},
  journal={arXiv preprint arXiv:1608.08614},
  year={2016}
}

@inproceedings{deng2009imagenet,
  title={Imagenet: A large-scale hierarchical image database},
  author={Deng, Jia and Dong, Wei and Socher, Richard and Li, Li-Jia and Li, Kai and Fei-Fei, Li},
  booktitle={2009 IEEE conference on computer vision and pattern recognition},
  pages={248--255},
  year={2009},
  organization={Ieee}
}

@article{chen2022adaptformer,
  title={Adaptformer: Adapting vision transformers for scalable visual recognition},
  author={Chen, Shoufa and Ge, Chongjian and Tong, Zhan and Wang, Jiangliu and Song, Yibing and Wang, Jue and Luo, Ping},
  journal={NeurIPS},
  volume={35},
  pages={16664--16678},
  year={2022}
}

@article{chen2023softmatch,
  title={Softmatch: Addressing the quantity-quality trade-off in semi-supervised learning},
  author={Chen, Hao and Tao, Ran and Fan, Yue and Wang, Yidong and Wang, Jindong and Schiele, Bernt and Xie, Xing and Raj, Bhiksha and Savvides, Marios},
  journal={arXiv preprint},
  year={2023}
}

@article{van2008visualizing,
  title={Visualizing data using t-SNE},
  author={Van der Maaten, Laurens and Hinton, Geoffrey},
  journal={Journal of Machine Learning Research},
  volume={9},
  number={11},
  year={2008}
}

@inproceedings{radford2021learning,
  title={Learning transferable visual models from natural language supervision},
  author={Radford, Alec and Kim, Jong Wook and Hallacy, Chris and Ramesh, Aditya and Goh, Gabriel and Agarwal, Sandhini and Sastry, Girish and Askell, Amanda and Mishkin, Pamela and Clark, Jack and others},
  booktitle={ICML},
  pages={8748--8763},
  year={2021}
}

@article{liu2022few,
  title={Few-shot parameter-efficient fine-tuning is better and cheaper than in-context learning},
  author={Liu, Haokun and Tam, Derek and Muqeeth, Mohammed and Mohta, Jay and Huang, Tenghao and Bansal, Mohit and Raffel, Colin A},
  journal={NeurIPS},
  volume={35},
  pages={1950--1965},
  year={2022}
}

@article{shi2023parameter,
  title={Parameter-Efficient Long-Tailed Recognition},
  author={Shi, Jiang-Xin and Wei, Tong and Zhou, Zhi and Han, Xin-Yan and Shao, Jie-Jing and Li, Yu-Feng},
  journal={arXiv preprint},
  year={2023}
}

@article{yang2022parameter,
  title={Parameter-efficient tuning makes a good classification head},
  author={Yang, Zhuoyi and Ding, Ming and Guo, Yanhui and Lv, Qingsong and Tang, Jie},
  journal={arXiv preprint},
  year={2022}
}

@article{dosovitskiy2020image,
  title={An image is worth 16x16 words: Transformers for image recognition at scale},
  author={Dosovitskiy, Alexey and Beyer, Lucas and Kolesnikov, Alexander and Weissenborn, Dirk and Zhai, Xiaohua and Unterthiner, Thomas and Dehghani, Mostafa and Minderer, Matthias and Heigold, Georg and Gelly, Sylvain and others},
  journal={arXiv preprint},
  year={2020}
}

@article{chen2022revisiting,
  title={Revisiting parameter-efficient tuning: Are we really there yet?},
  author={Chen, Guanzheng and Liu, Fangyu and Meng, Zaiqiao and Liang, Shangsong},
  journal={arXiv preprint},
  year={2022}
}

@article{hu2021lora,
  title={Lora: Low-rank adaptation of large language models},
  author={Hu, Edward J and Shen, Yelong and Wallis, Phillip and Allen-Zhu, Zeyuan and Li, Yuanzhi and Wang, Shean and Wang, Lu and Chen, Weizhu},
  journal={arXiv preprint},
  year={2021}
}

@article{zhang2021flexmatch,
  title={Flexmatch: Boosting semi-supervised learning with curriculum pseudo labeling},
  author={Zhang, Bowen and Wang, Yidong and Hou, Wenxin and Wu, Hao and Wang, Jindong and Okumura, Manabu and Shinozaki, Takahiro},
  journal={NeurIPS},
  volume={34},
  pages={18408--18419},
  year={2021}
}

@article{wang2022freematch,
  title={Freematch: Self-adaptive thresholding for semi-supervised learning},
  author={Wang, Yidong and Chen, Hao and Heng, Qiang and Hou, Wenxin and Fan, Yue and Wu, Zhen and Wang, Jindong and Savvides, Marios and Shinozaki, Takahiro and Raj, Bhiksha and others},
  journal={arXiv preprint},
  year={2022}
}

@inproceedings{zhong2021improving,
  title={Improving calibration for long-tailed recognition},
  author={Zhong, Zhisheng and Cui, Jiequan and Liu, Shu and Jia, Jiaya},
  booktitle={Proceedings of the IEEE/CVF conference on computer vision and pattern recognition},
  pages={16489--16498},
  year={2021}
}

@article{zhang2022self,
  title={Self-supervised aggregation of diverse experts for test-agnostic long-tailed recognition},
  author={Zhang, Yifan and Hooi, Bryan and Hong, Lanqing and Feng, Jiashi},
  journal={Advances in Neural Information Processing Systems},
  volume={35},
  pages={34077--34090},
  year={2022}
}

@article{gan2024erasing,
  title={Erasing the Bias: Fine-Tuning Foundation Models for Semi-Supervised Learning},
  author={Gan, Kai and Wei, Tong},
  journal={arXiv preprint arXiv:2405.11756},
  year={2024}
}

@article{wei2021robust,
  title={Robust long-tailed learning under label noise},
  author={Wei, Tong and Shi, Jiang-Xin and Tu, Wei-Wei and Li, Yu-Feng},
  journal={arXiv preprint arXiv:2108.11569},
  year={2021}
}

@article{ye2023bridging,
  title={Bridging the Gap: Learning Pace Synchronization for Open-World Semi-Supervised Learning},
  author={Ye, Bo and Gan, Kai and Wei, Tong and Zhang, Min-Ling},
  journal={arXiv preprint arXiv:2309.11930},
  year={2023}
}

@article{huang2023flatmatch,
  title={Flatmatch: Bridging labeled data and unlabeled data with cross-sharpness for semi-supervised learning},
  author={Huang, Zhuo and Shen, Li and Yu, Jun and Han, Bo and Liu, Tongliang},
  journal={Advances in Neural Information Processing Systems},
  volume={36},
  pages={18474--18494},
  year={2023}
}

@article{huang2024interlude,
  title={InterLUDE: Interactions between Labeled and Unlabeled Data to Enhance Semi-Supervised Learning},
  author={Huang, Zhe and Yu, Xiaowei and Zhu, Dajiang and Hughes, Michael C},
  journal={arXiv preprint arXiv:2403.10658},
  year={2024}
}

@inproceedings{CDMAD,
  author       = {Hyuck Lee and
                  Heeyoung Kim},
  title        = {{CDMAD:} Class-Distribution-Mismatch-Aware Debiasing for Class-Imbalanced
                  Semi-Supervised Learning},
  booktitle    = {{IEEE/CVF} Conference on Computer Vision and Pattern Recognition (CVPR)},
  pages        = {23891--23900},
  year         = {2024}
}

@inproceedings{gufourier,
  title={Fourier Clouds: Fast Bias Correction for Imbalanced Semi-Supervised Learning},
  author={Gu, Jiawei and Wang, Yidi and Sun, Qingqiang and Li, Xinming and Luo, Xiao and Qiao, Ziyue},
  booktitle={The Thirty-ninth Annual Conference on Neural Information Processing Systems},
  year = {2025}
}

@article{ouyang2025semantic,
  title={Semantic-Aware Pseudo-Labeling for Unsupervised Meta-Learning},
  author={Ouyang, Tianran and Dong, Xingping and Ye, Mang and Du, Bo and Shao, Ling and Shen, Jianbing},
  journal={IEEE Transactions on Pattern Analysis and Machine Intelligence},
  year={2025},
  publisher={IEEE}
}

@article{dong2024pseudo,
  title={Pseudo-labeling based practical semi-supervised meta-training for few-shot learning},
  author={Dong, Xingping and Ouyang, Tianran and Liao, Shengcai and Du, Bo and Shao, Ling},
  journal={IEEE Transactions on Image Processing},
  year={2024},
  publisher={IEEE}
}

@inproceedings{dong2022rethinking,
  title={Rethinking clustering-based pseudo-labeling for unsupervised meta-learning},
  author={Dong, Xingping and Shen, Jianbing and Shao, Ling},
  booktitle={European conference on computer vision},
  pages={169--186},
  year={2022},
  organization={Springer}
}

\clearpage
\onecolumn

\section*{Supplementary Material}
\beginsupplement

\section{Additional Experimental Results and Analysis}

\begin{table}[t]
\centering
\small
\setlength{\tabcolsep}{5pt}
\renewcommand{\arraystretch}{1.15}
\caption{Compact summary of the main priors/distributions and temperature parameters used in Section 3.}
\label{tab:notation_summary}
\begin{tabular}{p{1.8cm}p{3.6cm}p{2.6cm}p{5.1cm}}
\hline
\textbf{Symbol} & \textbf{Meaning} & \textbf{Where used} & \textbf{Role in DECON} \\
\hline
$\pi$ & Labeled empirical prior & Balanced branch, Eq.~(3) & Used in balanced softmax to mitigate the bias introduced by the imbalanced labeled data. \\

$\pi^{b}$ & EMA prior estimated by the balanced branch & Standard branch, Eq.~(4)--(6) & Used as the alignment prior for the standard branch. \\

$\pi^{s}$ & EMA prior estimated by the standard branch & Balanced branch, Eq.~(7) & Used for logit adjustment to generate balanced pseudo-labels for the balanced branch. \\

$\tau_{1}$ & Pseudo-label adjustment strength & Eq.~(7) & Controls the adjustment intensity for balanced pseudo-label generation. \\

$\tau_{2}$ & Alignment strength & Eq.~(4) & Controls the alignment intensity in the standard branch. \\

$\tau_{3}$ & Post-hoc adjustment strength & Eq.~(11) & Controls the calibration intensity during inference. \\
\hline
\end{tabular}
\end{table}

\begin{figure}[h]
\centering
\subfloat[]{\includegraphics[width=2.73in]{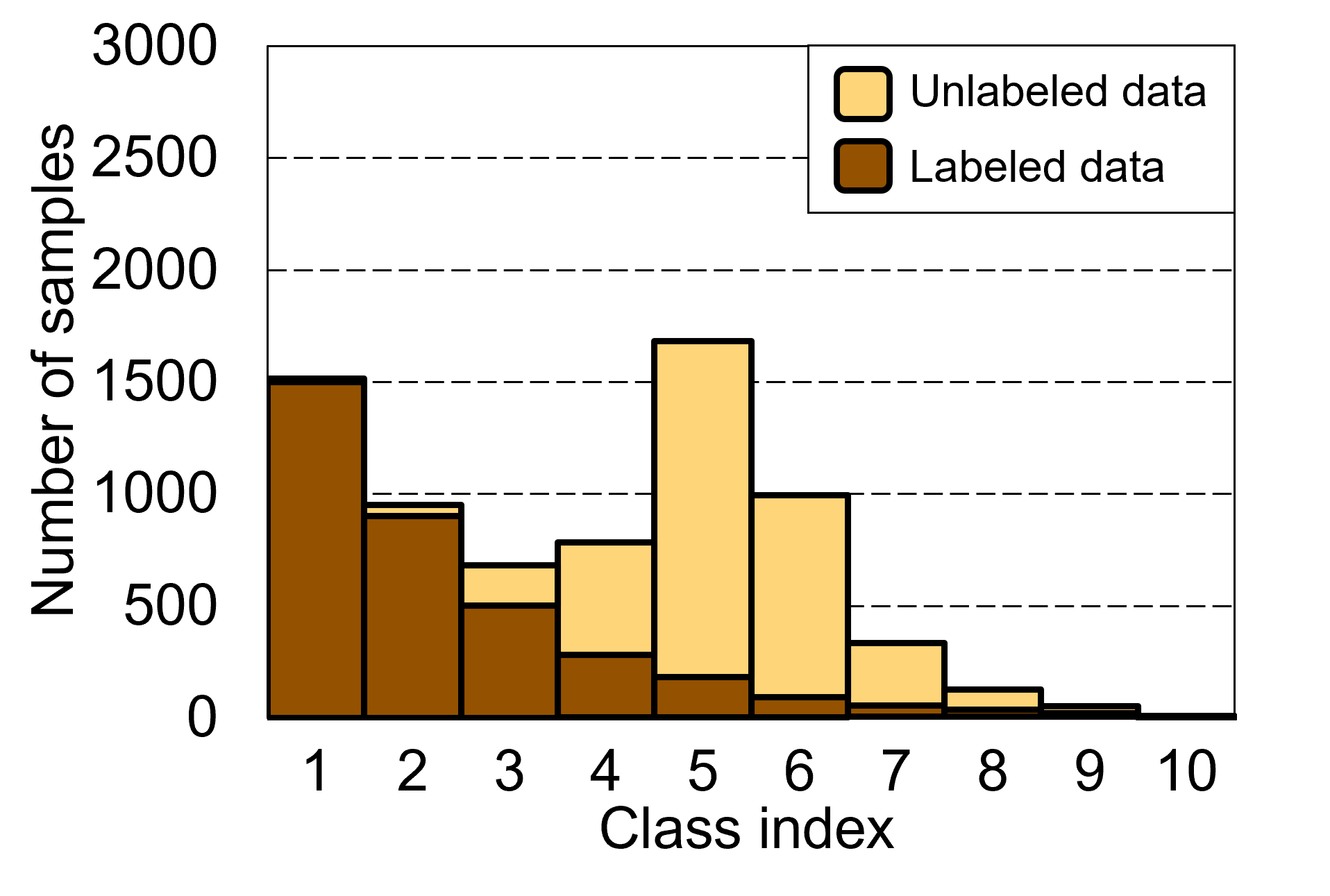}%
\label{fig:dist_middle}}
\hfil
\subfloat[]{\includegraphics[width=2.73in]{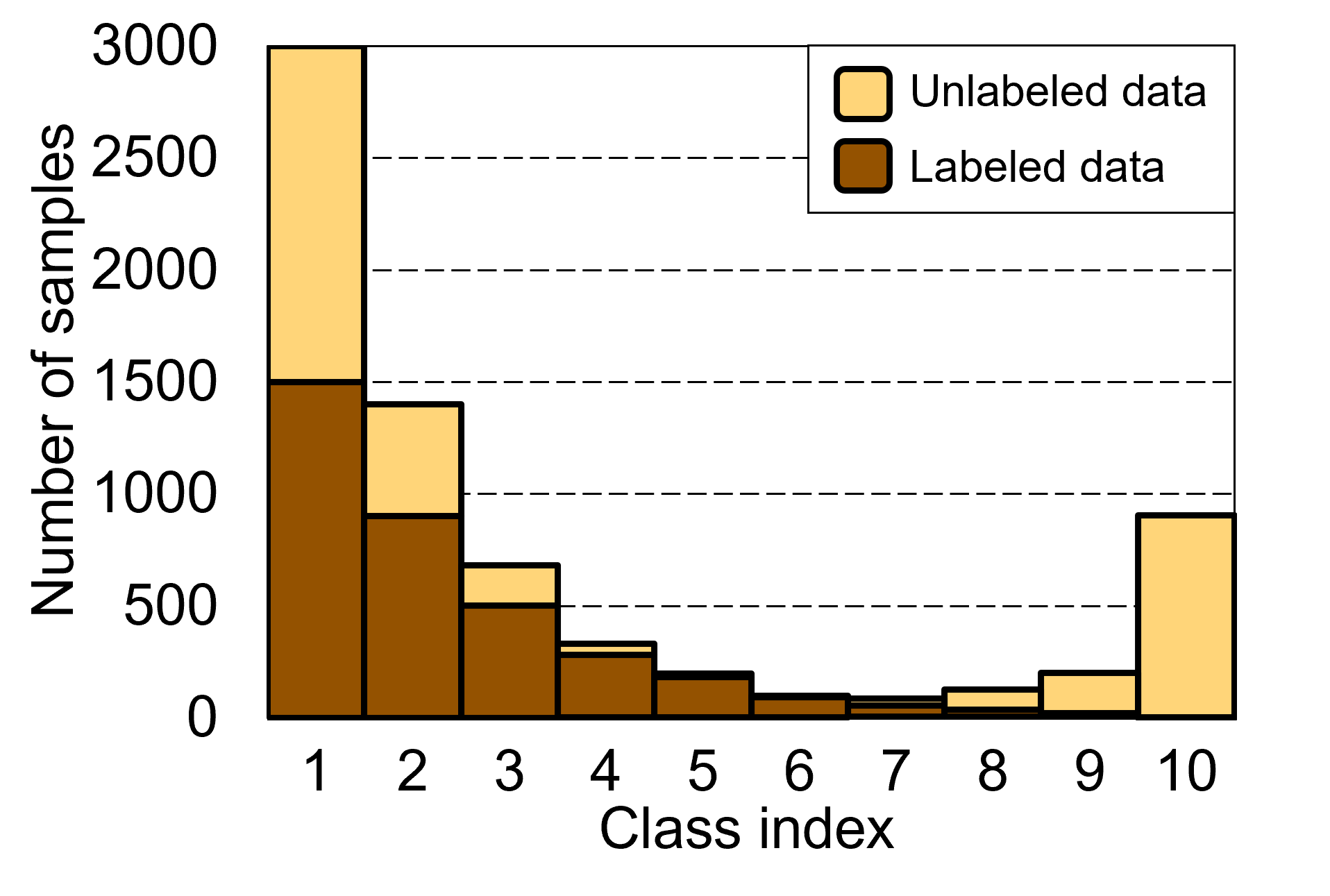}%
\label{fig:dist_headtail}}
\caption{More class distribution patterns for unlabeled data, i.e., \textit{middle} and \textit{headtail}.}
\label{fig:dist_more}
\end{figure}

\begin{figure*}[h]
\centering
\subfloat[]{\includegraphics[width=2.0in]{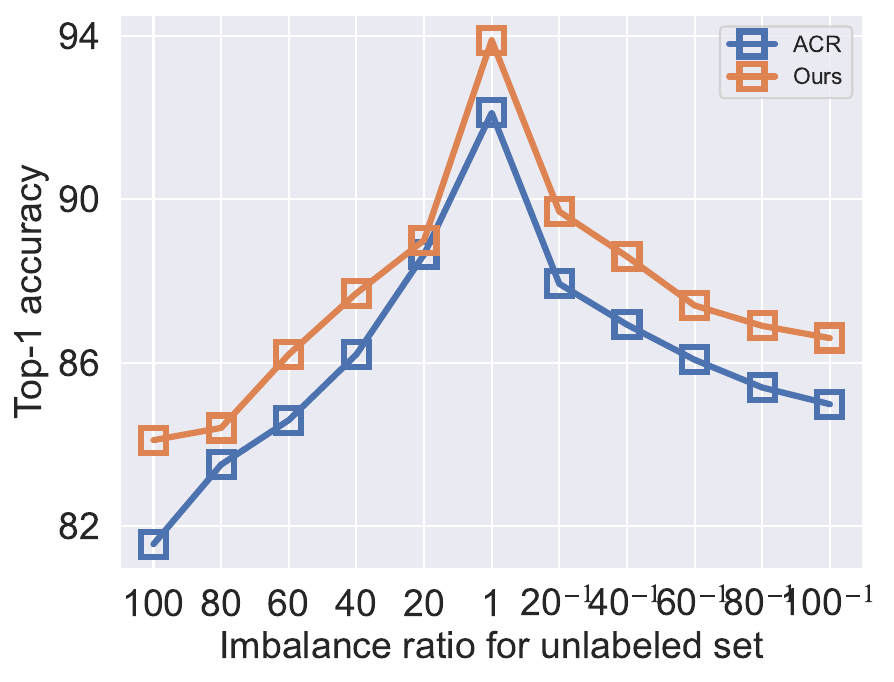}%
\label{fig:more_cifar10}}
\hfil
\subfloat[]{\includegraphics[width=2.0in]{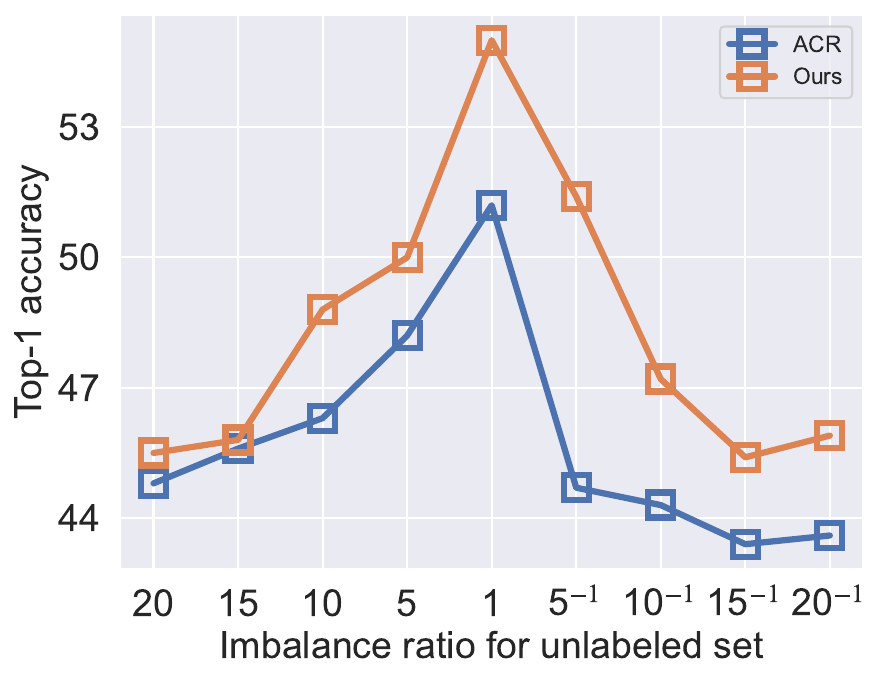}%
\label{fig:more_cifar100}}
\hfil
\subfloat[]{\includegraphics[width=2.0in]{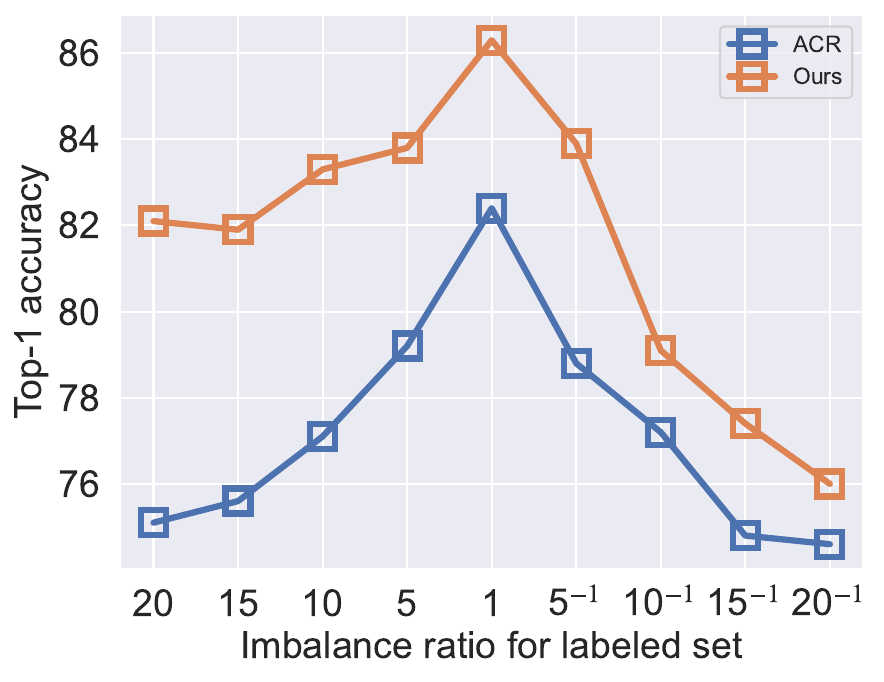}%
\label{fig:more_stl10}}
\caption{More realistic LTSSL settings with various imbalance ratio for unlabeled data or labeled data. (\labelcref{fig:more_cifar10}): $N_1=500, M_1=4000$ with $\gamma_l=100$ on CIFAR-10-LT. (\labelcref{fig:more_cifar100}): $N_1=50, M_1=400$ with $\gamma_l=20$ on CIFAR-100-LT. (\labelcref{fig:more_stl10}): $N_1=150$ with various imbalanced ratio for labeled set on STL10-LT.}
\label{fig:more_settings}
\end{figure*}

\begin{table}[h]
\centering
\caption{Test accuracy on CIFAR-10-LT and CIFAR-100-LT dataset under \textit{middle} and \textit{headtail} settings. We set $N_1=500, M_1=4000$ for CIFAR-10-LT and $N_1=50, M_1=400$ for CIFAR-100-LT. The best results are in \textbf{bold}.}
\resizebox{0.8\linewidth}{!}{%
\begin{tabular}{@{}llccccc@{}}
\toprule
                           &  & \multicolumn{2}{c}{CIFAR-10-LT ($\gamma_l=100$)} &  & \multicolumn{2}{c}{CIFAR-100-LT ($\gamma_l=20$)} \\ \midrule
\multirow{2}{*}{Algorithm} &  & \multicolumn{2}{c}{$\gamma_u=100$}             &  & \multicolumn{2}{c}{$\gamma_u=20$}              \\ \cmidrule(lr){3-4} \cmidrule(l){6-7} 
                           &  & middle                & headtail               &  & middle                & headtail               \\ \cmidrule(r){1-1} \cmidrule(lr){3-4} \cmidrule(l){6-7} 
\begin{tabular}[c]{@{}l@{}}FixMatch \cite{sohn2020fixmatch}\\ \quad w/ CReST+\cite{wei2021crest}\\ \quad w/ DASO \cite{oh2022daso}\\ \quad w/ SimPro \cite{du2024simpro}\\ \quad w/ ACR \cite{wei2023towards}\\ \quad w/ Ours\end{tabular} &
   &
  \begin{tabular}[c]{@{}c@{}}\ms{71.7}{0.46}\\ \ms{71.4}{0.60}\\ \ms{73.1}{0.68}\\ \ms{84.8}{0.54}\\ \ms{83.2}{0.41}\\ \ms{\textbf{86.8}}{0.54}\end{tabular} &
  \begin{tabular}[c]{@{}c@{}}\ms{66.6}{0.87}\\ \ms{67.2}{0.48}\\ \ms{71.1}{0.32}\\ \ms{83.0}{0.36}\\ \ms{80.6}{0.30}\\ \ms{\textbf{83.1}}{0.04}\end{tabular} &
   &
  \begin{tabular}[c]{@{}c@{}}\ms{39.7}{0.61}\\ \ms{36.9}{0.57}\\ \ms{43.1}{1.20}\\ \ms{43.6}{0.35}\\ \ms{44.6}{0.25}\\ \ms{\textbf{45.7}}{0.34}\end{tabular} &
  \begin{tabular}[c]{@{}c@{}}\ms{38.2}{0.82}\\ \ms{35.1}{1.10}\\ \ms{43.8}{0.43}\\ \ms{44.8}{0.56}\\ \ms{42.3}{0.17}\\ \ms{\textbf{44.9}}{0.53}\end{tabular} \\ \bottomrule
\end{tabular}%
}
\label{tab:middle_headtail}
\end{table}

\begin{figure*}[t]
\centering
\subfloat[ACR for \textit{consistent}]{\includegraphics[width=2.in]{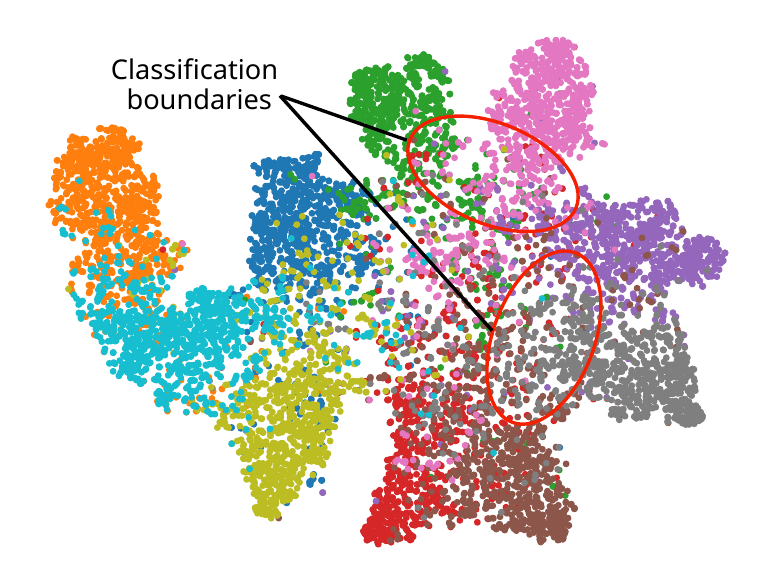}%
\label{fig:tsne1}}
\hfil
\subfloat[ACR for \textit{uniform}]{\includegraphics[width=2.0in]{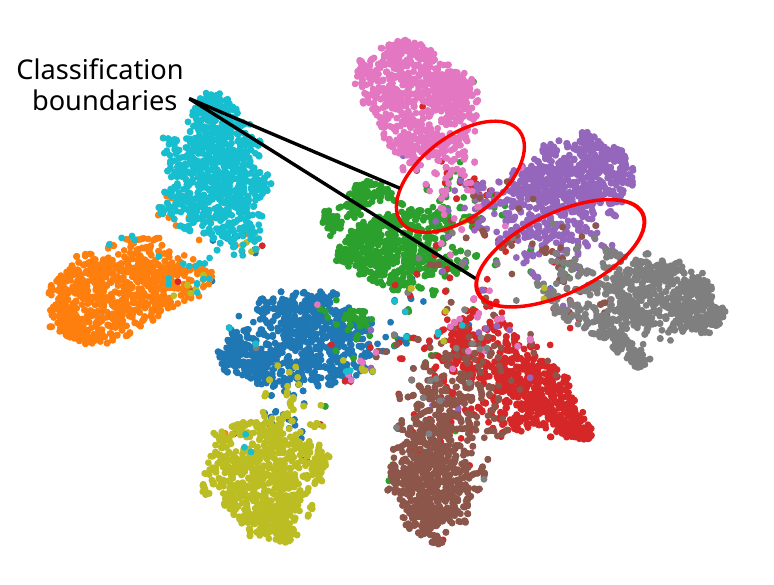}%
\label{fig:tsne2}}
\hfil
\subfloat[ACR for \textit{reversed}]{\includegraphics[width=2.0in]{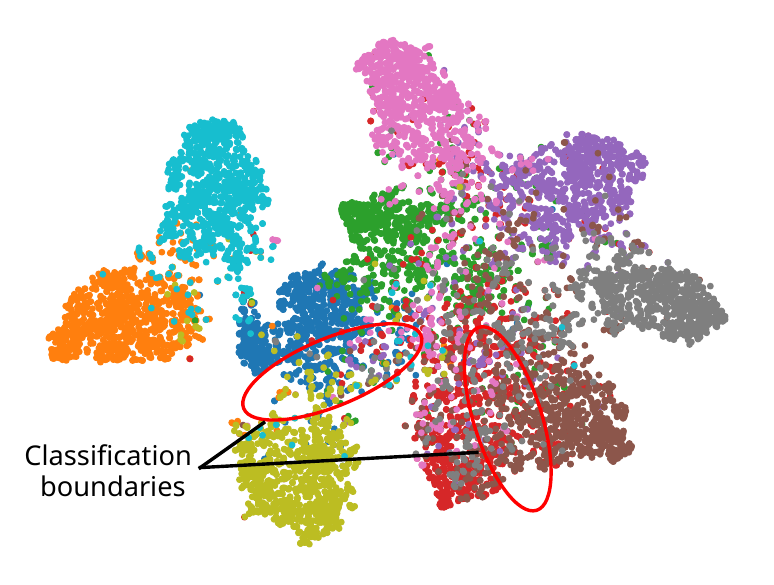}
\label{fig:tsne3}}

\medskip

\subfloat[\algo\ for \textit{consistent}]{\includegraphics[width=2.0in]{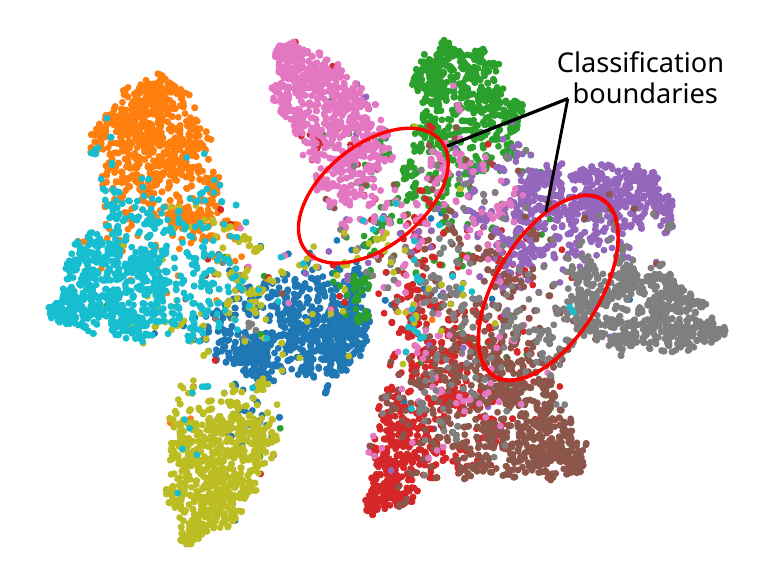}%
\label{fig:tsne4}}
\hfil
\subfloat[\algo\ for \textit{uniform}]{\includegraphics[width=2.0in]{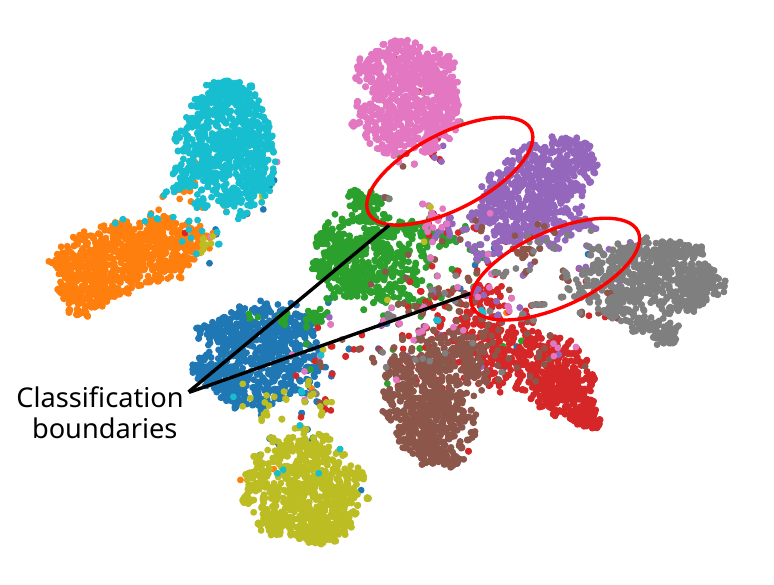}%
\label{fig:tsne5}}
\hfil
\subfloat[\algo\ for \textit{reversed}]{\includegraphics[width=2.0in]{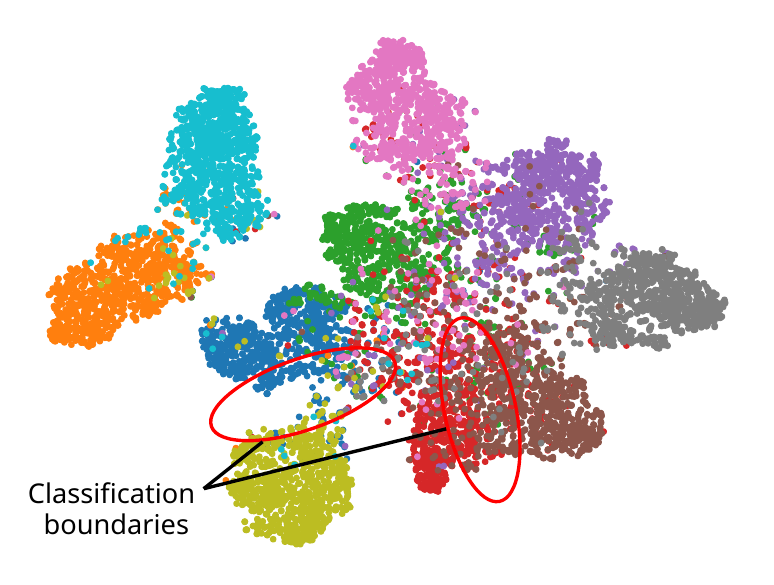}%
\label{fig:tsne6}}

\caption{The t-SNE visualization of the test set for ACR and \algo\ on CIFAR-10-LT with $N_1=500, M_1=4000$ and $\gamma_l=100$.}
\label{fig:tsne}
\end{figure*}

\subsection{Summary of Notations}

We provide a notation table to improve readability, which explicitly summarizes the key priors and temperature parameters used throughout the method. In particular, the summary highlights: (i) the labeled empirical prior used in the balanced branch; (ii) the EMA prior estimated by the balanced branch, which is used to align the standard branch; (iii) the EMA prior estimated by the standard branch, which is used to generate adjusted pseudo-labels for the balanced branch; and (iv) the roles of $\tau_{1}$ (pseudo-label adjustment strength), $\tau_{2}$ (alignment strength in the standard branch), and $\tau_{3}$ (inference-time post-hoc adjustment strength). The compact notation table is shown in \Cref{tab:notation_summary}.

\subsection{Results under more class distributions}
\label{exp:more_dist}

To examine the performance of our method in more imbalanced scenarios, we conduct additional experiments on CIFAR-10-LT (CIFAR-100-LT) by fixing $\gamma_l=100$ ($\gamma_l=20$) while varying the imbalance ratio of unlabeled data $\gamma_u$ from \textit{consistent} to \textit{reversed} by a step size 20 (step size 5). We set $N_1=500$ and $M_1=4000$ ($N_1=50$ and $M_1=400$) ($M_C=4000$ ($M_C=400$) in \textit{reversed} setting) and compare the performance with ACR\cite{wei2023towards}. \Cref{fig:more_settings} shows that our method \algo\ consistently outperforms ACR in all test cases. The performance gain becomes increasingly significant as the class distribution of unlabeled data changes from \textit{consistent} to \textit{reversed} on CIFAR-100-LT. In addition, for STL10-LT, we find that \algo\ can also achieve high performance under various imbalanced labeled ratio in \Cref{fig:more_stl10}. This demonstrates the capability of our method to handle various realistic LTSSL problems adaptively.

In addition, SimPro\cite{du2024simpro} introduces two more class distribution patterns for unlabeled data, i.e., \textit{middle} and \textit{headtail}, which are illustrated in \Cref{fig:dist_more}. The middle distribution represents a concentration of classes in the middle range of labeled data's classes, whereas the head-tail distribution indicates a concentration at both extremes. The results are reported in \Cref{tab:middle_headtail}, and we note that \algo\ achieves an average performance improvement of 1.1\% compared to SimPro, suggesting that our method can adapt to diverse distributions of unlabeled data without assuming any unlabeled distribution.

\subsection{Representation visualization}

We visualize the learned representations of \algo\ using the t-distributed stochastic neighbor embedding (t-SNE) \cite{van2008visualizing} and compare them with ACR. \Cref{fig:tsne} illustrates the comparison results on test set under \textit{consistent}, \textit{uniform}, and \textit{reversed} settings. It can be seen from the figure that the representations obtained by \algo\ allow for clearer classification boundaries, which can further enhance the model's performance in classification.

\subsection{Benefits for All Classes}

\begin{figure*}[t]
\centering
\includegraphics[width=1.3in]{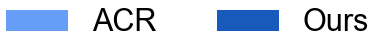}

\subfloat[$F_1$ score for \textit{consistent}]{\includegraphics[width=2.0in]{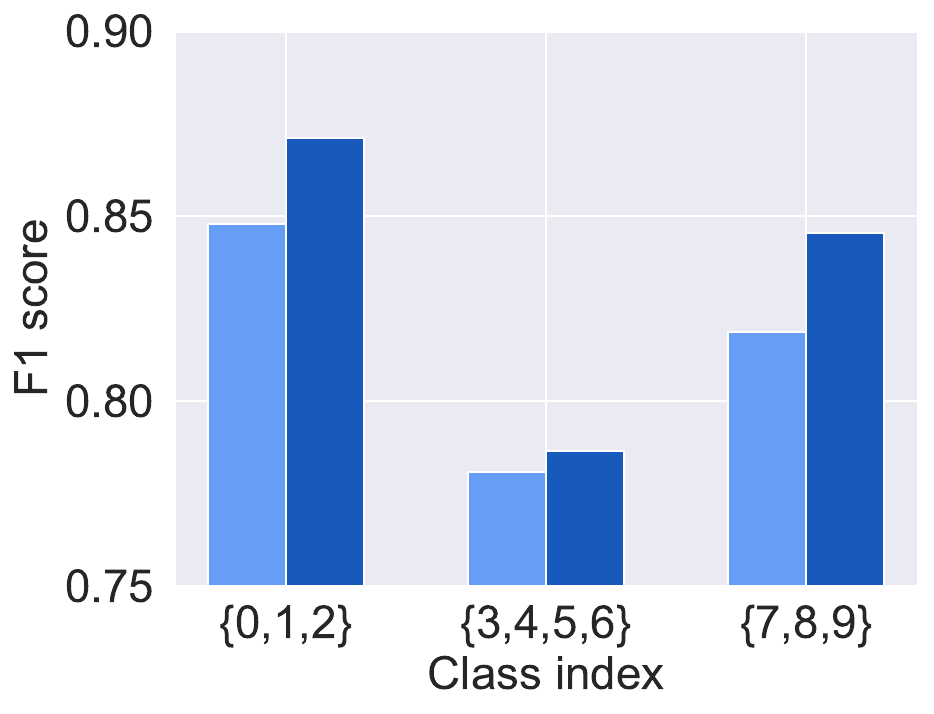}%
\label{fig:f1_con}}
\hfil
\subfloat[$F_1$ score for \textit{uniform}]{\includegraphics[width=2.0in]{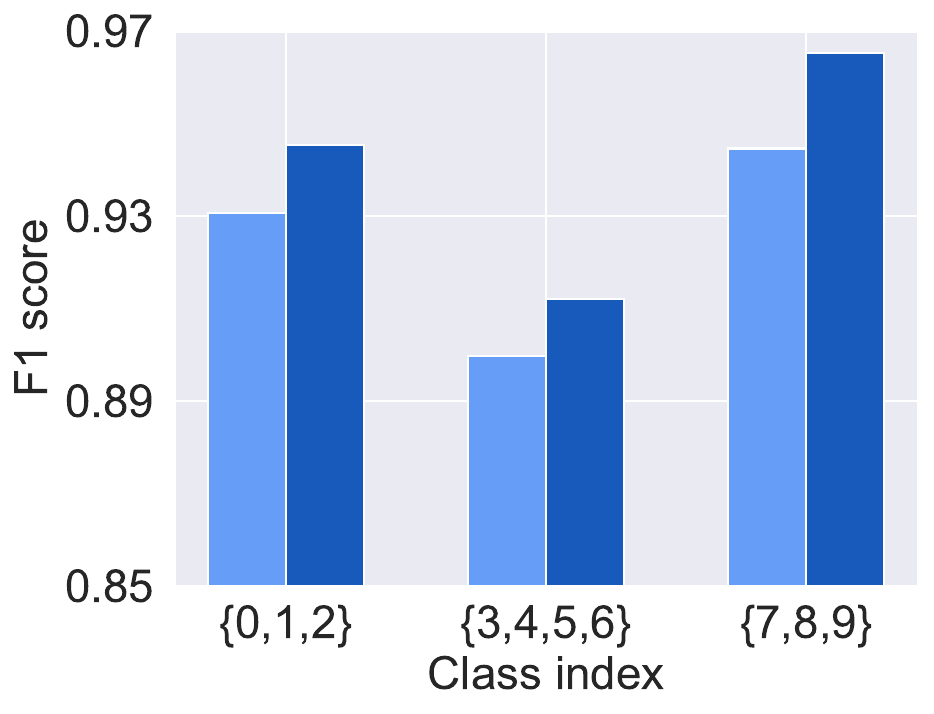}%
\label{fig:f1_uni}}
\hfil
\subfloat[$F_1$ score for \textit{reversed}]{\includegraphics[width=2.0in]{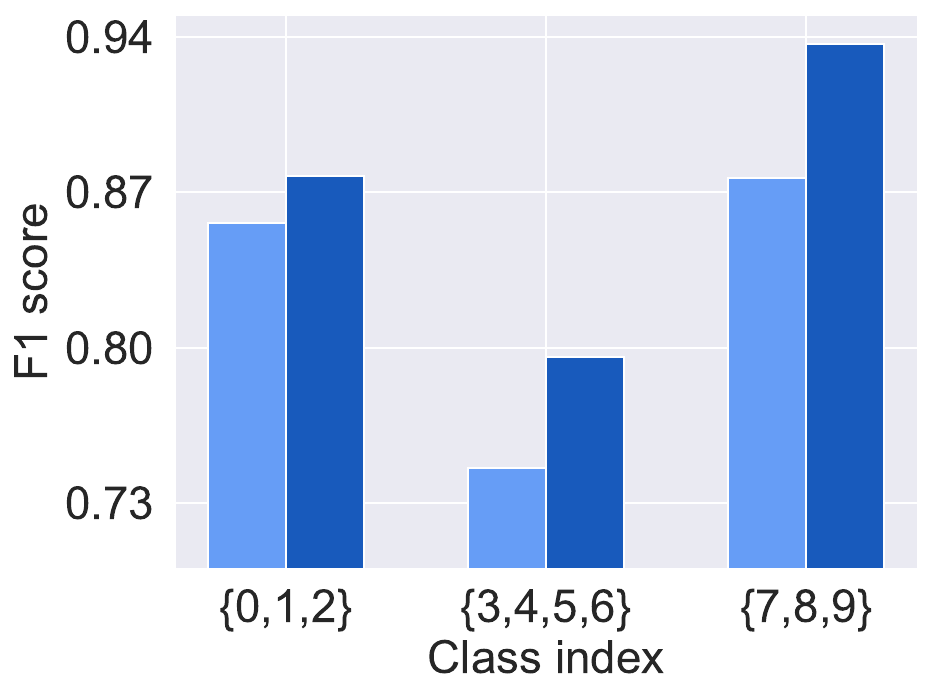}
\label{fig:f1_rev}}

\caption{The comparison of $F_1$ score for ACR and \algo\ on CIFAR-10-LT with $N_1=500, M_1=4000$ and $\gamma_l=100$.}
\label{fig:f1_score}
\end{figure*}

The two branches in \algo\ gradually converge during training, ultimately achieving strong performance across all classes. As illustrated in \Cref{fig:f1_score}, we observe that \algo\ can achieve higher $F_1$ score for various settings when compared to ACR. Specifically, whether in head or tail classes, \algo\ exhibits competitive $F_1$ scores, indicating its ability to improve tail-class performance while maintaining competitive head-class performance.

\subsection{Computational Overhead Analysis}
\label{supp:time}

It should be clarified that the ``dual-branch'' design in \algo\ does not involve two separate neural networks. Instead, it consists of a shared feature extractor (backbone) and two lightweight classifiers (heads).
Since the feature extractor (e.g., WideResNet-28-2) accounts for the vast majority of the parameters and computational operations, adding an auxiliary linear classification head (a single fully connected layer) introduces negligible overhead. To empirically demonstrate this, we compared the training speed and parameter count of \algo\ against the simple SSL baseline FixMatch under the exact same experimental settings (CIFAR-100-LT, WideResNet-28-2, Batch Size=64, single NVIDIA 3090 GPU). As shown in~\Cref{tab:re-time}, \algo\ only increases the parameter count about 0.01M and the training time per iteration by approximately 3.2\%. This slight increase is a worthy trade-off for the significant performance gain (+8.0\% accuracy).

\begin{table}[h]
\centering
  
\caption{Comparison of computational efficiency between FixMatch and \algo. The experiment is conducted on CIFAR-100-LT with WideResNet-28-2. \textit{Time/Iter} denotes the average time (milliseconds) for one training iteration (forward and backward pass).}
\resizebox{0.7\linewidth}{!}{
\begin{tabular}{lcccc}
\toprule
\textbf{Method} & \textbf{Params (M)} & \textbf{Time/Iter (ms)} & \textbf{Overhead} & \textbf{Accuracy (\%)} \\
\midrule
FixMatch & 1.47 & 124 & -- & 49.7 \\
\textbf{\algo}  & \textbf{1.48} & \textbf{128} & \textbf{+3.2\%} & \textbf{57.7} \\
\bottomrule
\end{tabular}
}
\label{tab:re-time}
  
\end{table}

\subsection{Experiments on Random Distributions}

To further substantiate our claim that \algo\ can handle a wide range of unlabeled distributions, we conducted additional experiments using 3 \textbf{Randomly Sampled} unlabeled distributions (sampled via Dirichlet distribution with $\alpha \in \{0.5, 1.0\}$). These distributions are highly irregular and jagged. The results in \Cref{tab:re-arbit-dist} show that while ACR struggles with these unstructured distributions, \algo\ maintains superior performance. This confirms that our Decouple-then-Converge mechanism effectively handles diverse unlabeled class priors under label-prior shift.

\begin{table}[h!]
\centering
\caption{Comparison of test accuracy (\%) on CIFAR-100-LT ($\gamma_l = 20$, $N_1=50$, $M_1=400$) with randomly generated unlabeled distributions. The results show that \algo\ consistently outperforms other methods, demonstrating superior robustness to varying label distributions.}
\label{tab:random_dist}
\resizebox{0.8\linewidth}{!}{
\begin{tabular}{lcccc}
\toprule
\textbf{Method} & \textbf{Random Dist. 1} & \textbf{Random Dist. 2} & \textbf{Random Dist. 3} & \textbf{Average} \\
\midrule
ACR & 43.1 $\pm$ 0.45 & 44.2 $\pm$ 0.21 & 43.5 $\pm$ 0.53 & 43.6 $\pm$ 0.40 \\
SimPro & 44.9 $\pm$ 0.20 & 45.8 $\pm$ 0.36 & 45.2 $\pm$ 0.45 & 45.3 $\pm$ 0.30 \\
\textbf{\algo} & \textbf{45.3 $\pm$ 0.28} & \textbf{46.1 $\pm$ 0.35} & \textbf{45.8 $\pm$ 0.22} & \textbf{45.7 $\pm$ 0.28} \\
\bottomrule
\end{tabular}
}
\label{tab:re-arbit-dist}
\end{table}

\subsection{Fine-Grained Ablations based on ACR}

To explicitly disentangle the contribution of each new component in \algo\ over our conference version (ACR), we conduct a fine-grained incremental ablation study. Starting from the ACR baseline, we add each key module of \algo\ one by one. The primary enhancements of \algo\ over ACR are:

\begin{itemize}
    \item \textbf{Cross-Branch Pseudo-Labeling (CBPL):} This module replaces ACR's strategy of using the labeled data distribution for adjustment. Instead, it generates pseudo-labels for the balanced branch by correcting the standard branch's outputs using the EMA-estimated unlabeled class prior from the standard branch ($\pi^{s}_t$) (Eq.~(7) in the main text). This highlights the cross-branch interaction.
    \item \textbf{Decoupled Distribution Alignment (DDA):} This is the interaction in the reverse direction. The standard branch's loss on labeled data is modified to align its predictions towards the decoupled distribution ($\tilde{\pi}_t$) estimated by the balanced branch (Eq.~(4) in the main text), mitigating confirmation bias.
    \item \textbf{Importance Weighting (IW):} \algo\ introduces a sample weighting scheme (Eq.~(8) and Eq.~(9) in the main text) for the consistency loss of the balanced branch, which measures the stability of pseudo-labels post-correction rather than relying on a simple confidence threshold.
\end{itemize}

We present the results of this incremental analysis on the CIFAR-10-LT and CIFAR-100-LT datasets in \Cref{tab:ablation_incremental}. We start with the ACR baseline and successively integrate CBPL, DDA, and IW to build up to the full \algo\ (before post-hoc adjustment). The results clearly demonstrate where the incremental gains originate.
\begin{itemize}
    \item Adding \textbf{CBPL} brings consistent improvements in the ``Uni'' and ``Rev'' settings, where the unlabeled distribution mismatches the labeled one. This supports the benefit of using a dynamically estimated unlabeled prior for pseudo-label correction.
    \item Further adding \textbf{DDA} continues this trend, especially in the ``Uni'' setting, by fostering beneficial interaction between the two branches and preventing the standard branch from overfitting to the (potentially misleading) labeled distribution.
    \item The introduction of \textbf{IW} provides a final small boost before post-hoc adjustment.
\end{itemize}

\begin{table*}[!t]
\begin{center}
\caption{Fine-grained ablation study showing the incremental performance gain from ACR to \algo. All results are reported before the final post-hoc adjustment step to isolate the contributions during training. The final row shows the full \algo\ model with post-hoc adjustment for reference. ``Con'', ``Uni'', and ``Rev'' denote consistent, uniform, and reversed distributions, respectively. Best results for each setting are in \textbf{bold}.}
\label{tab:ablation_incremental}
\resizebox{0.95\textwidth}{!}{%
\begin{tabular}{@{}lcccccc@{}}
\toprule
\multirow{2}{*}{Method} & \multicolumn{3}{c}{CIFAR-10-LT ($\gamma_l=100$)} & \multicolumn{3}{c}{CIFAR-100-LT ($\gamma_l=10$)} \\ \cmidrule(l){2-4} \cmidrule(l){5-7} 
                        & Con.       & Uni.       & Rev.       & Con.       & Uni.       & Rev.       \\ \midrule
ACR (Baseline)         & 81.6         & 92.1         & 84.9         & 51.3         & 57.2         & 51.6         \\
+ CBPL                   & 78.1         & 92.8         & 85.3         & 51.2         & 58.0         & 52.1         \\
+ CBPL + DDA              & 78.4         & 93.5         & 85.4         & 51.4         & 58.9         & 52.6         \\
+ CBPL + DDA + IW (\algo\ w/o post-hoc) & 77.8 & 93.8 & 85.5 & 51.1 & 59.6 & 52.7 \\
\midrule
\textbf{\algo\ (Full Model, with post-hoc)} & \textbf{84.1} & \textbf{93.9} & \textbf{86.6} & \textbf{52.0} & \textbf{60.4} & \textbf{53.3} \\
\bottomrule
\end{tabular}%
}
\end{center}
\end{table*}

\noindent Interestingly, in the ``Con'' setting, combining these modules decreases the pre-adjustment accuracy (e.g., from 81.6\% to 77.8\% on CIFAR-10-LT). However, as shown in Table~5 in the main paper, these components substantially improve the model's \emph{amenability to post-hoc rebalancing}. Concretely, under the same post-hoc protocol, the full \algo\ model gains a substantial \textbf{+6.3\%} on CIFAR-10-LT (``Con'') after post-hoc adjustment (77.8\% $\rightarrow$ 84.1\%), whereas ACR improves by only +0.8\%. This indicates that the added modules encourage learning representations and decision scores that are easier to calibrate for balanced inference, ultimately leading to better final performance.

Taken together, the ablations indicate that the gains of \algo\ over ACR stem from a synergistic combination of more adaptive pseudo-label correction, cross-branch knowledge transfer, and a more robust training process, which collectively improve the effectiveness of the final inference-time rebalancing.


\begin{table*}[ht]
\centering
\caption{The results are based on fine-tuning the pre-trained CLIP model. The fine-tuning strategies employed in the experiments are visual prompt tuning (VPT) \cite{jia2022visual} and AdaptFormer \cite{chen2022adaptformer}. We set $N_1=500, M_1=4000, \gamma_l=100$ for CIFAR-10-LT, $N_1=50, M_1=400, \gamma_l=10$ for CIFAR-100-LT, and $N_1=150$ for STL10-LT.}
\resizebox{0.95\linewidth}{!}{%
\begin{tabular}{@{}lcccccccc@{}}
\toprule
\multirow{2}{*}{Algorithm} & \multicolumn{3}{c}{CIFAR-10-LT} & \multicolumn{3}{c}{CIFAR-100-LT} & \multicolumn{2}{c}{STL10-LT}  \\ \cmidrule(l){2-9} 
                           & Con       & Uni      & Rev      & Con       & Uni      & Rev      & $\gamma_l=10$ & $\gamma_l=20$ \\ \midrule
\begin{tabular}[c]{@{}l@{}}FixMatch + VPT \cite{jia2022visual}\\ \quad w/ ACR \cite{wei2023towards}\\ \quad w/ Ours\end{tabular} &
  \begin{tabular}[c]{@{}c@{}}\ms{92.8}{0.22}\\ \ms{95.5}{0.54}\\ \ms{96.0}{0.30}\end{tabular} &
  \begin{tabular}[c]{@{}c@{}}\ms{95.1}{0.11}\\ \ms{97.1}{0.05}\\ \ms{97.3}{0.29}\end{tabular} &
  \begin{tabular}[c]{@{}c@{}}\ms{95.2}{0.65}\\ \ms{96.2}{0.42}\\ \ms{96.7}{0.34}\end{tabular} &
  \begin{tabular}[c]{@{}c@{}}\ms{78.6}{0.65}\\ \ms{80.8}{0.20}\\ \ms{81.3}{0.21}\end{tabular} &
  \begin{tabular}[c]{@{}c@{}}\ms{81.3}{0.44}\\ \ms{81.2}{0.23}\\ \ms{82.9}{0.15}\end{tabular} &
  \begin{tabular}[c]{@{}c@{}}\ms{79.7}{0.21}\\ \ms{80.3}{0.34}\\ \ms{81.9}{0.23}\end{tabular} &
  \begin{tabular}[c]{@{}c@{}}\ms{98.4}{0.04}\\ \ms{98.2}{0.14}\\ \ms{98.7}{0.25}\end{tabular} &
  \begin{tabular}[c]{@{}c@{}}\ms{97.3}{0.42}\\ \ms{97.8}{0.78}\\ \ms{98.5}{0.23}\end{tabular} \\ \midrule
\begin{tabular}[c]{@{}l@{}}FixMatch + AdaptFormer \cite{chen2022adaptformer}\\ \quad w/ ACR \cite{wei2023towards}\\ \quad w/ Ours\end{tabular} &
  \begin{tabular}[c]{@{}c@{}}\ms{92.7}{0.23}\\ \ms{96.1}{0.33}\\ \ms{96.7}{0.41}\end{tabular} &
  \begin{tabular}[c]{@{}c@{}}\ms{95.6}{0.15}\\ \ms{97.6}{0.23}\\ \ms{97.8}{0.25}\end{tabular} &
  \begin{tabular}[c]{@{}c@{}}\ms{95.8}{0.11}\\ \ms{96.2}{0.76}\\ \ms{96.8}{0.34}\end{tabular} &
  \begin{tabular}[c]{@{}c@{}}\ms{79.7}{0.45}\\ \ms{81.7}{0.56}\\ \ms{82.2}{0.13}\end{tabular} &
  \begin{tabular}[c]{@{}c@{}}\ms{81.7}{0.23}\\ \ms{83.3}{0.65}\\ \ms{83.6}{0.32}\end{tabular} &
  \begin{tabular}[c]{@{}c@{}}\ms{80.2}{0.45}\\ \ms{81.4}{0.38}\\ \ms{83.1}{0.19}\end{tabular} &
  \begin{tabular}[c]{@{}c@{}}\ms{98.4}{0.30}\\ \ms{98.6}{0.52}\\ \ms{98.6}{0.56}\end{tabular} &
  \begin{tabular}[c]{@{}c@{}}\ms{98.3}{0.23}\\ \ms{98.4}{0.45}\\ \ms{98.9}{0.20}\end{tabular} \\ \bottomrule
\end{tabular}%
}
\label{tab:peft}
\end{table*}

\section{The Potential Applications in Few-Shot Learning and Future Work}

Few-shot learning shares a common spirit with LTSSL: \textbf{leveraging unlabeled data through pseudo-labeling to overcome data scarcity}. We discuss several connections and future directions below:

\begin{itemize}
    \item \textbf{Complementing the Focus on Pseudo-Label Quality (TPAMI 2025~\cite{ouyang2025semantic}):}
    This work studies the issues of ``clustering noise'' and ``semantic chaos'' in pseudo-labels for unsupervised meta-learning, aiming to improve the \textit{semantic coherence} of pseudo-labeled clusters.
    Our method, \algo, addresses a complementary challenge: handling an \textbf{unknown and potentially long-tailed class prior} in the auxiliary unlabeled data.
    In practical FSL scenarios with auxiliary unlabeled data, class frequencies are rarely balanced. Even if clusters are semantically clean, a strongly imbalanced auxiliary set can bias the meta-training tasks constructed from it.
    One possible extension is to integrate \algo's distribution-aware dual-branch framework with their semantic-stability index, so that pseudo-labels are both semantically coherent and less affected by distributional bias.

    \item \textbf{An Alternative to Two-Stage Frameworks (TIP 2024~\cite{dong2024pseudo}):}
    The work proposes a practical two-stage framework (PLML) for Semi-Supervised FSL: (1) pre-train a model using a standard SSL algorithm, and (2) generate pseudo-labels and perform meta-training.
    \algo\ offers an alternative as an \textbf{end-to-end, single-stage framework}: its standard and balanced branches interact throughout training, allowing feature representation and pseudo-label correction to be jointly refined.
    Exploring \algo\ as a replacement for the first stage of PLML, or as a single-stage alternative, is a compelling direction for future research.

    \item \textbf{Addressing Distributional Bias in Clustering-Based Meta-Learning (ECCV 2022~\cite{dong2022rethinking}):}
    The work focuses on the quality of the feature space, proposing a clustering-friendly embedding (PL-CFE) to minimize intra-class distance and maximize inter-class discriminability.
    However, their task construction phase relies on standard clustering (e.g., $k$-means), which can be sensitive to class imbalance.
    In realistic scenarios with long-tailed auxiliary unlabeled data, $k$-means may be biased towards majority classes, potentially merging tail classes into head clusters despite a good embedding.
    \algo\ provides a complementary tool by handling \textbf{unknown and mismatched unlabeled class priors}.
    A direction for future work is to incorporate \algo's distribution estimation and logit correction into the clustering phase, improving robustness to class-frequency mismatch when synthesizing few-shot tasks.
\end{itemize}

In summary, while the cited works emphasize semantic purity and practical training strategies for FSL with unlabeled data, \algo\ contributes a complementary capability: \textbf{robustness to unknown mismatches in the unlabeled class prior (label shift)}.

\section{Practical Relevance and Applicability of Distribution Mismatch in Real-World Semi-Supervised Learning}

We briefly motivate why the distribution mismatch setting is practically relevant.

First, the mismatch between labeled and unlabeled \emph{class frequencies} is not a theoretical artifact but a common reality in large-scale applications. For example, labeled data are often curated under specific protocols (e.g., medical annotations from hospitals), while unlabeled data are collected in-the-wild (e.g., user-uploaded images or sensor logs), naturally leading to divergent class priors—even when sourced from a similar domain.

Second, generative models may produce synthetic data whose class composition deviates from real data. This aligns with our setting: the unlabeled data (real or synthetic) may have an \emph{unknown} and mismatched class prior. Our method \algo\ does not require knowing the unlabeled class prior in advance and is designed to adapt to such mismatches.

Overall, the problem we address is realistic and increasingly important as SSL scales to large, heterogeneous data collection pipelines.

\section{More Clarification for \algo's Novelty}

While both logit adjustment and dual-branch architectures have appeared in prior work, the novelty of \algo\ lies in how these components are integrated to address the unknown unlabeled class prior---a setting where existing methods can fail.

Specifically, unlike ACR or other two-branch approaches that rely on fixed anchor distributions or static pseudo-labeling, \algo\ introduces cross-branch interaction mechanisms that enable the standard and balanced branches to dynamically converge during training. The balanced branch does not operate in isolation; instead, it receives refined pseudo-labels from the standard branch (via adaptive logit correction based on an estimated unlabeled prior), while the standard branch aligns its predictions using the decoupled distribution inferred from the balanced branch (Eq.~(4)--(6) in the main text). This mutual bootstrapping---decouple then converge---is absent in prior works and is key to handling unknown unlabeled class-prior shifts.

Moreover, as shown in Sec.~3.5 and Tables~1--4 in the main paper, \algo\ consistently outperforms ACR and other two-branch methods under diverse unlabeled class priors, confirming that our design goes beyond a simple combination of existing techniques.

\section{Diagram for \algo}

\begin{figure}[h]
\centering
\includegraphics[width=0.8\linewidth]{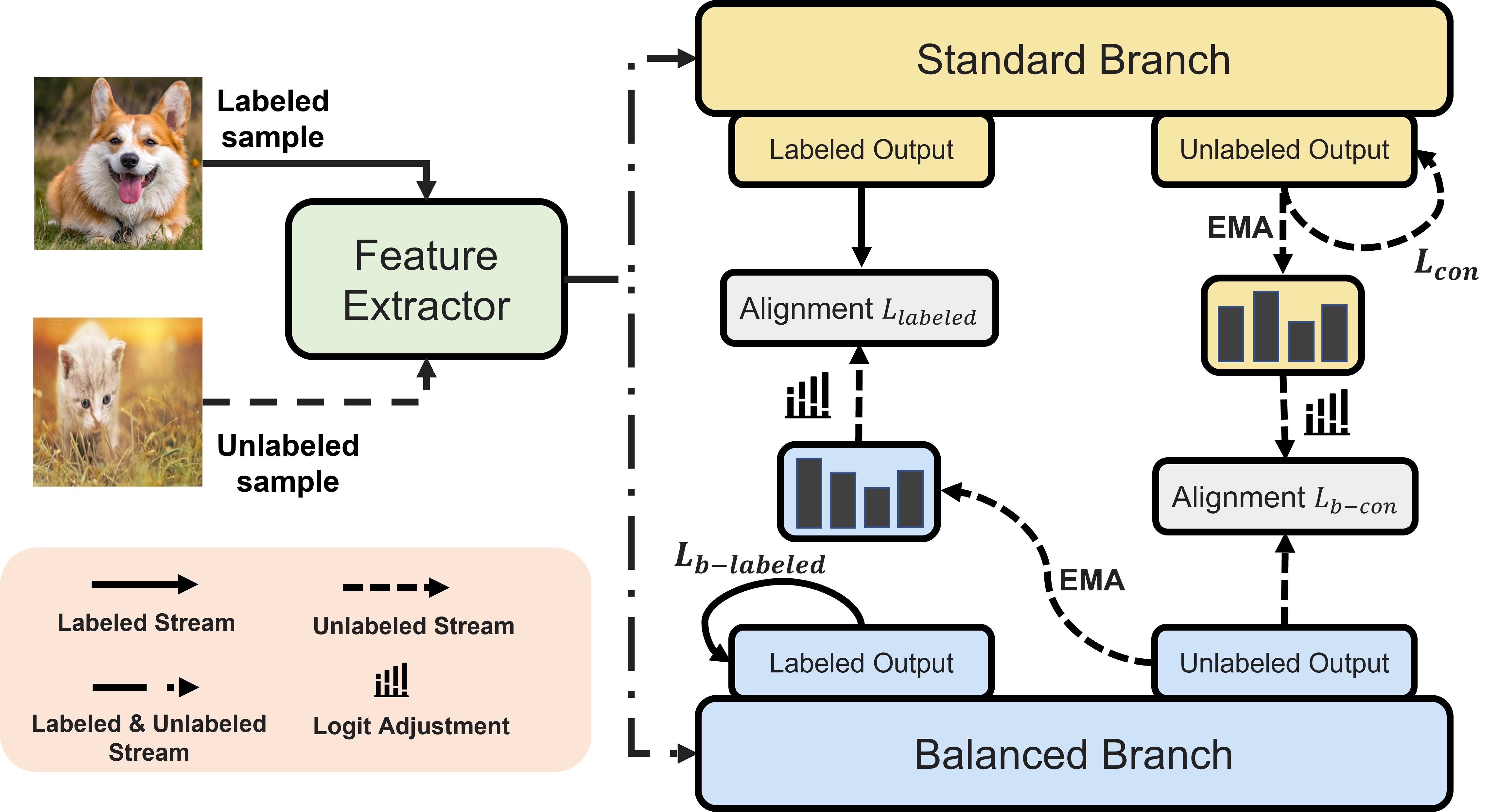}
\caption{Schematic overview of the proposed \algo\ framework.}
\label{fig:framework}
\end{figure}

\Cref{fig:framework} provides a schematic overview of \algo, including:

\begin{itemize}
\item the dual-branch architecture (standard and balanced branches),
\item cross-branch interactions (pseudo-label transfer with logit correction and distribution alignment).
\end{itemize}

\section{Looking ahead: future LTSSL methods}

While \algo\ has achieved outstanding performance, we have observed that recent popular foundation models \cite{radford2021learning} can achieve excellent performance on downstream tasks through fine-tuning \cite{liu2022few,shi2023parameter,yang2022parameter,gan2024erasing}. Therefore, we investigate whether fine-tuning foundation models can further enhance the performance of the LTSSL method. We choose CLIP \cite{radford2021learning} for fine-tuning due to its strong generalization ability. CLIP is a multimodal model to align images and text in a shared embedding space, facilitating tasks like image classification, generation, and zero-shot learning across various domains \cite{shi2023parameter}. When considering full fine-tuning the model, the target tasks often require a substantial amount of data for effective adaptation \cite{dosovitskiy2020image,chen2022revisiting}. However, for LTSSL tasks, labeled data are scarce and severely imbalanced, hindering full fine-tuning from achieving satisfactory performance. Therefore, motivated by \cite{shi2023parameter}, we attempt to leverage the parameter-efficient fine-tuning (PEFT) strategy \cite{jia2022visual,chen2022adaptformer,hu2021lora} for fine-tuning to enhance the generalization performance of LTSSL models. PEFT strategies typically freeze the pre-trained model while introducing a few learnable parameters for adaptation. The results are shown in \Cref{tab:peft}, and we utilize Visual Prompt Tuning (VPT) \cite{jia2022visual} and AdaptFormer \cite{chen2022adaptformer} for fine-tuning. VPT prepends learnable prompts optimized during the training at each transformer layer. AdaptFormer adds a bottleneck module, which is parallel to the FFN layer. The incorporation of VPT and AdaptFormer facilitates a notable 17.6\% and 18.2\% performance improvement for \algo\ compared to the Wide ResNet-based model \cite{zagoruyko2016wide}. Additionally, we find that \algo\ outperforms ACR by 0.6\% when employing PEFT on CIFAR-10-LT and CIFAR-100-LT, further demonstrating the broad applicability of \algo. Regarding the STL10-LT, we observe that the performance of all methods has almost saturated, with error rates generally below 2\%. We advocate PEFT as a promising direction for LTSSL with foundation models in the future.



Next, we further clarify the relationship between \algo\ and CLIP, as well as practical challenges in this setting.

\paragraph{Complementarity between CLIP and \algo}
CLIP provides strong pre-trained representations that can improve feature quality, but it does not inherently solve classifier bias caused by long-tailed labeled data in the target task.
\begin{itemize}
\item \textbf{Representation vs.\ decision boundary:} CLIP mainly improves the feature space (representation learning). However, when fine-tuned (including via PEFT) on a highly imbalanced labeled set, the resulting decision boundaries can still skew towards head classes. \algo\ addresses this issue at the objective/optimization level through cross-branch interactions and distribution-aware correction.
\item \textbf{Empirical evidence:} As shown in \Cref{tab:peft}, applying \algo\ on top of CLIP (via VPT or AdaptFormer) yields consistent improvements (e.g., surpassing ACR by 0.6\% and FixMatch by 1.9\%). This suggests that \algo\ remains beneficial even with strong foundation-model representations.
\end{itemize}

\paragraph{Practical challenges when using CLIP for LTSSL}
Integrating foundation models into LTSSL is non-trivial and faces several challenges beyond standard fine-tuning:
\begin{itemize}
\item \textbf{Biased adaptation with scarce, imbalanced labels:} Adapting a generic model to a specific task with extremely scarce and long-tailed labeled data can quickly introduce bias towards head classes.
\item \textbf{Noisy pseudo-labeling under imbalance:} In SSL, the model generates pseudo-labels for unlabeled data. If adaptation is biased, pseudo-labels can be skewed and reinforce the imbalance. A key challenge is how to leverage unlabeled data---with an unknown class prior---to regularize adaptation, which is aligned with the motivation of \algo's dual-branch interaction.
\end{itemize}

\section{Theoretical Analysis for \algo}
\label{sec:th_ana}

\subsection{Setup and Notation}

We consider a labeled set $\mathcal{D}^l=\{(x^{(l)}_i,y^{(l)}_i)\}_{i=1}^{N}$ and an unlabeled set
$\mathcal{D}^u=\{x^{(u)}_j\}_{j=1}^{M}$ with $C$ classes.
Let $\pi_l\in\Delta^{C-1}$ be the labeled class prior and $\pi_u\in\Delta^{C-1}$ be the (unknown) unlabeled/test prior.
We write $[C]:=\{1,2,\ldots,C\}$ and $\Delta^{C-1}$ for the probability simplex.


\paragraph{Two branches and two EMA priors}
Let $f_t$ denote \emph{standard-branch} logits and $\tilde f_t$ denote \emph{balanced-branch} logits at step $t$.
The method maintains two EMA priors:
(i) $\tilde\pi_t$ estimated from the balanced branch (used in decoupled distribution alignment, Eq.~(4) in the main text);
(ii) $\pi^{s}_t$ estimated from the standard branch (used in pseudo-label correction, Eq.~(7) in the main text).
Let $\delta(\cdot)$ be the softmax, so $\delta(z)\in\mathrm{int}(\Delta^{C-1})$ for any finite logits $z$.

\paragraph{Temperatures}
$\tau_{2}>0$ is the scale used in standard-branch decoupled distribution alignment (Eq.~(4)).
$\tau_{1}>0$ is the scale used in pseudo-label correction (Eq.~(7)).
$\tau_{3}>0$ is the post-hoc adjustment scale used at inference (Eq.~(11)).

\begin{assumption}[Label shift]
\label{ass:label_shift}
There exists a shared class-conditional distribution $p(x\mid y)$ such that
$p_l(x,y)=\pi_l(y)p(x\mid y)$ and $p_u(x,y)=\pi_u(y)p(x\mid y)$.
\end{assumption}

\paragraph{Positivity for log terms}
All priors used inside $\log(\cdot)$ are strictly positive.
Softmax outputs are strictly positive; EMA priors remain in the simplex interior by convexity.
Implementations may clip probabilities for numerical stability.

\subsection{Decouple: Balanced Softmax Recovers Class-Conditional Scores in an Idealized Limit}

For any prior $\pi\in\Delta^{C-1}$ define the \emph{prior-weighted} predictive distribution
\[
p_{\tilde f}^{\pi}(y\mid x)
:=\frac{\exp(\tilde f_y(x))\pi(y)}{\sum_{c=1}^C \exp(\tilde f_c(x))\pi(c)}.
\]

\begin{proposition}[Labeled-only decoupling property]
\label{prop:bs_recover}
Assume Assumption~\ref{ass:label_shift} and realizability:
there exists $\tilde f$ such that $p_{\tilde f}^{\pi_l}(\cdot\mid x)=p_l(\cdot\mid x)$ almost surely.
Then any population minimizer $\tilde f^{\star}$ of
$\mathbb{E}_{(x,y)\sim p_l}\big[-\log p_{\tilde f}^{\pi_l}(y\mid x)\big]$ satisfies
\[
\tilde f^{\star}_y(x)=\log p(x\mid y)+b(x),
\]
for some measurable $b(x)$ independent of $y$.
\end{proposition}

\begin{proposition}[Decoupled scores imply \emph{prior-adaptable} Bayes prediction]
\label{prop:any_prior}
Assume Assumption~\ref{ass:label_shift}. Fix any target prior $\pi_{\mathrm{tar}}\in\Delta^{C-1}$.
If a model $\tilde f$ satisfies $\tilde f_y(x)=\log p(x\mid y)+b(x)$ for some $b(x)$ independent of $y$, then
\[
\arg\max_{y\in[C]}\{\tilde f_y(x)+\log \pi_{\mathrm{tar}}(y)\}
=
\arg\max_{y\in[C]}\{\log p(x\mid y)+\log \pi_{\mathrm{tar}}(y)\},
\]
which is Bayes-optimal under label shift with target prior $\pi_{\mathrm{tar}}$.
\end{proposition}


\subsection{EMA Prior Estimators: Deterministic Dynamics (Mean-Field) and Bounded Variation}

\paragraph{Mean-field vs. mini-batch.}
In the implementation, priors are updated using mini-batch averages.
For a careful and non-misleading convergence statement, we analyze the mean-field (population) update:
\[
\bar p_t := \mathbb{E}_{x\sim p_u}\big[\delta(s_t(x))\big]\in\Delta^{C-1},
\]
where $s_t$ is either $f_t$ (for $\pi^{s}_t$) or $\tilde f_t$ (for $\tilde\pi_t$).
Mini-batch updates are unbiased estimators of $\bar p_t$ and EMA smoothing controls their variance;
we do not claim almost-sure pointwise convergence for the stochastic EMA with persistent sampling noise.

\paragraph{Generic EMA recursion.}
Let $\pi_t$ denote any EMA prior updated by
\begin{equation}
\label{eq:ema_update_generic}
\pi_{t+1}=m\pi_t+(1-m)\bar p_t,\qquad m\in(0,1).
\end{equation}

\begin{lemma}[Closed form and bounded step size]
\label{lem:ema_closed}
The EMA recursion \eqref{eq:ema_update_generic} admits the closed form
\[
\pi_t
=
m^t\pi_0+(1-m)\sum_{k=0}^{t-1}m^{t-1-k}\bar p_k.
\]
Moreover,
\[
\|\pi_{t+1}-\pi_t\|_1
=(1-m)\|\bar p_t-\pi_t\|_1\le 2(1-m).
\]
\end{lemma}

\begin{lemma}[Tracking bound and convergence under convergent drive]
\label{lem:ema_track}
Define $e_t:=\pi_t-\bar p_t$. Then for all $t\ge 1$,
\begin{equation}
\label{eq:ema_track_exact}
\|e_t\|_1
\le
m^t\|e_0\|_1+\sum_{k=1}^{t} m^{t-k}\|\bar p_k-\bar p_{k-1}\|_1.
\end{equation}
In particular, if $\bar p_t\to \bar p_\infty$, then $\pi_t\to \bar p_\infty$.
\end{lemma}

\subsection{What Prediction-Based Prior Estimation Can (and Cannot) Guarantee}

\paragraph{Balanced-branch prediction marginal}
Define
\[
\pi_t^{\mathrm{pred}}
:=\mathbb{E}_{x\sim p_u}\big[\delta(\tilde f_t(x))\big]\in\Delta^{C-1},
\qquad
\hat\pi_t:=\frac{1}{M}\sum_{j=1}^{M}\delta(\tilde f_t(x^{(u)}_j)).
\]

\begin{theorem}[Error decomposition for prediction-marginal estimation]
\label{thm:piA_decomp}
Fix $t$ and treat $\tilde f_t$ as fixed. With probability at least $1-\nu$ over $\mathcal{D}^u$ (i.i.d.\ from $p_u(x)$),
\[
\|\hat\pi_t-\pi_u\|_1
\le
\underbrace{\|\pi_t^{\mathrm{pred}}-\pi_u\|_1}_{\text{model-induced bias}}
+
2\sqrt{\frac{C}{M}}+\sqrt{\frac{2\log(1/\nu)}{M}}.
\]
Moreover,
\[
\|\pi_t^{\mathrm{pred}}-\pi_u\|_1
\le
\mathbb{E}_{x\sim p_u}\big\|\delta(\tilde f_t(x))-p_u(\cdot\mid x)\big\|_1.
\]
\end{theorem}

\paragraph{Key takeaway}
Theorem~\ref{thm:piA_decomp} \emph{does not} claim that $\hat\pi_t$ is close to $\pi_u$.
It cleanly separates two effects: sampling noise (controlled by $M$) and model-induced bias (depends on prediction quality).
No prediction-only estimator can generally avoid this dependence.
Therefore, our algorithmic justification \emph{does not rely on} assuming $\tilde\pi_t$ or $\pi^{s}_t$ is a universally consistent estimator of $\pi_u$.

\begin{proposition}[Uniform-prior posterior effect]
\label{prop:unif_posterior}
Assume $\tilde f$ satisfies the labeled-only decoupling form in Proposition~\ref{prop:bs_recover}. Then
\[
\delta(\tilde f(x))_y=\frac{p(x\mid y)}{\sum_{c=1}^C p(x\mid c)}=:p^{\mathrm{unif}}(y\mid x),
\]
i.e., the posterior induced by a uniform class prior.
Consequently, in the large-sample limit, $\hat\pi$ converges to
\[
\pi^{\mathrm{unif}}:=\mathbb{E}_{x\sim p_u}\big[p^{\mathrm{unif}}(\cdot\mid x)\big],
\]
which may differ from $\pi_u$ when classes overlap.
\end{proposition}

\paragraph{Why Proposition~\ref{prop:unif_posterior} is \emph{not} a contradiction to \algo}
Proposition~\ref{prop:unif_posterior} shows only that a perfectly decoupled model produces a \emph{uniform-prior posterior}.
It does \emph{not} imply that the induced marginal $\pi^{\mathrm{unif}}$ must be the uniform vector,
nor does it invalidate the algorithm, because \algo\ does not require $\tilde\pi_t=\pi_u$ to be exact.
Instead, \algo\ uses prior signals through (i) margin-robust logit adjustments, and (ii) a self-paced importance weight
that suppresses DA/PLC-sensitive pseudo-labels.

\subsection{Pseudo-Label Correction (PLC) and Self-Paced Importance Weighting (Eq.~(7--9) in Main Text)}

\paragraph{PLC distribution.}
Let $p^{\mathrm{pre}}_t(\cdot\mid x):=\delta(f_t(x))$ and define
\begin{equation}
\label{eq:plc_def}
p^{\mathrm{plc}}_{t}(y\mid x)
=
\frac{p^{\mathrm{pre}}_t(y\mid x)\cdot \big(\pi^{s}_t(y)\big)^{-\tau_{1}}}
{\sum_{c=1}^C p^{\mathrm{pre}}_t(c\mid x)\cdot \big(\pi^{s}_t(c)\big)^{-\tau_{1}}}.
\end{equation}
The pseudo-label used to supervise the balanced branch is
$\hat q^{\mathrm{plc}}_t(x):=\arg\max_y p^{\mathrm{plc}}_{t}(y\mid x)$, matching Eq.~(7).

\begin{lemma}[Logit form of PLC]
\label{lem:plc_logit}
$\hat q^{\mathrm{plc}}_t(x)$ is equivalently
\[
\hat q^{\mathrm{plc}}_t(x)
=
\arg\max_{y\in[C]}\Big\{\log p^{\mathrm{pre}}_t(y\mid x)-\tau_{1}\log\pi^{s}_t(y)\Big\}.
\]
\end{lemma}

\paragraph{Overlap score and weight (implementation-aligned).}
Define
\[
\alpha_{\mathrm{pre}}(x):=\max_{c} p^{\mathrm{pre}}_t(c\mid x),\quad
\alpha_{\mathrm{plc}}(x):=\max_{c} p^{\mathrm{plc}}_{t}(c\mid x),\quad
\eta_t(x):=\max_{c} \, p^{\mathrm{pre}}_t(c\mid x)\,p^{\mathrm{plc}}_{t}(c\mid x)\in[0,1].
\]
The implementation uses $\psi_t(x)=\gamma_t\eta_t(x)$ (Eq.~(8--9)) for a global scalar $\gamma_t$.

\begin{lemma}[Self-paced gating property of $\eta$ (no accuracy assumption)]
\label{lem:self_paced_eta}
For any $x$,
\[
\eta_t(x)\le \alpha_{\mathrm{pre}}(x)\,\alpha_{\mathrm{plc}}(x).
\]
If moreover the maximizer class agrees (i.e., $\arg\max p^{\mathrm{pre}}_t(\cdot\mid x)=\arg\max p^{\mathrm{plc}}_t(\cdot\mid x)$)
and is unique, then
\[
\eta_t(x)= \alpha_{\mathrm{pre}}(x)\,\alpha_{\mathrm{plc}}(x).
\]
Consequently, large $\eta_t(x)$ is only possible when both (i) both distributions are confident and (ii) they agree on the top class.
\end{lemma}

\begin{lemma}[Disagreement implies small overlap]
\label{lem:overlap}
If $\arg\max p^{\mathrm{pre}}_t(\cdot\mid x)\neq \arg\max p^{\mathrm{plc}}_{t}(\cdot\mid x)$, then
\[
\eta_t(x)\le 1-\min\{\alpha_{\mathrm{pre}}(x),\alpha_{\mathrm{plc}}(x)\}.
\]
\end{lemma}

\paragraph{How \algo\ avoids ``early inaccurate prior'' harm}
Lemma~\ref{lem:self_paced_eta} is the crucial non-circular mechanism statement:
even if $\pi^{s}_t$ is inaccurate early (hence PLC is noisy), the unlabeled term is automatically suppressed when
the model is uncertain (small $\alpha_{\mathrm{pre}},\alpha_{\mathrm{plc}}$), and also suppressed when PLC flips labels
(Lemma~\ref{lem:overlap}).
Thus the balanced branch is not forced to learn from unstable pseudo-labels at the beginning;
the interaction ramps up only as the standard branch becomes confident, breaking the ``you must already be accurate''
loop.


\subsection{Decoupled Distribution Alignment (DDA) on Standard Branch (Eq.~(4)) and Margin Robustness}

Let $p^{\mathrm{std}}_t(\cdot\mid x):=\delta(f_t(x))$.
DDA uses $\tilde\pi_t$ to form
\begin{equation}
\label{eq:dda_def}
p^{\mathrm{dda}}_{t}(y\mid x)
=
\frac{p^{\mathrm{std}}_t(y\mid x)\cdot \left(\frac{\pi_l(y)}{\tilde\pi_t(y)}\right)^{\tau_{2}}}
{\sum_{c=1}^C p^{\mathrm{std}}_t(c\mid x)\cdot \left(\frac{\pi_l(c)}{\tilde\pi_t(c)}\right)^{\tau_{2}}}.
\end{equation}

So the logit form of DDA is:
\begin{equation}
\label{lem:dda_logit}
\arg\max_{y\in[C]} p^{\mathrm{dda}}_{t}(y\mid x)
=
\arg\max_{y\in[C]}\Big\{\log p^{\mathrm{std}}_t(y\mid x)+\tau_{2}\big(\log\pi_l(y)-\log\tilde\pi_t(y)\big)\Big\}.
\end{equation}

\begin{lemma}[Argmax robustness to log-prior perturbation]
\label{lem:dda_margin_stable}
Define $\delta_t^{(e)}:=\|\log \tilde\pi_t-\log \pi_u\|_\infty$ and
\[
s^{\mathrm{dda}}_t(y\mid x;\pi):=\log p^{\mathrm{std}}_t(y\mid x)+\tau_{2}\big(\log\pi_l(y)-\log\pi(y)\big).
\]
Let $y^\star(x;\pi_u)$ be the argmax and define the margin
\[
\Delta^{\mathrm{dda}}_t(x;\pi_u)
:=
s^{\mathrm{dda}}_t\big(y^\star(x;\pi_u)\mid x;\pi_u\big)
-
\max_{y\neq y^\star(x;\pi_u)} s^{\mathrm{dda}}_t(y\mid x;\pi_u).
\]
If $\Delta^{\mathrm{dda}}_t(x;\pi_u) > 2\tau_{2} \delta_t^{(e)}$, then the argmax under $\tilde\pi_t$ equals that under $\pi_u$.
\end{lemma}

\paragraph{Remark}
Lemma~\ref{lem:dda_margin_stable} is a conservative robustness statement (prevents catastrophic flips on high-margin points).
We do \emph{not} claim it proves performance gain by itself.

\subsection{Eventual Stationarity Without Circular ``Assume Convergence''}

We now give a non-circular stationarity theorem.
Instead of assuming the scores converge, we use a standard optimization-tail condition that implies the logits
(and hence the coupling signals) are Cauchy.

\begin{lemma}[Cauchy logits under summable step sizes (one sufficient condition)]
\label{lem:cauchy_logits}
Let $\theta_t$ be parameters of a branch and $h_{\theta}(x)\in\mathbb{R}^C$ its logits (either $f$ or $\tilde f$).
Assume:
(i) the update satisfies $\theta_{t+1}=\theta_t-\alpha_t g_t$ with $\|g_t\|_2\le G$ almost surely;
(ii) $\sum_{t=0}^\infty \alpha_t<\infty$;
(iii) $h_{\theta}(x)$ is $L$-Lipschitz in parameters: $\|h_{\theta}(x)-h_{\theta'}(x)\|_\infty\le L\|\theta-\theta'\|_2$.
Then for every $x$,
\[
\sum_{t=0}^\infty \|h_{\theta_{t+1}}(x)-h_{\theta_t}(x)\|_\infty < \infty,
\]
hence $h_{\theta_t}(x)$ converges (Cauchy) as $t\to\infty$.
\end{lemma}

\begin{theorem}[Eventual stationarity of PLC pseudo-labels and overlap weights (mean-field)]
\label{thm:stationary}
Assume the mean-field EMA prior updates \eqref{eq:ema_update_generic} for both $\tilde\pi_t$ and $\pi^{s}_t$,
with $\bar p_t$ defined as population prediction marginals.
Assume furthermore that both branches satisfy the conditions of Lemma~\ref{lem:cauchy_logits}
(so $f_t(x)$ and $\tilde f_t(x)$ converge pointwise), and that for almost every $x$,
the limiting PLC score vector
\[
u_\infty(y\mid x):=\lim_{t\to\infty}\Big(\log \delta(f_t(x))_y-\tau_{1}\log \pi^{s}_t(y)\Big)
\]
has a unique maximizer (no ties).
Then for almost every $x$:
\begin{enumerate}
\item (\textbf{PLC pseudo-label stabilization}) There exists $T(x)$ such that for all $t\ge T(x)$,
$\hat q^{\mathrm{plc}}_t(x)$ is constant.
\item (\textbf{Prior stabilization}) The EMA priors $\tilde\pi_t$ and $\pi^{s}_t$ converge in $\ell_1$.
\item (\textbf{Overlap/weight stabilization}) $\eta_t(x)$ converges. If $\gamma_t\to\gamma_\infty<\infty$, then $\psi_t(x)=\gamma_t\eta_t(x)$ converges.
\end{enumerate}
\end{theorem}

\paragraph{What this theorem does and does not claim}
Theorem~\ref{thm:stationary} proves a \emph{stationarity} property of the coupled discrete/continuous signals
under a standard ``optimization tail'' condition.
It does \emph{not} claim convergence to a globally optimal classifier, nor that the stationary point is correct.

\subsection{Post-hoc Adjustment (Eq.~(11)) and Robustness}

At inference,
\[
\hat q^{(b)}(x)=\arg\max_y\{\tilde f_t(x)_y-\tau_{3}\log\tilde\pi_t(y)\}.
\]

\begin{lemma}[Probability form of post-hoc adjustment]
\label{lem:posthoc_prob}
Let $p_t(y\mid x)=\delta(\tilde f_t(x))_y$ and $p_t^{\mathrm{adj}}(y\mid x)=\delta(\tilde f_t(x)-\tau_{3}\log\tilde\pi_t)_y$.
Then
\[
p_t^{\mathrm{adj}}(y\mid x)
=
\frac{p_t(y\mid x)\cdot \tilde\pi_t(y)^{-\tau_{3}}}
{\sum_{c=1}^C p_t(c\mid x)\cdot \tilde\pi_t(c)^{-\tau_{3}}}.
\]
\end{lemma}

\begin{lemma}[Post-hoc argmax robustness to prior error]
\label{lem:posthoc_margin}
Fix $x$ and define scores
\[
r_t(y\mid x;\pi):=\tilde f_t(x)_y-\tau_{3}\log\pi(y).
\]
Let $y^\star(x;\pi_u)=\arg\max_y r_t(y\mid x;\pi_u)$ and define the margin
\[
\Gamma_t(x;\pi_u)
:=
r_t\big(y^\star(x;\pi_u)\mid x;\pi_u\big)
-
\max_{y\neq y^\star(x;\pi_u)} r_t(y\mid x;\pi_u).
\]
If $\Gamma_t(x;\pi_u)>2\tau_{3}\|\log\tilde\pi_t-\log\pi_u\|_\infty$, then
\[
\arg\max_y r_t(y\mid x;\tilde\pi_t)=\arg\max_y r_t(y\mid x;\pi_u).
\]
\end{lemma}

\section{Proofs}
\label{sec:th_proof}

\subsection{Proof of Proposition~\ref{prop:bs_recover}}
\begin{proof}
At a realizable population optimum, $p_{\tilde f^\star}^{\pi_l}(\cdot\mid x)=p_l(\cdot\mid x)$ almost surely.
Thus for each $x,y$,
\[
\frac{\exp(\tilde f^\star_y(x))\pi_l(y)}{\sum_c \exp(\tilde f^\star_c(x))\pi_l(c)}
=
\frac{\pi_l(y)p(x\mid y)}{\sum_c \pi_l(c)p(x\mid c)}.
\]
Cancel $\pi_l(y)$ to obtain $\exp(\tilde f^\star_y(x))=A(x)p(x\mid y)$ for some $A(x)>0$ independent of $y$,
hence $\tilde f^\star_y(x)=\log p(x\mid y)+b(x)$ where $b(x):=\log A(x)$.
\end{proof}

\subsection{Proof of Proposition~\ref{prop:any_prior}}
\begin{proof}
If $\tilde f_y(x)=\log p(x\mid y)+b(x)$ with $b(x)$ independent of $y$, then for any $\pi_{\mathrm{tar}}$,
\[
\arg\max_y\{\tilde f_y(x)+\log\pi_{\mathrm{tar}}(y)\}
=
\arg\max_y\{\log p(x\mid y)+\log\pi_{\mathrm{tar}}(y)\},
\]
since adding $b(x)$ does not change the argmax over $y$.
Under label shift, the right-hand side is the Bayes rule for target prior $\pi_{\mathrm{tar}}$.
\end{proof}

\subsection{Proof of Lemma~\ref{lem:ema_closed}}
\begin{proof}
Unroll $\pi_{t+1}=m\pi_t+(1-m)\bar p_t$ to obtain the closed form.
For the step bound,
\[
\|\pi_{t+1}-\pi_t\|_1
=(1-m)\|\bar p_t-\pi_t\|_1\le 2(1-m),
\]
since any two simplex vectors differ by at most $2$ in $\ell_1$.
\end{proof}

\subsection{Proof of Lemma~\ref{lem:ema_track}}
\begin{proof}
Define $e_t:=\pi_t-\bar p_t$. Then
\[
e_{t+1}
=\pi_{t+1}-\bar p_{t+1}
=m\pi_t+(1-m)\bar p_t-\bar p_{t+1}
=m(\pi_t-\bar p_t)+(\bar p_t-\bar p_{t+1})
=me_t+(\bar p_t-\bar p_{t+1}).
\]
Iterating,
\[
e_t=m^t e_0+\sum_{k=1}^{t}m^{t-k}(\bar p_{k-1}-\bar p_{k}),
\]
hence \eqref{eq:ema_track_exact} by triangle inequality.
If $\bar p_t\to\bar p_\infty$, then $\|\bar p_k-\bar p_{k-1}\|_1\to 0$ and the convolution with $m^{t-k}$ vanishes,
so $\pi_t-\bar p_t=e_t\to 0$ and thus $\pi_t\to \bar p_\infty$.
\end{proof}

\subsection{Proof of Theorem~\ref{thm:piA_decomp}}
\begin{proof}
Let $X_j:=\delta(\tilde f_t(x^{(u)}_j))\in\Delta^{C-1}$, and let $\mu:=\mathbb{E}[X_j]=\pi_t^{\mathrm{pred}}$.
Define $\hat\mu:=\frac{1}{M}\sum_{j=1}^{M}X_j=\hat\pi_t$.

\emph{Step 1 (McDiarmid).}
Consider $F(X_1,\dots,X_M):=\big\|\frac{1}{M}\sum_{j=1}^{M}X_j-\mu\big\|_1$.
Changing one sample changes the mean by at most $\frac{2}{M}$ in $\ell_1$,
so McDiarmid gives with probability at least $1-\nu$,
\[
\|\hat\mu-\mu\|_1
\le
\mathbb{E}\|\hat\mu-\mu\|_1+\sqrt{\frac{2\log(1/\nu)}{M}}.
\]

\emph{Step 2 (bound the expectation).}
By symmetrization and $\|v\|_1=\sup_{\|w\|_\infty\le 1}\langle w,v\rangle$,
\[
\mathbb{E}\|\hat\mu-\mu\|_1
\le
2\,\mathbb{E}\Big\|\frac{1}{M}\sum_{j=1}^{M}\sigma_j X_j\Big\|_1,
\]
with i.i.d.\ Rademacher $\sigma_j$.
Conditioned on $\{X_j\}$, by Khintchine,
$\mathbb{E}_\sigma\big|\sum_{j}\sigma_j X_j(c)\big|\le \sqrt{\sum_j X_j(c)^2}$.
Thus
\[
\mathbb{E}_\sigma\Big\|\sum_{j}\sigma_j X_j\Big\|_1
\le
\sum_{c=1}^{C}\sqrt{\sum_{j=1}^{M}X_j(c)^2}
\le
\sqrt{C\sum_{c=1}^{C}\sum_{j=1}^{M}X_j(c)^2}.
\]
Since $X_j\in\Delta^{C-1}$, $\sum_c X_j(c)^2\le 1$, so the last term is $\le \sqrt{CM}$.
Therefore $\mathbb{E}\|\hat\mu-\mu\|_1\le 2\sqrt{C/M}$.
Combine with Step 1 and then triangle inequality
$\|\hat\pi_t-\pi_u\|_1\le \|\pi_t^{\mathrm{pred}}-\pi_u\|_1+\|\hat\pi_t-\pi_t^{\mathrm{pred}}\|_1$.

For the second claim, since $\pi_u=\mathbb{E}[p_u(\cdot\mid x)]$,
\[
\|\pi_t^{\mathrm{pred}}-\pi_u\|_1
=
\Big\|\mathbb{E}\big[\delta(\tilde f_t(x))-p_u(\cdot\mid x)\big]\Big\|_1
\le
\mathbb{E}\big\|\delta(\tilde f_t(x))-p_u(\cdot\mid x)\big\|_1
\]
by Jensen.
\end{proof}

\subsection{Proof of Proposition~\ref{prop:unif_posterior}}
\begin{proof}
If $\tilde f_y(x)=\log p(x\mid y)+b(x)$, then
\[
\delta(\tilde f(x))_y
=
\frac{\exp(\log p(x\mid y)+b(x))}{\sum_c \exp(\log p(x\mid c)+b(x))}
=
\frac{p(x\mid y)}{\sum_c p(x\mid c)}.
\]
Taking expectation over $x\sim p_u(x)$ gives $\pi^{\mathrm{unif}}$.
\end{proof}

\subsection{Proof of Lemma~\ref{lem:plc_logit}}
\begin{proof}
From \eqref{eq:plc_def}, the unnormalized score is
$p^{\mathrm{pre}}_t(y\mid x)\,(\pi^{s}_t(y))^{-\tau_{1}}$.
Taking logs yields $\log p^{\mathrm{pre}}_t(y\mid x)-\tau_{1}\log\pi^{s}_t(y)$.
The normalizer does not depend on $y$, hence the argmax form.
\end{proof}

\subsection{Proof of Lemma~\ref{lem:self_paced_eta}}
\begin{proof}
For any $c$, $p^{\mathrm{pre}}_t(c\mid x)\le \alpha_{\mathrm{pre}}(x)$ and
$p^{\mathrm{plc}}_t(c\mid x)\le \alpha_{\mathrm{plc}}(x)$, hence
$p^{\mathrm{pre}}_t(c\mid x)p^{\mathrm{plc}}_t(c\mid x)\le \alpha_{\mathrm{pre}}(x)\alpha_{\mathrm{plc}}(x)$.
Taking max over $c$ gives $\eta_t(x)\le \alpha_{\mathrm{pre}}(x)\alpha_{\mathrm{plc}}(x)$.

If the unique maximizer agrees, say $a=\arg\max p^{\mathrm{pre}}_t(\cdot\mid x)=\arg\max p^{\mathrm{plc}}_t(\cdot\mid x)$, then
$p^{\mathrm{pre}}_t(a\mid x)=\alpha_{\mathrm{pre}}(x)$ and $p^{\mathrm{plc}}_t(a\mid x)=\alpha_{\mathrm{plc}}(x)$,
so $\eta_t(x)\ge p^{\mathrm{pre}}_t(a\mid x)p^{\mathrm{plc}}_t(a\mid x)=\alpha_{\mathrm{pre}}(x)\alpha_{\mathrm{plc}}(x)$.
Combine with the upper bound to get equality.
\end{proof}

\subsection{Proof of Lemma~\ref{lem:overlap}}
\begin{proof}
Let $a:=\arg\max p^{\mathrm{pre}}_t(\cdot\mid x)$ and $b:=\arg\max p^{\mathrm{plc}}_t(\cdot\mid x)$ with $a\neq b$.
Then $p^{\mathrm{pre}}_t(a\mid x)=\alpha_{\mathrm{pre}}(x)$ and $p^{\mathrm{plc}}_t(b\mid x)=\alpha_{\mathrm{plc}}(x)$.
Since $a\neq b$, $p^{\mathrm{plc}}_t(a\mid x)\le 1-\alpha_{\mathrm{plc}}(x)$, so
\[
p^{\mathrm{pre}}_t(a\mid x)p^{\mathrm{plc}}_t(a\mid x)\le \alpha_{\mathrm{pre}}(x)(1-\alpha_{\mathrm{plc}}(x))\le 1-\alpha_{\mathrm{plc}}(x).
\]
Similarly $p^{\mathrm{pre}}_t(b\mid x)\le 1-\alpha_{\mathrm{pre}}(x)$, so
\[
p^{\mathrm{pre}}_t(b\mid x)p^{\mathrm{plc}}_t(b\mid x)\le (1-\alpha_{\mathrm{pre}}(x))\alpha_{\mathrm{plc}}(x)\le 1-\alpha_{\mathrm{pre}}(x).
\]
For any other $c\notin\{a,b\}$, both factors are $\le 1-\alpha_{\mathrm{pre}}(x)$ and $\le 1-\alpha_{\mathrm{plc}}(x)$, so the product
is $\le (1-\alpha_{\mathrm{pre}}(x))(1-\alpha_{\mathrm{plc}}(x))\le 1-\min\{\alpha_{\mathrm{pre}}(x),\alpha_{\mathrm{plc}}(x)\}$.
Taking the maximum over $c$ yields the claim.
\end{proof}

\subsection{Proof of Lemma~\ref{lem:dda_margin_stable}}
\begin{proof}
For any $y$,
\[
|s^{\mathrm{dda}}_t(y\mid x;\tilde\pi_t)-s^{\mathrm{dda}}_t(y\mid x;\pi_u)|
=
\tau_{2}|\log\tilde\pi_t(y)-\log\pi_u(y)|
\le \tau_{2}\delta_t^{(e)}.
\]
Thus both the top score and the best competing score can change by at most $\tau_{2}\delta_t^{(e)}$,
so the margin can shrink by at most $2\tau_{2}\delta_t^{(e)}$.
If the margin remains larger, the ordering cannot change.
\end{proof}

\subsection{Proof of Lemma~\ref{lem:cauchy_logits}}
\begin{proof}
Since $\theta_{t+1}-\theta_t=-\alpha_t g_t$ and $\|g_t\|_2\le G$, we have
$\|\theta_{t+1}-\theta_t\|_2\le G\alpha_t$.
By Lipschitzness,
\[
\|h_{\theta_{t+1}}(x)-h_{\theta_t}(x)\|_\infty
\le L\|\theta_{t+1}-\theta_t\|_2
\le LG\alpha_t.
\]
Summing over $t$ gives $\sum_t \|h_{\theta_{t+1}}(x)-h_{\theta_t}(x)\|_\infty \le LG\sum_t \alpha_t<\infty$.
Hence $h_{\theta_t}(x)$ is Cauchy and converges.
\end{proof}

\subsection{Proof of Theorem~\ref{thm:stationary}}
\begin{proof}
By Lemma~\ref{lem:cauchy_logits}, for every $x$, $f_t(x)\to f_\infty(x)$ and $\tilde f_t(x)\to \tilde f_\infty(x)$.
By continuity of softmax on $\mathbb{R}^C$, $\delta(f_t(x))\to \delta(f_\infty(x))$ pointwise.
By dominated convergence (bounded in $[0,1]$), the population marginals
$\bar p_t=\mathbb{E}_{x\sim p_u}[\delta(f_t(x))]$ and similarly for $\tilde f_t$ converge.
Then Lemma~\ref{lem:ema_track} implies $\pi^{s}_t$ and $\tilde\pi_t$ converge.

Now $\log\delta(f_t(x))_y$ is continuous in $f_t(x)$ whenever probabilities stay positive; this holds since logits converge to finite
values and softmax outputs are in the simplex interior.
Also $\log\pi^{s}_t(y)$ converges since $\pi^{s}_t$ converges in the interior.
Therefore the PLC score vector
$u_t(y\mid x)=\log \delta(f_t(x))_y-\tau_{1}\log \pi^{s}_t(y)$ converges to $u_\infty(\cdot\mid x)$.

By the unique-maximizer assumption, there exists $T(x)$ such that for all $t\ge T(x)$ the argmax of $u_t(\cdot\mid x)$ equals that of
$u_\infty(\cdot\mid x)$, hence $\hat q^{\mathrm{plc}}_t(x)$ is constant.

Finally, $p^{\mathrm{pre}}_t(\cdot\mid x)=\delta(f_t(x))$ and
$p^{\mathrm{plc}}_t(\cdot\mid x)=\delta(u_t(\cdot\mid x))$ converge by continuity, hence
$\eta_t(x)=\max_c p^{\mathrm{pre}}_t(c\mid x)p^{\mathrm{plc}}_t(c\mid x)$ converges as a continuous mapping.
If $\gamma_t\to\gamma_\infty$, then $\psi_t(x)=\gamma_t\eta_t(x)$ converges.
\end{proof}

\subsection{Proof of Lemma~\ref{lem:posthoc_prob}}
\begin{proof}
Expand the softmax:
\[
\delta(\tilde f_t(x)-\tau_{3}\log\tilde\pi_t)_y
=
\frac{\exp(\tilde f_t(x)_y)\tilde\pi_t(y)^{-\tau_{3}}}
{\sum_c \exp(\tilde f_t(x)_c)\tilde\pi_t(c)^{-\tau_{3}}}.
\]
Divide numerator and denominator by $\sum_c\exp(\tilde f_t(x)_c)$ to obtain the stated probability form.
\end{proof}

\subsection{Proof of Lemma~\ref{lem:posthoc_margin}}
\begin{proof}
For any $y$,
\[
|r_t(y\mid x;\tilde\pi_t)-r_t(y\mid x;\pi_u)|
=
\tau_{3}|\log\tilde\pi_t(y)-\log\pi_u(y)|
\le \tau_{3}\|\log\tilde\pi_t-\log\pi_u\|_\infty.
\]
Thus the top-1 and runner-up scores can each change by at most that amount,
so the margin can shrink by at most $2\tau_{3}\|\log\tilde\pi_t-\log\pi_u\|_\infty$.
If $\Gamma_t(x;\pi_u)$ exceeds it, the argmax cannot change.
\end{proof}

\vspace{1em}


\clearpage

\vfill

\end{document}